\documentclass[11pt, a4paper, goog]{google}

\usepackage[T1]{fontenc}
\usepackage[utf8]{inputenc}
\usepackage[authoryear, sort&compress, round]{natbib}
\usepackage{multirow}
\usepackage{array}
\usepackage{tabularx}
\usepackage{pdflscape}
\usepackage{arydshln}
\usepackage{placeins}
\usepackage{tikz}
\usetikzlibrary{arrows.meta, positioning, fit, backgrounds, calc, shapes.geometric}
\usepackage[most]{tcolorbox}
\usepackage{soul}

\newcommand{\yes}{\ding{51}}
\newcommand{\no}{\ding{55}}
\newcommand{\partsup}{$\diamond$}

\definecolor{datablue}{HTML}{4472C4}
\definecolor{sharedgray}{HTML}{6C757D}
\definecolor{retgreen}{HTML}{548235}
\definecolor{tvorange}{HTML}{C55A11}
\definecolor{lightbg}{HTML}{F2F2F2}
\definecolor{lightgreen}{HTML}{E2EFDA}
\definecolor{lightorange}{HTML}{FCE4CC}
\definecolor{lightblue}{HTML}{D6E4F0}
\definecolor{antgreen}{HTML}{548235}
\definecolor{conpurple}{HTML}{7030A0}
\definecolor{highcolor}{HTML}{2E75B6}
\definecolor{medcolor}{HTML}{BF8F00}
\definecolor{lowcolor}{HTML}{C00000}
\definecolor{lightpurple}{HTML}{E8D5F5}

\newcommand{\gymname}{\textsc{FinanceGym}}
\newcommand{\workname}{\textsc{FinanceHarness}}

\uselogo{}

\title{\textbf{FinanceHarness}: Autonomous Financial Deep Research Framework}

\worknote{* This work was done while Yijia Xiao was a Student Researcher at Google Cloud AI Research.}

\correspondingauthor{yijia.xiao@cs.ucla.edu, rujunh@google.com, weiwang@cs.ucla.edu, chenyulee@google.com}

\renewcommand{\today}{2026-08-06}

\author[1,2]{Yijia Xiao\textsuperscript{*}}
\author[1]{Rujun Han}
\author[1]{Yanfei Chen}
\author[1]{Zifeng Wang}
\author[1]{Ke Jiang}
\author[1]{Zhongying CuiZhu}
\author[1]{Vishy Tirumalashetty}
\author[2]{Wei Wang}
\author[1]{Burak Gokturk}
\author[1]{Tomas Pfister}
\author[1]{Chen-Yu Lee}

\affil[1]{Google Cloud AI Research}
\affil[2]{University of California, Los Angeles}

\begin{abstract}
Powered by advances in LLMs and autonomous agents, deep research has become one of the most widely adopted agentic products. However, most deep research systems write general-purpose reports, which are inadequate for financial deep research. Financial research demands specialized knowledge to analyze historical patterns and forecast upcoming events. Automating financial deep research therefore requires both a layered harness to drive the research agent and a verifiable, point-in-time benchmark that prevents leakage of future information. We present \workname{}, a harness that runs finance-oriented tools and practitioner-guided workflows, automating financial deep research end to end: environment and data construction, the agent execution loop, and reward modeling. We further propose \gymname{}, comprising thesis-driven research questions and rubrics that combine pre-cutoff and post-cutoff criteria. Professional expert validation yields an 82\% pass rate. With the same open-weight backbone, \workname{} improves the overall rubric score from 25.3\% to 32.4\%, demonstrating the effectiveness of our specialized harness design. However, even pairing \workname{} with the most cutting edge LLM (e.g. Opus-5), the \gymname{} score is below 45\%, showing that it is a challenging benchmark for financial deep research. Leaderboard is available at: \url{https://financegym.github.io/} and \workname{} code is available at: \url{https://github.com/Yijia-Xiao/FinanceHarness}.
\end{abstract}

\begin{document}

\maketitle

\section{Introduction}
\label{sec:intro}

Deep research is now among the most widely adopted agentic products in the industry \citep{OpenAI_DR, Perplexity_DR, team2025tongyi, Gemini_DR}. Equipped with leading LLMs and a carefully designed agent harness, deep research conducts comprehensive searches that can answer challenging multi-hop reasoning questions and synthesize insightful long-form reports from multiple sources and topics \citep{TTD_DR, team2025tongyi, team2025mirothinker, team2026mirothinker}.

Despite the impressive progress, a general deep research harness falls short for highly specialized domains~\citep{wu2023bloomberggpt, xie2024finben, xie2023pixiu}. In this work, we target \emph{financial deep research}, which requires capabilities such as evidence gathering, cross-source validation, and report synthesis that go beyond a generic deep research skillset. It further requires the system to grasp the relationships among financial entities and events, and anticipate their evolution over time~\citep{ding2015deep, xu2018stock}. For example, an equity analyst investigating a semiconductor company needs to examine historical earnings reports, analyze current competitor dynamics, and predict future demand and macroeconomic conditions to derive a reasonable rating for the stock. A general deep research system without financial tools and expertise to analyze earnings reports and economic events may fail to identify specific information for such complex analysis.

Moreover, despite the abundance of recent deep research benchmarks \citep{coelho2025deepresearchgym, du2025deepresearch, li2026deepresearch, sharma2025researchrubrics, wang2025liveresearchbench}, none of them provide scalable evaluation data for financial research. Existing deep research benchmarks focus mostly on testing past or current conditions. Professional financial research reports typically need to reason about future events (post-cutoff reasoning), which requires point-in-time (PIT) evaluation to avoid leakage of future information. On the other hand, existing financial NLP benchmarks target isolated capabilities (sentiment, entity recognition, filing QA)~\citep{shah2022flue, islam2023financebench, yang2023fingpt}, which are insufficient to measure the quality of comprehensive, temporally constrained financial reports.

We address these gaps with a financial deep research suite built on a PIT search sandbox. We construct a large-scale web corpus with real publication dates from the public web, and build a dense retrieval system with cutoff-date access control. This ensures a financial deep research agent can locate finance-related knowledge effectively, while information published beyond each question's cutoff date is excluded from the research reports.

Leveraging this sandbox, we construct \gymname{}, a large-scale, expert-validated financial deep research benchmark. We build an entity graph from the corpus and sample financial situations from this graph to ground each research question and its rubric. Drawing on input from financial practitioners, we generate an investment thesis and paired pre-cutoff and post-cutoff rubrics for every question. A multi-stage filtering pipeline then removes low-quality records, and expert annotators validate the rest to form the final benchmark.

Building on this new environment, we propose \workname{}, an expert-knowledge-guided agent harness for financial deep research (Section~\ref{sec:harness}). Following the common definition in the community, \workname{} implements a harness as layered services of LLM-facing tools, APIs, execution workflows, and evaluation rubrics \citep{lee2026metaharness, ning2026codeasharness}. It runs agents against the PIT sandbox and scores them with the \gymname{} rubrics, so that evaluation and in-environment optimization share one contract.

We summarize our contributions. (1) We build a point-in-time financial search sandbox to serve as the foundation for \gymname{}: an expert-validated financial deep research benchmark of 400 high-quality research questions and rubrics that separate pre-cutoff evidence retrieval from post-cutoff reasoning. (2) We propose \workname{}, an expert-knowledge-guided harness that automates environment and data construction, agent execution, and reward modeling for financial deep research under a strict point-in-time contract. (3) We benchmark a wide range of leading LLMs and agents and show that \gymname{} is challenging, with every system scoring below 40\%, making it a useful target for advancing financial deep research.

\section{Related Work}
\label{sec:related}

\paragraph{Financial benchmarks.}
Financial NLP benchmarks have traditionally targeted isolated skills such as sentiment analysis, named entity recognition~\citep{shah2022flue}, factual question answering over SEC filings~\citep{islam2023financebench}, or broad evaluation of finance-specialized LLMs~\citep{yang2023fingpt}. These tasks are valuable, but they do not capture analyst-style financial deep research: long-form reports that connect entities, sectors, events, and time. To bridge this gap, \gymname{} builds questions from a finance entity graph and evaluates reports with a two-tier rubric that separates pre-cutoff evidence retrieval from post-cutoff outcome anticipation.

\paragraph{Research agents.}
Modern research agents build on retrieval-augmented generation~\citep{lewis2020retrieval} and ReAct-style tool use~\citep{yao2022react}, but differ in where the research policy lives.
Scaffolded agents encode orchestration in code, including Test-Time Diffusion for Deep Research (TTD-DR)~\citep{TTD_DR}, GPT-Researcher,\footnote{\url{https://github.com/assafelovic/gpt-researcher}}, STORM~\citep{shao2024assisting}, and subagent-based frameworks.
Specialized deep-research models instead train the backbone on agentic-search trajectories with a fixed tool stack, as in Tongyi-DeepResearch-30B-A3B (Tongyi-DR)~\citep{team2025tongyi}
and MiroThinker-1.7-mini~\citep{team2025mirothinker, team2026mirothinker}.
Our evaluation includes trained models, a fixed ReAct search wrapper with
multiple backbones, and agentic search systems on the same PIT corpus retriever,
which lets us separate backbone capability, scaffold contribution, and
tool-distribution shift.
Within finance, LLM agents have largely targeted trading decisions, through multi-agent frameworks~\citep{xiao2024tradingagents} and reasoning models trained with reinforcement learning~\citep{xiao2025trading}, whereas we target
analyst-style research and report generation evaluated with verifiable rubrics.

\paragraph{Deep-research evaluation.}
Recent deep-research benchmarks emphasize long-form, citation-grounded answers
and richer rubrics~\citep{coelho2025deepresearchgym, du2025deepresearch,
li2026deepresearch, sharma2025researchrubrics, wang2025liveresearchbench}, but
most rely on live-web retrieval or a single global snapshot rather than a
per-question publication-date cutoff.
LLM-as-judge methods offer scalable evaluation~\citep{zheng2023judging}, yet
finance requires criterion-level attribution: a report may retrieve the right
historical facts while failing to anticipate the later outcome, or reason
plausibly while omitting source-grounded evidence.
We therefore use evidence-aware rubric judging with explicit pre-cutoff and
post-cutoff rubrics.

\begin{table*}[tbp]
\centering
\small
\setlength{\tabcolsep}{4pt}
\caption{Comparison with representative deep-research and search-agent benchmarks. \yes\,/\,\partsup\,/\,\no{} denote full, partial, and no support. \emph{Reproducible} means evaluation uses a fixed corpus rather than live-web access. \emph{PIT} means retrieval is constrained by a per-question publication-date cutoff. \emph{Verifiable} means evaluation rests on externally checkable signals such as gold answers, binary rubric items, citation matches, or post-cutoff-date outcomes.}
\label{tab:positioning}
\resizebox{\textwidth}{!}{
\begin{tabular}{lcccccc}
\toprule
\textbf{Benchmark} & \textbf{Domain} & \textbf{Long-form} & \textbf{Reproducible} & \textbf{PIT} & \textbf{Rubric} & \textbf{Verifiable} \\
\midrule
FinanceBench~\citep{islam2023financebench}                 & Finance      & \no  & \yes     & \no  & Gold answer                     & \yes \\
GAIA~\citep{mialon2024gaia}                                & General      & \no  & \partsup & \no  & Gold answer                     & \yes \\
BrowseComp~\citep{wei2025browsecomp}                       & General      & \no  & \no      & \no  & Gold answer                     & \yes \\
BrowseComp-Plus~\citep{chen2025browsecomp}                 & General      & \no  & \yes     & \no  & Gold answer + citation          & \yes \\
FRAMES~\citep{krishna2025fact}                             & General      & \no  & \yes     & \no  & Gold answer                     & \yes \\
DeepResearchGym~\citep{coelho2025deepresearchgym}          & General      & \yes & \yes     & \no  & Citation + LLM-judge            & \partsup \\
DeepResearch Bench~\citep{du2025deepresearch}              & Multi-domain & \yes & \no      & \no  & RACE + FACT                     & \partsup \\
DeepResearch Bench~II~\citep{li2026deepresearch}           & Multi-domain & \yes & \no      & \no  & 9.4k binary rubrics             & \yes \\
ResearchRubrics~\citep{sharma2025researchrubrics}          & Multi-domain & \yes & \no      & \no  & 2.6k expert rubrics             & \yes \\
LiveResearchBench~\citep{wang2025liveresearchbench}        & Multi-domain & \yes & \no      & \no  & DeepEval checklist              & \yes \\
MiroEval~\citep{ye2026miroeval}                            & Multi-domain & \yes & \no      & \no  & Adaptive + process eval         & \yes \\
DeepScholar-Bench~\citep{patel2025deepscholar}             & Academic     & \yes & \no      & \yes & Automated, 3 dimensions         & \yes \\
\midrule
\rowcolor{lightblue}
\textbf{\gymname{} (Ours)} & \textbf{Finance} & \textbf{\yes} & \textbf{\yes} & \textbf{\yes} & \textbf{Pre-cutoff/post-cutoff rubrics} & \textbf{\yes} \\
\bottomrule
\end{tabular}
}
\end{table*}

Table~\ref{tab:positioning} highlights the gap \gymname{} targets.
FinanceBench~\citep{islam2023financebench} is the closest finance-domain benchmark,
but evaluates short-form QA over filings rather than analyst-style research.
BrowseComp-Plus~\citep{chen2025browsecomp} and
DeepResearchGym~\citep{coelho2025deepresearchgym} improve reproducibility
through fixed corpora, but do not apply per-question publication-date cutoffs.
DeepScholar-Bench~\citep{patel2025deepscholar} is closest on the PIT axis, but
tied to the evolving arXiv index.
DeepResearch Bench~II~\citep{li2026deepresearch} and
ResearchRubrics~\citep{sharma2025researchrubrics} provide rich rubrics, but do
not isolate pre-cutoff-date evidence retrieval from post-cutoff-date outcome
anticipation, which is the central difficulty in financial deep research.

\section{\gymname{} Construction}
\label{sec:method}

\begin{figure*}[!b]
\centering
\includegraphics[width=\textwidth]{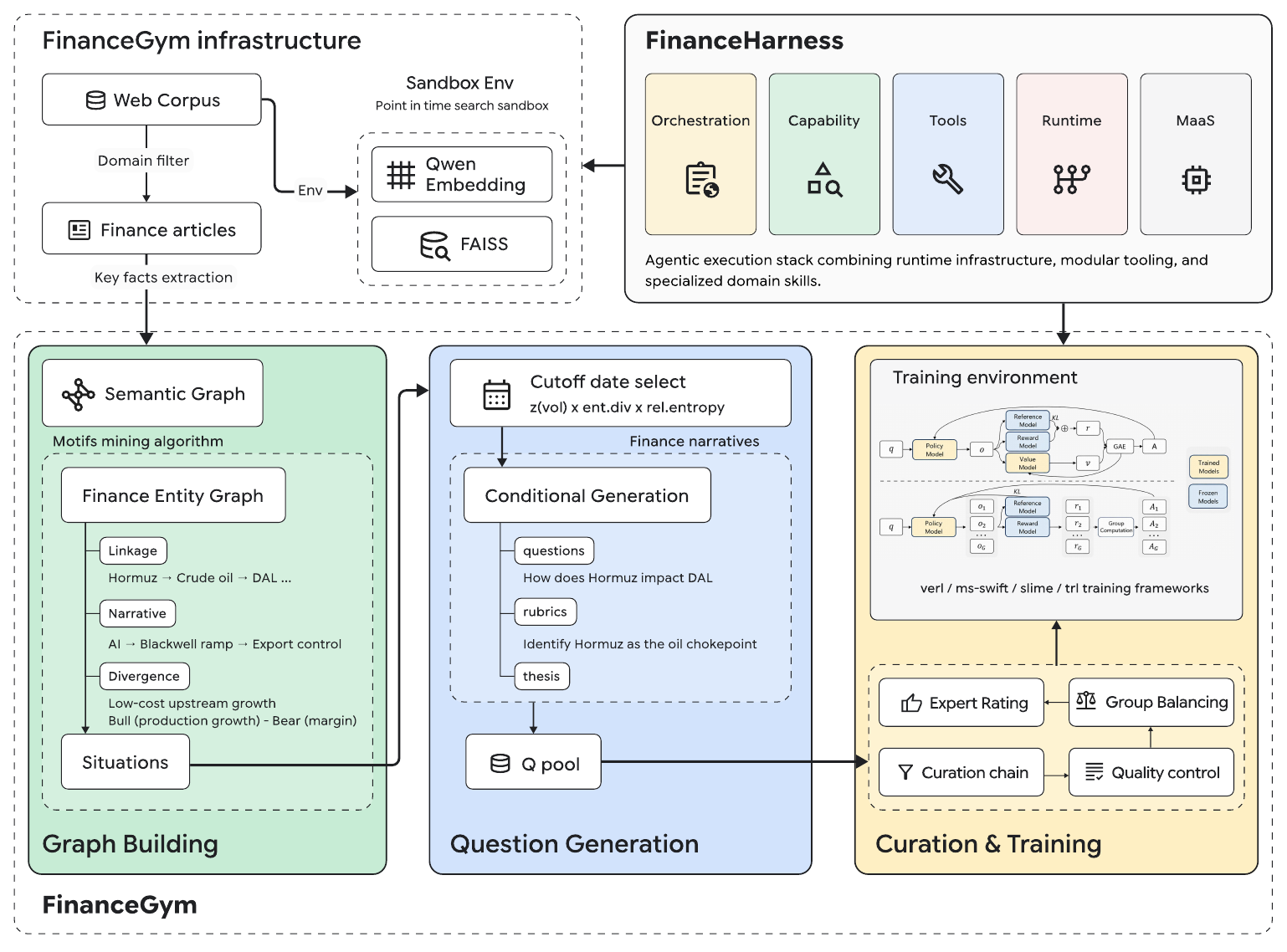}
\caption{End-to-end benchmark construction. Web articles flow through
a domain filter and FAISS + PIT embedding store into a semantic graph, where
motif mining yields the finance entity graph and downstream situations;
publication-date cutoff selection then drives unconstrained question generation,
followed by curation,
balancing, expert annotation, and final evaluation.}
\label{fig:system-overview}
\end{figure*}

As mentioned in Section~\ref{sec:intro}, our benchmark infrastructure consists of a point-in-time (PIT)
financial search sandbox and \gymname{}, the benchmark constructed on top
of it (Figure~\ref{fig:system-overview}). In this section, we first describe
the corpus, retrieval, and temporal access contract that define the search sandbox
(\S\ref{sec:corpus}). We then describe how \gymname{} is generated from a
finance entity graph, filtered, balanced, and expert-annotated
(\S\ref{sec:questions}). Finally, we define the scoring contract used by the
benchmark and by downstream harness training (\S\ref{sec:eval}).

\subsection{Point-in-Time Financial Search Sandbox}
\label{sec:corpus}

\paragraph{Source corpus.}
We built the search sandbox from a large-scale web corpus that we collected from thousands of public web domains.
We built the corpus so that all articles carry reliably extracted
publication dates: each article's date was extracted with
\texttt{htmldate}~\citep{barbaresi2020htmldate}, allowing the environment to
enforce PIT access.
We extracted clean text with
\texttt{trafilatura}~\citep{barbaresi2021trafilatura} with normalized metadata.
The resulting corpus contains 100+ million articles, deliberately
preserving the noise and breadth of web-scale financial search rather than
reducing the task to a curated filing set.
The corpus serves only as the internal testbed for building and evaluating the
harness. Users can input any public corpus in our pipeline to produce research questions and rubrics. The released benchmark consists of standalone research questions that are decoupled from the underlying corpus (\S\ref{sec:questions}).

\paragraph{Embedding and storage.}
\label{sec:embedding}
Articles are embedded with Qwen3-Embedding-4B~\citep{zhang2025qwen3}, producing
normalized dense vectors for retrieval.
Full texts are stored separately from the vector index so agents can retrieve
candidate documents and then fetch source text for citation-grounded synthesis.

\paragraph{Search server.}
\label{sec:search}
We build a FAISS~\citep{douze2025faiss} IVF-SQ8 index over the normalized
embeddings and serve it through a lightweight API.
This API defines the environment contract used throughout the paper: agents can
search and read historical documents, but they cannot access articles published
after the assigned cutoff date.

\subsection{\gymname{}: Situation-Driven Benchmark Construction}
\label{sec:questions}

\paragraph{Finance graph construction.}
\label{sec:kg}
We construct a finance entity graph from the corpus by first filtering to
finance-relevant sources and then extracting entity-relation triples with
Gemini-3.5-Flash~\citep{team2023gemini} under structured JSON output.
Each edge stores a head entity, relation, tail entity, local context, source
URL, and publication date.
The extractor produces 5.74M raw edges; after filtering generic or
malformed nodes, the working graph contains 4.37M edges.
These edges span 1.11M unique entities from 1.20M source
articles and serve as the basis for benchmark generation.

\paragraph{Situation mining.}
\label{sec:situations}
Rather than clustering the graph into broad themes, we mine situations that
match how financial analysts discover research questions.
The miner has three complementary modes: cross-category linkages between
entities and event types, temporal narrative arcs around high-degree entities,
and polar divergences such as upgrades versus downgrades or earnings beats
versus misses.
These modes instantiate the fine-grained situation types reported in the
harness ablation (Appendix~\ref{sec:appendix-harness-ablation}): a multihop-path
type from linkages, a temporal-narrative type, and several divergence
(\emph{tension}) types.
An entity budget limits repeated sampling of the same primary name, preventing
mega-cap firms from dominating the benchmark.

\paragraph{Publication-date cutoff selection and unconstrained generation.}
\label{sec:unconstrained}
Each candidate situation is paired with a cutoff date.
For each candidate day $d$, let $v(d)$ be its event volume, $e(d)$ entity
diversity, and $r(d)$ relation entropy, and let $z(\cdot)$ denote a
$z$-score across candidate days.
Cutoff dates are then selected by a volume, entity-diversity, and
relation-entropy objective:
\begin{equation}
    \text{score}(d) =
    z\bigl(v(d)\bigr)\,
    \bigl(1 + e(d)\bigr)\,
    \bigl(1 + \tfrac{r(d)}{10}\bigr).
\end{equation}
The volume term favors eventful days; the diversity and entropy terms penalize
batch artifacts dominated by a single entity or one relation type.

Given a (situation, cutoff-date) pair, an LLM generates the benchmark record
\emph{without category priming}: the prompt does not provide a topic, sector,
or reasoning taxonomy.
The output contains an analyst-style question, a reference investment thesis,
and a two-tier rubric.
Pre-cutoff criteria test facts findable before the cutoff date; post-cutoff criteria
test outcomes verifiable only after the cutoff date.
This split is central to \gymname{}: it separates evidence retrieval from
forward-looking synthesis while keeping both sides auditable.

\paragraph{Bottom-up taxonomy.}
\label{sec:taxonomy}
After questions are generated, we assign taxonomy labels.
LLMs classify batches of generated questions into natural categories, and the
raw labels are consolidated into three axes: topic, sector, and reasoning type.
This bottom-up taxonomy avoids imposing a prompt-time template on question
generation while still allowing the final benchmark to be balanced and
reported by interpretable groups.

\begin{figure*}
\centering
\includegraphics[width=\textwidth]{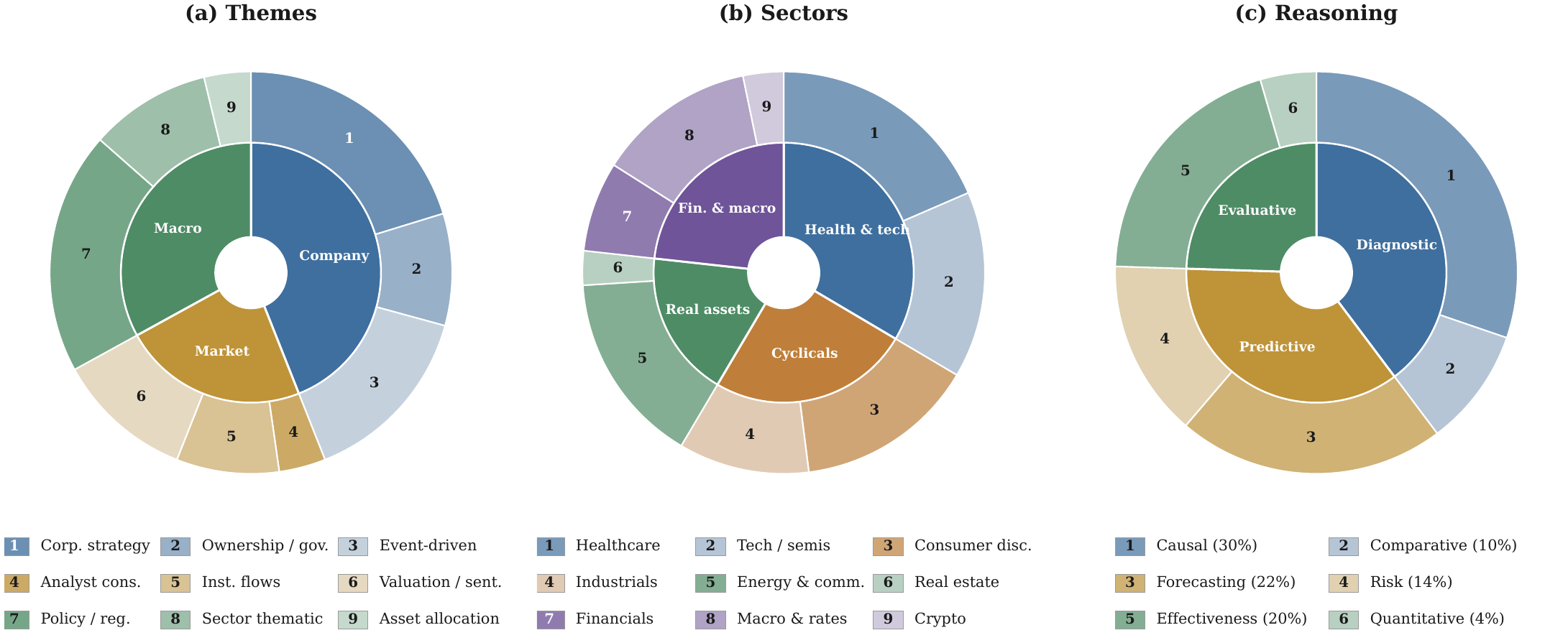}
\caption{Distribution of the 400-question \gymname{} subset across the three
bottom-up taxonomy axes. Topics and sectors are shown as two-ring sunbursts
with super-groups; reasoning types appear as a donut.}
\label{fig:taxonomy-mix}
\end{figure*}

\paragraph{Quality filtering and balancing.}
\label{sec:filters}
Generated questions pass through a sequence of LLM quality gates before
selection for expert review.
The filters test feasibility, institutional relevance, coherence, groundedness
against evidence, and final balance across topic, sector, reasoning type, and
cutoff date.
This yields a data-determined 2{,}078-question publication pool from
29{,}669 unconstrained generations.
For benchmarking, an integer linear program (ILP) selects a 500-question subset that
maximizes quality subject to balance constraints across the taxonomy axes and
monthly cutoff-date buckets.

\begin{figure*}[!ht]
\centering
\includegraphics[width=\textwidth]{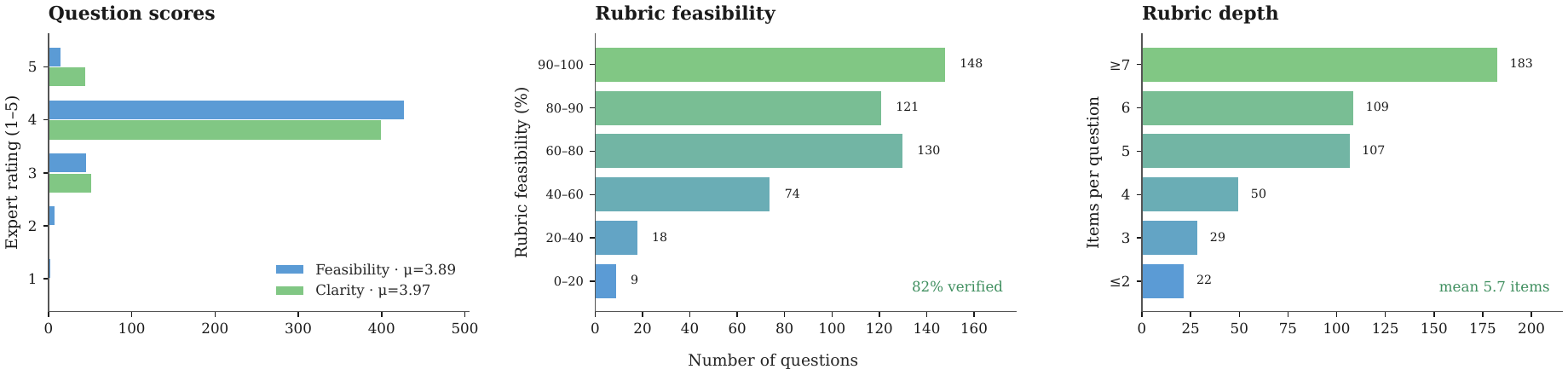}
\caption{Expert-annotation quality over the annotated question pool:
distribution of annotator feasibility and clarity ratings used as the
quality-control gate that curates \gymname{}.}
\label{fig:annotation}
\end{figure*}

\begin{figure*}
\centering
\includegraphics[width=\textwidth]{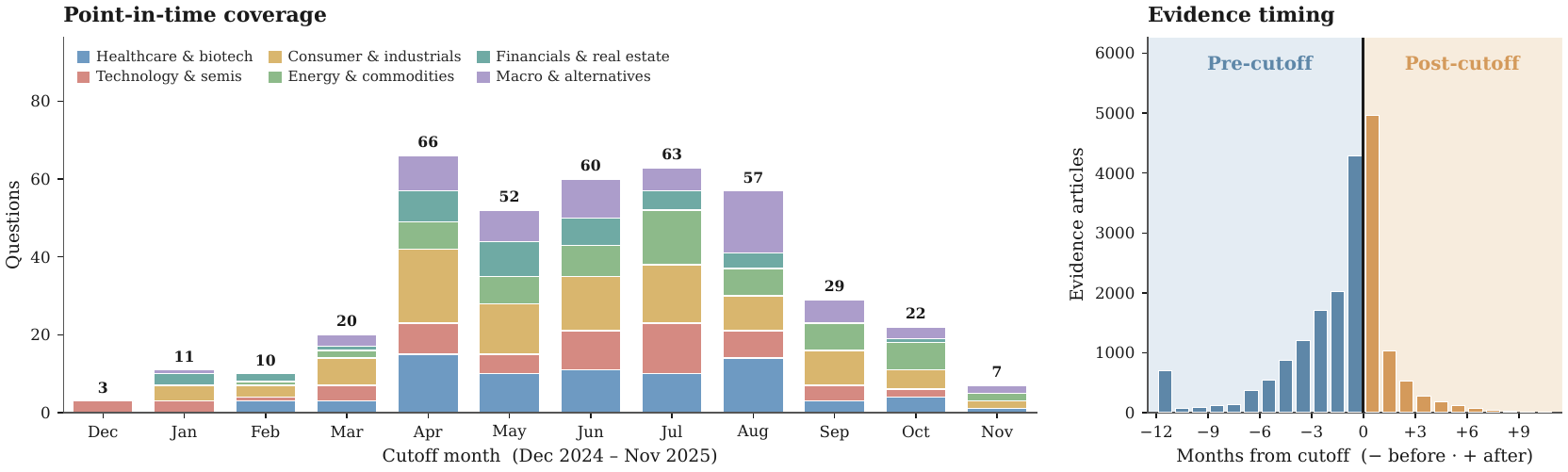}
\caption{Temporal coverage of \gymname{}: distribution of question cutoff dates
across the 12 monthly cutoff-date buckets.}
\label{fig:temporal}
\end{figure*}

\paragraph{Expert annotation.}
\label{sec:human-curation}
The balanced 500-question subset receives a final expert-annotation pass
through an external professional data vendor, followed by LLM-assisted curation
using the annotation labels.
Annotators score question feasibility and clarity and mark each rubric item
as feasible or infeasible with a short justification.
Curation then keeps records with sufficiently clear questions and enough
feasible rubrics to grade an agent report: 411 of the 500
annotated questions meet this bar, an 82\% professional-annotation pass rate.
Each sample requires approximately 1.2 hours of expert work, reflecting the
difficulty of validating analyst-style questions and rubric items under a
publication-date cutoff.
We further balance this pool to the released \gymname{} benchmark of
400 expert-annotated questions with 2{,}464 annotated rubric items, balanced
across 9 topics, 11 sectors (merged into 9 leaves), 6 reasoning types, and 12
monthly cutoff-date buckets (Figures~\ref{fig:taxonomy-mix},
\ref{fig:annotation}, and~\ref{fig:temporal}).
\paragraph{Release policy.}
As a final release gate, human experts review every released question along two axes. For quality, they confirm the question is well posed and answerable from public information, with a clear analytical target. For privacy and provenance, they confirm the question only \emph{queries} information. In other words, it refers to publicly observable market events and asks the agent to research them, rather than embedding, quoting, or leaking content from the underlying corpus, which ensures that each question contains neither personal data nor copyrighted text. Questions that fail any check are removed.

After annotations, we do not publicly release the underlying corpus. We release the research
questions and the report submission code. The grading runs through our leaderboard with private rubrics to preserve the integrity of our scoring criteria.

\subsection{Evaluation Contract}
\label{sec:eval}

\paragraph{Rubric-based scoring.}
\label{sec:scoring}
An LLM judge evaluates each agent report against the question-specific rubric.
For each criterion, the judge assigns a score on a 5-tier (0--4) scale with
explicit anchors:
0~(not addressed),
1~(mentioned),
2~(partial),
3~(substantive), and
4~(fully grounded).
The judge sees the question, thesis, cutoff date, target rubrics, and the agent's
report with cited URLs.
Because the rubrics are professionally validated and evidence-linked, the LLM
judge is used as a consistent rubric applier rather than as a source of new
evaluation criteria.

The primary metric is the \emph{outcome score}: the average over questions of
the fraction of rubric points each report earns,
\begin{equation}
    \text{Outcome} \;=\;
    \frac{1}{|Q|} \sum_{q \in Q}
    \frac{\sum_{c \in \mathcal{R}(q)} \mathrm{score}(q, c)}
    {4 \,|\mathcal{R}(q)|}\,,
    \label{eq:outcome}
\end{equation}
so each question contributes equally regardless of its number of rubric items.
Scores can be disaggregated by criterion type, topic, sector, reasoning type,
or cutoff-date bucket using the same formula.

\paragraph{Point-in-time protocol.}
\label{sec:pit}
Each question has an assigned cutoff date.
During evaluation, the FAISS server enforces PIT filtering: every search result
must have publication date $\leq$ the cutoff date, and agents do not access the live
web.
Pre-cutoff synthesis rubrics test retrieval and synthesis from pre-cutoff evidence.
Post-cutoff reasoning rubrics test whether the agent's analysis anticipates developments
that are only verifiable after the cutoff date.
Any detected PIT leakage invalidates the run rather than being treated as an
ordinary scoring error.

\section{\workname{}}
\label{sec:harness}
In this section, we explain \workname{} design and its connection to \gymname{}. The point-in-time (PIT) search sandbox defines what information an agent may retrieve before each question's cutoff date, and \gymname{} defines how the resulting report is graded.
\workname{} is the executable harness built around that contract, which enables users to produce highly specialized reports for their financial research queries.
It provides the same retrieval tools, orchestration loop, and rubric signal for both evaluation and optimization, so a model trained inside the harness sees the same tool-use distribution at test time.

\subsection{Expert-guided Harness Design}
Financial deep research is not only a retrieval problem: an analyst must connect relationships among entities, sector context, numerical evidence, and forward-looking risks into an auditable view.
\workname{} therefore exposes a layered finance tool interface rather than a single generic web-search action. The core loop covers PIT search, source reading, citation composition, and report finalization. Auxiliary finance tools and workflow skills add structured support for exact data retrieval, calculation, valuation, comparables, risk analysis, and scenario reasoning without expanding the base prompt for every query.

\paragraph{Model and serving layer.}
The serving layer hosts the orchestrator backbone and a lightweight reader used
for long-document evidence extraction.
It is deliberately opaque to the rest of the harness: the runtime sees only a
request/response contract, so the backbone can be swapped across local serving
stacks or closed-model SDKs without changing the orchestration logic.

\paragraph{Runtime layer.}
The runtime is the control plane.
It owns the bounded agent loop, parsing and dispatch, schema validation, tier
loading, prompt-mode selection, recovery, and the registries for tools and
workflow skills.
It contains no model intelligence; instead, it enforces cross-layer invariants
such as schema conformance, tool-result chaining, citation finalization, and
run-level budget limits.

\paragraph{Tool surface.}
The tool surface is tiered to keep the initial prompt small while preserving a
broad set of financial capabilities.
The always-loaded tier contains the core-loop actions.
Deferred tools are exposed through a compact catalog, with full schemas loaded
only after the model commits to a tool family.
The extension tier supports MCP or CLI-backed connectors, including paid data
vendors and internal systems.

\paragraph{Modes and skills.}
\workname{} uses prompt-variant modes over a consistent tool surface.
The research mode emphasizes web-first evidence gathering; the analytical mode
emphasizes data and computation tools; the automatic mode lets the agent choose
between them.
Keeping the registry consistent across modes prevents a trajectory from being
stranded by tools that disappear after a mode switch.
Skills are reusable workflow specifications that the model can activate on
demand.
Each skill declares the tools and specialist context it expects, and routing is
based on the skill description rather than a hard-coded name.

\paragraph{Grounding and robustness.}
The orchestrator can respond to a step with direct tool calls or by loading a
workflow skill that brings in the tools needed for a structured analysis.
For research-style reports, the harness composes numbered citations from visited
sources, can retrofit citations when a draft omits the citation tool, and runs a
light grounding-review pass that asks the backbone to soften or attribute claims
it cannot tie to read sources or tool outputs.
Recovery policies handle transient model failures, malformed tool calls,
context-budget pressure, and truncation without changing the evaluation contract.

\paragraph{Operational guardrails.}
URL pre-fetch validates document IDs before expensive reads.
In ablations, disabling this guardrail raises the visit error rate from
$2.1\%$ to $39.4\%$; the end score barely moves because the backbone often
self-corrects, but trajectories become substantially more expensive.
Search-budget caps show a different pattern: excessive searching correlates with
question difficulty rather than causing low scores, so budget caps mainly
control cost rather than quality.

\subsection{Evaluation and Optimization}
The harness is tightly coupled to the environment: tool calls hit the PIT search sandbox's FAISS server directly, and the rubric judge used for benchmark scoring is also available as a reward component.
This matters because specialized research models are often trained against one
tool stack and evaluated against another.
In \workname{}, swapping the backbone model changes only the learnable model layer. The search API, document-fetch contract, citation requirements, and scoring rubric stay fixed.
Appendices~\ref{sec:appendix-cases} and~\ref{sec:appendix-traces} give worked examples: complete cited reports produced by the harness, and the round-by-round tool trajectories behind them.

\section{Experimental Setup}
\label{sec:experiments}

\subsection{Benchmark Dataset}
\label{sec:dataset}

The publication dataset comprises 2{,}078 questions, the data-determined output of the end-to-end LLM quality filter chain (\S\ref{sec:filters}).
For benchmarking, the upstream pipeline selects a 500-question ILP-balanced subset, which then passes through external expert annotation and LLM-assisted curation (\S\ref{sec:human-curation}) to produce \gymname{}: 400 expert-annotated questions with 2{,}464 annotated rubric items (mean 6.16 items per question), jointly balanced across 9 topics, 11 sectors (merged into 9 leaves), 6 reasoning types, and 12 monthly cutoff-date buckets.
Sector coverage spans healthcare/biotech, technology/semiconductors, consumer discretionary, financial services, real estate, energy/natural resources, industrials/transportation, macro/policy, commodities, crypto/digital assets, and fixed income.
The expert-annotation curation pass (\S\ref{sec:human-curation}) closely preserves the ILP sector balance.

\subsection{Agent Baselines}
\label{sec:baselines}

Following Tongyi-DR's framing~\citep{team2025tongyi}, we group baselines by scaffolding complexity rather than by base-model identity.
The three groups share our PIT-filtered FAISS retriever and \gymname{} as the headline evaluation set:

\noindent\textbf{Fine-tuned open-weight}: open-weight models that internalize the research loop without an external scaffold, fine-tuned end-to-end with RL on agentic-search trajectories (Tongyi-DR~\citep{team2025tongyi}, OpenResearcher~\citep{li2026openresearcher}, MiroThinker~\citep{team2025mirothinker, team2026mirothinker}).

\noindent\textbf{Foundation model with search tool}: a fixed 30-step ReAct wrapper with one search tool, one final-answer tool, and 9 interchangeable backbones, which isolates model capability under a constant scaffold; the result tables split this group into open-weight and proprietary backbones. We also report \workname{}, our open-weight layered harness on a Qwen3.6-27B backbone (Section~\ref{sec:harness}), in this group for the open-weight comparison; Appendix~\ref{sec:appendix-capability-profile} reports a complementary cross-backbone evaluation of \workname{} on a component-coverage suite.

\noindent\textbf{Agentic search systems}: engineered scaffolds that make orchestration decisions in code while using an LLM for planning, retrieval, synthesis, or critique. This group includes TTD-DR~\citep{TTD_DR}, GPT-Researcher,\footnote{\url{https://github.com/assafelovic/gpt-researcher}} STORM~\citep{shao2024assisting}, OpenClaw,\footnote{\url{https://github.com/openclaw/openclaw}} and deepagents.\footnote{\url{https://github.com/hwchase17/deepagents}}
TTD-DR, GPT-Researcher, STORM, OpenClaw, and deepagents all use Gemini-3-Flash as the LLM backbone.

Comparing the three groups on identical questions and the same corpus retriever isolates (a) how well training-time tool distributions transfer to a new search environment (Fine-tuned open-weight vs.\ Agentic search systems); (b) how much hand-engineered scaffolding adds over a minimal ReAct loop on the same backbone (Foundation model with search tool vs.\ Agentic search systems); and (c) how performance varies with model identity at fixed scaffolding (within the foundation-model rows).

The fine-tuned open-weight models were trained against live-web tool stacks (e.g., Serper, Jina); we evaluate them through corpus-backed equivalents (\texttt{Search} $\to$ FAISS+PIT, \texttt{Visit} $\to$ SQLite), a substitution that drives the tool-distribution-shift analysis (\S\ref{sec:tool-shift}).

\subsection{Evaluation Protocol}
\label{sec:protocol}

Agents are run once on the full 500-question ILP-balanced subset; headline metrics then aggregate scores against \gymname{}'s 400 expert-annotated questions and their 2{,}464 annotated rubric items (\S\ref{sec:human-curation}) by subsetting the per-criterion judge outputs, and the broader 500-question run is retained for auxiliary diagnostics.
All baselines run under identical search infrastructure (PIT-filtered FAISS over the same Qwen3-Embedding-4B index).
The foundation-model matrix holds the wrapper fixed across 9 frontier and open-weight backbones.
Each agent receives only the question text and cutoff date; the rubric, thesis, and supporting evidence are withheld for blind evaluation.
The judge (Gemini-3.5-Flash, distinct from the Gemini-3-Flash baseline backbone) scores each rubric criterion on the 5-tier (0--4) scale described in~\S\ref{sec:scoring}, with access to the question, thesis, the rubric item being scored, pre-cutoff edge evidence, post-cutoff edge evidence, and the agent's full report with cited URLs.
Headline scores are reported with bootstrap standard errors ($n{=}1000$).

\section{Results and Analysis}
\label{sec:results}
\label{sec:headline}

We evaluate 23 baselines on \gymname{} under the protocol above.
Scores are averaged with the 5-tier rubric (Eq.~\ref{eq:outcome}); the $\pm$\,SE column reports bootstrap standard errors over questions.
The main text reports the decision-relevant comparisons, while per-axis and harness-ablation breakdowns appear in the appendix.

\paragraph{Cost-quality Frontier.} Fig.~\ref{fig:headline-pareto} shows the performance v.s. cost trade-offs among all benchmarked systems. We observe our \workname{} stays on the frontier with Opus-5 achieves the best overall performance of 44.9\%.

Table~\ref{tab:headline} reports all
evaluated systems grouped by system category; the fixed-backbone \workname{}
ablation is deferred to \S\ref{sec:rl-training}.

\begin{figure*}[tbp]
\centering
\includegraphics[width=0.9\linewidth]{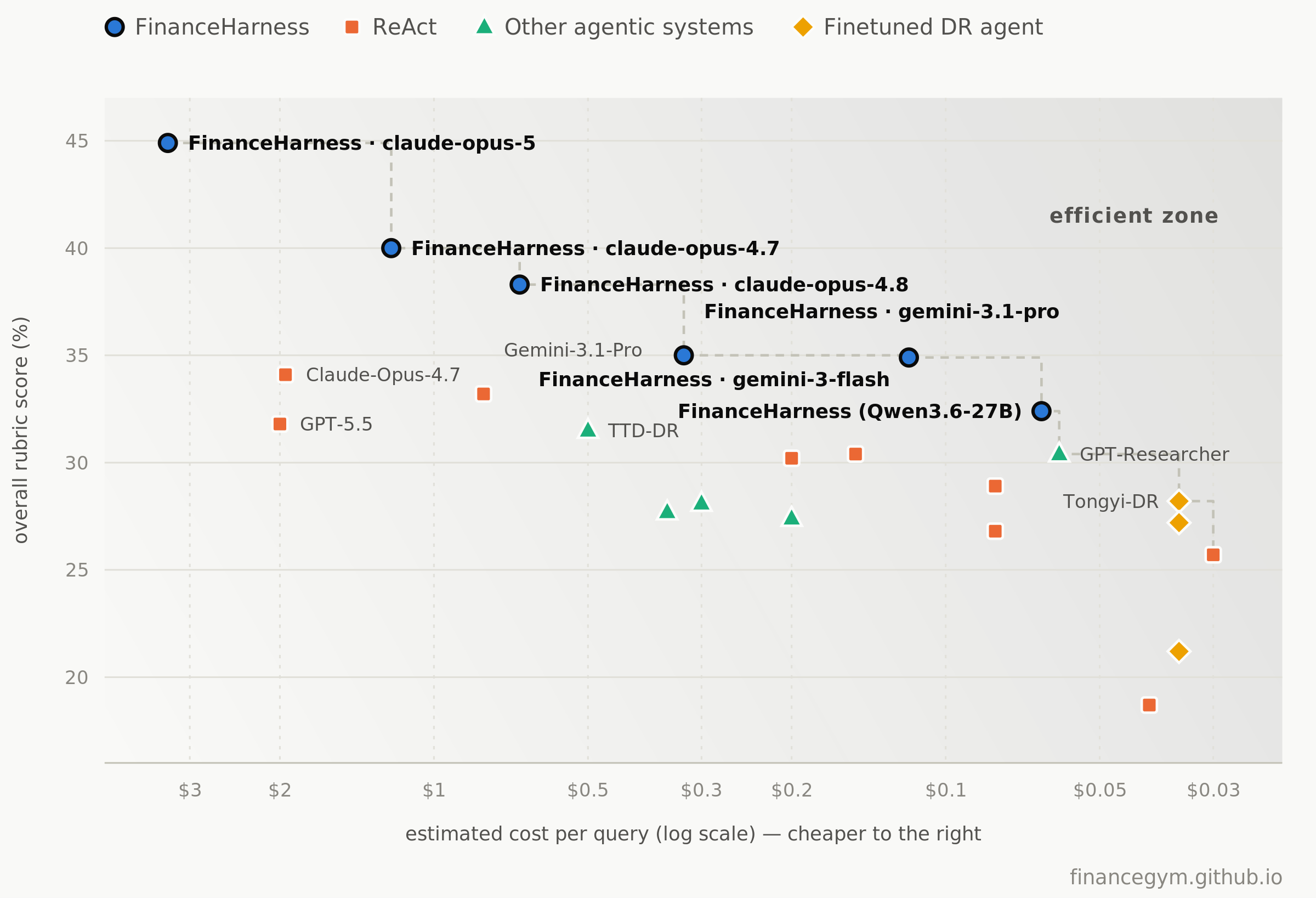}
\caption{Pareto-frontier for \gymname{}. \workname{} reaches an overall frontier, indicating a favorable performance-to-cost trade-off.}
\label{fig:headline-pareto}
\end{figure*}

\begin{table*}[t]
\centering
\small
\setlength{\tabcolsep}{5pt}
\renewcommand{\arraystretch}{1.18}
\caption{Headline results on \gymname{} by system category. Normalized rubric means (\%); within each group, \textbf{bold} marks the best and \underline{underline} the second best per column (Overall/Pre-cutoff/Post-cutoff). $\pm$SE is the bootstrap SE of the Overall score.}
\label{tab:headline}
\resizebox{\textwidth}{!}{%
\begin{tabular}{@{}l@{\hspace{0.04\textwidth}}l@{}}
\begin{tabular}[t]{@{}>{\raggedright\arraybackslash}p{3.35cm} r r r r@{}}
\multicolumn{5}{@{}l}{\textbf{Fine-tuned open-weight}}\\[2pt]
\toprule
\textbf{System} & \textbf{Overall}\,$\uparrow$ & \textbf{Pre-cutoff}\,$\uparrow$ & \textbf{Post-cutoff}\,$\uparrow$ & \textbf{$\pm$SE} \\
\midrule
Tongyi-DR & \textbf{28.2} & \underline{39.7} & \textbf{10.7} & 0.8 \\
OpenResearcher & \underline{27.2} & \textbf{39.9} & \underline{8.1} & 0.7 \\
MiroThinker & 21.2 & 29.6 & 7.8 & 0.6 \\
\bottomrule
\end{tabular} &
\begin{tabular}[t]{@{}>{\raggedright\arraybackslash}p{3.35cm} r r r r@{}}
\multicolumn{5}{@{}l}{\textbf{Agentic search systems}}\\[2pt]
\toprule
\textbf{System} & \textbf{Overall}\,$\uparrow$ & \textbf{Pre-cutoff}\,$\uparrow$ & \textbf{Post-cutoff}\,$\uparrow$ & \textbf{$\pm$SE} \\
\midrule
TTD-DR & \textbf{31.5} & \textbf{45.5} & \underline{9.8} & 0.7 \\
GPT-Researcher & \underline{30.4} & \underline{42.2} & \textbf{12.0} & 0.8 \\
deepagents & 28.1 & 40.7 & 8.6 & 1.1 \\
OpenClaw & 27.7 & 40.2 & 8.3 & 0.7 \\
STORM & 27.4 & 39.3 & 9.4 & 0.6 \\
\bottomrule
\end{tabular} \\[10pt]
\begin{tabular}[b]{@{}>{\raggedright\arraybackslash}p{3.35cm} r r r r@{}}
\multicolumn{5}{@{}l}{\textbf{Open-weight with search}}\\[2pt]
\toprule
\textbf{System} & \textbf{Overall}\,$\uparrow$ & \textbf{Pre-cutoff}\,$\uparrow$ & \textbf{Post-cutoff}\,$\uparrow$ & \textbf{$\pm$SE} \\
\midrule
\textbf{\workname{}} & \textbf{32.4} & \textbf{45.7} & \textbf{11.8} & 0.8 \\
GLM-5 & \underline{30.4} & \underline{44.7} & \underline{8.5} & 0.9 \\
DeepSeek-v3.2 & 28.9 & 42.4 & 8.4 & 0.8 \\
Qwen3-235B-A22B & 26.8 & 39.7 & 7.3 & 0.7 \\
Gemma-4-26B & 25.7 & 37.6 & 7.7 & 0.6 \\
gpt-oss-120b & 18.7 & 27.8 & 5.2 & 0.8 \\
\bottomrule
\end{tabular} &
\begin{tabular}[b]{@{}>{\raggedright\arraybackslash}p{3.35cm} r r r r@{}}
\multicolumn{5}{@{}l}{\textbf{Proprietary with search}}\\[2pt]
\toprule
\textbf{System} & \textbf{Overall}\,$\uparrow$ & \textbf{Pre-cutoff}\,$\uparrow$ & \textbf{Post-cutoff}\,$\uparrow$ & \textbf{$\pm$SE} \\
\midrule
Claude-Opus-4.7 & \textbf{34.1} & \textbf{50.1} & \underline{9.9} & 0.8 \\
Gemini-3.1-Pro & \underline{33.2} & 46.8 & \textbf{12.8} & 0.7 \\
GPT-5.5 & 31.8 & \underline{47.5} & 8.0 & 0.7 \\
Gemini-3-Flash & 30.2 & 43.7 & 9.8 & 0.7 \\
\bottomrule
\end{tabular} \\
\end{tabular}}
\end{table*}

As Table~\ref{tab:headline} shows, on a 27B open-weight backbone, \workname{} ranks among the strongest systems on \gymname{}: it leads all open-weight models, slightly outperforms the best agentic search system, and trails only two proprietary backbones.

\subsection{What Drives Performance}
\label{sec:backbone-vs-scaffold}

The main comparison is between model choice and research procedure under the
same point-in-time corpus.
Changing the backbone within the same search-tool wrapper produces a wider spread
than changing the scaffold around Gemini-3-Flash, where the engineered agentic systems cluster near a minimal ReAct loop and several fall below it.
This suggests that current financial deep-research performance is still
strongly constrained by the underlying model, while task-matched scaffolding can
recover additional evidence and organize it more reliably.

The fine-tuned open-weight models show a related transfer issue: trained around
live-web tool stacks, they are evaluated here through our corpus-backed FAISS/PIT
interface, where several general-purpose search backbones outperform them,
consistent with the tool-distribution-shift analysis in \S\ref{sec:tool-shift}.

The per-topic breakdown (Table~\ref{tab:topic-macro}) and the per-sector and per-reasoning-type breakdowns (Appendix Tables~\ref{tab:sector-macro} and~\ref{tab:reasoning-macro}) follow the aggregate ordering within each system group, with causal and comparative analysis among the hardest reasoning types across systems.

\begin{table*}[t]
\centering
\small
\caption{Per-topic averaged scores (\%) on \gymname{}, by system group. Columns follow the Company/Market/Macro topic levels of Figure~\ref{fig:taxonomy-mix} (Corp.\ strat., Own./gov., Event $\mid$ Analyst, Inst.\ flows, Valu./sent. $\mid$ Policy, Thematic, Alloc.). Each panel is one system group; shading (blue~$=$~higher, amber~$=$~lower), \textbf{bold}~(best) and \underline{underline}~(second) are normalized per column \emph{within the panel}.}
\label{tab:topic-macro}
\textbf{(a) Fine-tuned open-weight}\par\smallskip
\resizebox{\textwidth}{!}{%
\begin{tabular}{l ccccccccc}
\toprule
\textbf{System} & \textbf{Corp.\ strat.} & \textbf{Own./gov.} & \textbf{Event} & \textbf{Analyst} & \textbf{Inst.\ flows} & \textbf{Valu./sent.} & \textbf{Policy} & \textbf{Thematic} & \textbf{Alloc.} \\
\midrule
Tongyi-DR & \cellcolor[HTML]{D6E5FC} \textbf{27.6} & \cellcolor[HTML]{D6E5FC} \textbf{32.5} & \cellcolor[HTML]{E2ECFD} \underline{33.6} & \cellcolor[HTML]{D6E5FC} \textbf{27.5} & \cellcolor[HTML]{D6E5FC} \textbf{20.9} & \cellcolor[HTML]{DDEAFD} \underline{23.8} & \cellcolor[HTML]{D6E5FC} \textbf{29.2} & \cellcolor[HTML]{D6E5FC} \textbf{26.0} & \cellcolor[HTML]{D6E5FC} \textbf{31.1} \\
OpenResearcher & \cellcolor[HTML]{F0F5FE} \underline{25.9} & \cellcolor[HTML]{F8FBFE} \underline{27.8} & \cellcolor[HTML]{D6E5FC} \textbf{35.1} & \cellcolor[HTML]{D8E6FC} \underline{27.3} & \cellcolor[HTML]{FFFEFA} \underline{19.0} & \cellcolor[HTML]{D6E5FC} \textbf{24.0} & \cellcolor[HTML]{DAE7FC} \underline{28.8} & \cellcolor[HTML]{DCE8FD} \underline{25.6} & \cellcolor[HTML]{FEF0CA} 25.5 \\
MiroThinker & \cellcolor[HTML]{FEF0CA} 22.1 & \cellcolor[HTML]{FEF0CA} 21.3 & \cellcolor[HTML]{FEF0CA} 24.1 & \cellcolor[HTML]{FEF0CA} 17.4 & \cellcolor[HTML]{FEF0CA} 17.4 & \cellcolor[HTML]{FEF0CA} 21.6 & \cellcolor[HTML]{FEF0CA} 19.8 & \cellcolor[HTML]{FEF0CA} 20.2 & \cellcolor[HTML]{FEF2D2} \underline{25.9} \\
\bottomrule
\end{tabular}}
\par\medskip
\textbf{(b) Agentic search systems}\par\smallskip
\resizebox{\textwidth}{!}{%
\begin{tabular}{l ccccccccc}
\toprule
\textbf{System} & \textbf{Corp.\ strat.} & \textbf{Own./gov.} & \textbf{Event} & \textbf{Analyst} & \textbf{Inst.\ flows} & \textbf{Valu./sent.} & \textbf{Policy} & \textbf{Thematic} & \textbf{Alloc.} \\
\midrule
TTD-DR & \cellcolor[HTML]{D6E5FC} \textbf{31.8} & \cellcolor[HTML]{DDE9FD} \underline{34.5} & \cellcolor[HTML]{D6E5FC} \textbf{37.2} & \cellcolor[HTML]{F2F6FE} 28.3 & \cellcolor[HTML]{D6E5FC} \textbf{33.2} & \cellcolor[HTML]{D6E5FC} \textbf{28.4} & \cellcolor[HTML]{F6FAFE} \underline{29.6} & \cellcolor[HTML]{EAF1FE} \underline{28.1} & \cellcolor[HTML]{EEF4FE} 26.8 \\
GPT-Researcher & \cellcolor[HTML]{F6FAFE} \underline{29.4} & \cellcolor[HTML]{D6E5FC} \textbf{35.2} & \cellcolor[HTML]{ECF3FE} 36.3 & \cellcolor[HTML]{D6E5FC} \textbf{30.8} & \cellcolor[HTML]{F9FBFE} 28.6 & \cellcolor[HTML]{E6EEFE} \underline{26.9} & \cellcolor[HTML]{FFFCF4} 28.9 & \cellcolor[HTML]{ECF3FE} 28.0 & \cellcolor[HTML]{DEEAFD} \underline{28.0} \\
deepagents & \cellcolor[HTML]{FEF0CE} 25.9 & \cellcolor[HTML]{FEF4D8} 27.4 & \cellcolor[HTML]{EAF1FE} \underline{36.4} & \cellcolor[HTML]{FEF0CA} 23.3 & \cellcolor[HTML]{F1F6FE} \underline{29.7} & \cellcolor[HTML]{FEF0CA} 20.4 & \cellcolor[HTML]{D6E5FC} \textbf{30.9} & \cellcolor[HTML]{FEFBF1} 26.9 & \cellcolor[HTML]{FEF0CA} 22.8 \\
OpenClaw & \cellcolor[HTML]{FEF5DE} 26.8 & \cellcolor[HTML]{FEF9EB} 29.0 & \cellcolor[HTML]{FEF0CA} 33.8 & \cellcolor[HTML]{FEF8E5} 25.2 & \cellcolor[HTML]{FEF0CA} 22.4 & \cellcolor[HTML]{FBFCFE} 24.8 & \cellcolor[HTML]{FEF8E7} 28.5 & \cellcolor[HTML]{FEF0CA} 25.8 & \cellcolor[HTML]{D6E5FC} \textbf{28.5} \\
STORM & \cellcolor[HTML]{FEF0CA} 25.7 & \cellcolor[HTML]{FEF0CA} 26.2 & \cellcolor[HTML]{FEF4DA} 34.3 & \cellcolor[HTML]{DAE7FC} \underline{30.5} & \cellcolor[HTML]{FEF0CC} 22.6 & \cellcolor[HTML]{FFFCF7} 23.8 & \cellcolor[HTML]{FEF0CA} 27.6 & \cellcolor[HTML]{D6E5FC} \textbf{28.8} & \cellcolor[HTML]{F4F8FE} 26.4 \\
\bottomrule
\end{tabular}}
\par\medskip
\textbf{(c) Proprietary with search}\par\smallskip
\resizebox{\textwidth}{!}{%
\begin{tabular}{l ccccccccc}
\toprule
\textbf{System} & \textbf{Corp.\ strat.} & \textbf{Own./gov.} & \textbf{Event} & \textbf{Analyst} & \textbf{Inst.\ flows} & \textbf{Valu./sent.} & \textbf{Policy} & \textbf{Thematic} & \textbf{Alloc.} \\
\midrule
Claude-Opus-4.7 & \cellcolor[HTML]{D6E5FC} \textbf{33.9} & \cellcolor[HTML]{D6E5FC} \textbf{34.6} & \cellcolor[HTML]{D6E5FC} \textbf{40.2} & \cellcolor[HTML]{E4EEFE} \underline{30.4} & \cellcolor[HTML]{FEF2D2} \underline{22.7} & \cellcolor[HTML]{D9E6FC} \underline{28.4} & \cellcolor[HTML]{D6E5FC} \textbf{38.7} & \cellcolor[HTML]{D6E5FC} \textbf{32.9} & \cellcolor[HTML]{D6E5FC} \textbf{35.3} \\
Gemini-3.1-Pro & \cellcolor[HTML]{EAF2FE} \underline{33.4} & \cellcolor[HTML]{D8E6FC} \underline{34.5} & \cellcolor[HTML]{F6F9FE} \underline{38.6} & \cellcolor[HTML]{D6E5FC} \textbf{30.8} & \cellcolor[HTML]{D6E5FC} \textbf{25.3} & \cellcolor[HTML]{D6E5FC} \textbf{28.5} & \cellcolor[HTML]{F7FAFE} 35.4 & \cellcolor[HTML]{E0ECFD} \underline{32.3} & \cellcolor[HTML]{FFFEFA} \underline{32.2} \\
GPT-5.5 & \cellcolor[HTML]{FEF7E4} 32.3 & \cellcolor[HTML]{F2F7FE} 33.3 & \cellcolor[HTML]{FEF3D7} 36.5 & \cellcolor[HTML]{F0F6FE} 30.0 & \cellcolor[HTML]{FEF0CA} 22.5 & \cellcolor[HTML]{FEF0CA} 25.5 & \cellcolor[HTML]{F2F7FE} \underline{35.9} & \cellcolor[HTML]{FEFEFF} 30.5 & \cellcolor[HTML]{FEF0CA} 29.7 \\
Gemini-3-Flash & \cellcolor[HTML]{FEF0CA} 31.8 & \cellcolor[HTML]{FEF0CA} 30.8 & \cellcolor[HTML]{FEF0CA} 36.0 & \cellcolor[HTML]{FEF0CA} 28.3 & \cellcolor[HTML]{FEF0CA} 22.5 & \cellcolor[HTML]{FEF6E0} 26.1 & \cellcolor[HTML]{FEF0CA} 30.5 & \cellcolor[HTML]{FEF0CA} 28.0 & \cellcolor[HTML]{FFFCF6} 32.0 \\
\bottomrule
\end{tabular}}
\par\medskip
\textbf{(d) Open-weight with search}\par\smallskip
\resizebox{\textwidth}{!}{%
\begin{tabular}{l ccccccccc}
\toprule
\textbf{System} & \textbf{Corp.\ strat.} & \textbf{Own./gov.} & \textbf{Event} & \textbf{Analyst} & \textbf{Inst.\ flows} & \textbf{Valu./sent.} & \textbf{Policy} & \textbf{Thematic} & \textbf{Alloc.} \\
\midrule
\textbf{FinanceHarness} & \cellcolor[HTML]{D6E5FC} \textbf{31.7} & \cellcolor[HTML]{D6E5FC} \textbf{34.7} & \cellcolor[HTML]{D6E5FC} \textbf{39.1} & \cellcolor[HTML]{D6E5FC} \textbf{33.4} & \cellcolor[HTML]{D6E5FC} \textbf{25.9} & \cellcolor[HTML]{D6E5FC} \textbf{28.6} & \cellcolor[HTML]{D6E5FC} \textbf{33.6} & \cellcolor[HTML]{D6E5FC} \textbf{29.6} & \cellcolor[HTML]{D6E5FC} \textbf{29.6} \\
GLM-5 & \cellcolor[HTML]{E0EBFD} \underline{30.2} & \cellcolor[HTML]{DDE9FD} \underline{33.5} & \cellcolor[HTML]{D8E6FC} \underline{38.6} & \cellcolor[HTML]{FFFEFC} 28.7 & \cellcolor[HTML]{FFFDF8} 19.9 & \cellcolor[HTML]{F4F8FE} 23.9 & \cellcolor[HTML]{DCE8FD} \underline{32.7} & \cellcolor[HTML]{DCE8FD} \underline{28.8} & \cellcolor[HTML]{ECF2FE} 26.7 \\
DeepSeek-v3.2 & \cellcolor[HTML]{F2F7FE} 27.2 & \cellcolor[HTML]{EBF2FE} 30.9 & \cellcolor[HTML]{E6EFFE} 35.6 & \cellcolor[HTML]{F8FAFE} \underline{29.7} & \cellcolor[HTML]{FFFCF4} 19.5 & \cellcolor[HTML]{E5EEFE} \underline{26.3} & \cellcolor[HTML]{DCE9FD} 32.5 & \cellcolor[HTML]{F0F5FE} 26.0 & \cellcolor[HTML]{FDFEFF} 24.4 \\
Qwen3-235B-A22B & \cellcolor[HTML]{FCFEFF} 25.6 & \cellcolor[HTML]{FEFFFF} 27.2 & \cellcolor[HTML]{ECF2FE} 34.4 & \cellcolor[HTML]{FEFAEC} 27.3 & \cellcolor[HTML]{FEF8E7} 18.1 & \cellcolor[HTML]{F5F8FE} 23.7 & \cellcolor[HTML]{F4F8FE} 28.3 & \cellcolor[HTML]{F4F8FE} 25.3 & \cellcolor[HTML]{F4F8FE} 25.5 \\
Gemma-4-26B & \cellcolor[HTML]{FFFCF4} 23.8 & \cellcolor[HTML]{FFFEFB} 26.6 & \cellcolor[HTML]{FAFCFE} 31.2 & \cellcolor[HTML]{FEF0CA} 24.4 & \cellcolor[HTML]{FFFEFC} \underline{20.2} & \cellcolor[HTML]{FAFCFE} 22.9 & \cellcolor[HTML]{FAFCFE} 27.2 & \cellcolor[HTML]{FBFCFE} 24.4 & \cellcolor[HTML]{DBE8FD} \underline{29.0} \\
gpt-oss-120b & \cellcolor[HTML]{FEF0CA} 18.7 & \cellcolor[HTML]{FEF0CA} 19.6 & \cellcolor[HTML]{FEF0CA} 21.0 & \cellcolor[HTML]{FEF1D0} 24.9 & \cellcolor[HTML]{FEF0CA} 15.2 & \cellcolor[HTML]{FEF0CA} 15.6 & \cellcolor[HTML]{FEF0CA} 19.2 & \cellcolor[HTML]{FEF0CA} 18.0 & \cellcolor[HTML]{FEF0CA} 18.6 \\
\bottomrule
\end{tabular}}
\end{table*}

\begin{figure*}[tbp]
\centering
\includegraphics[width=0.8\linewidth]{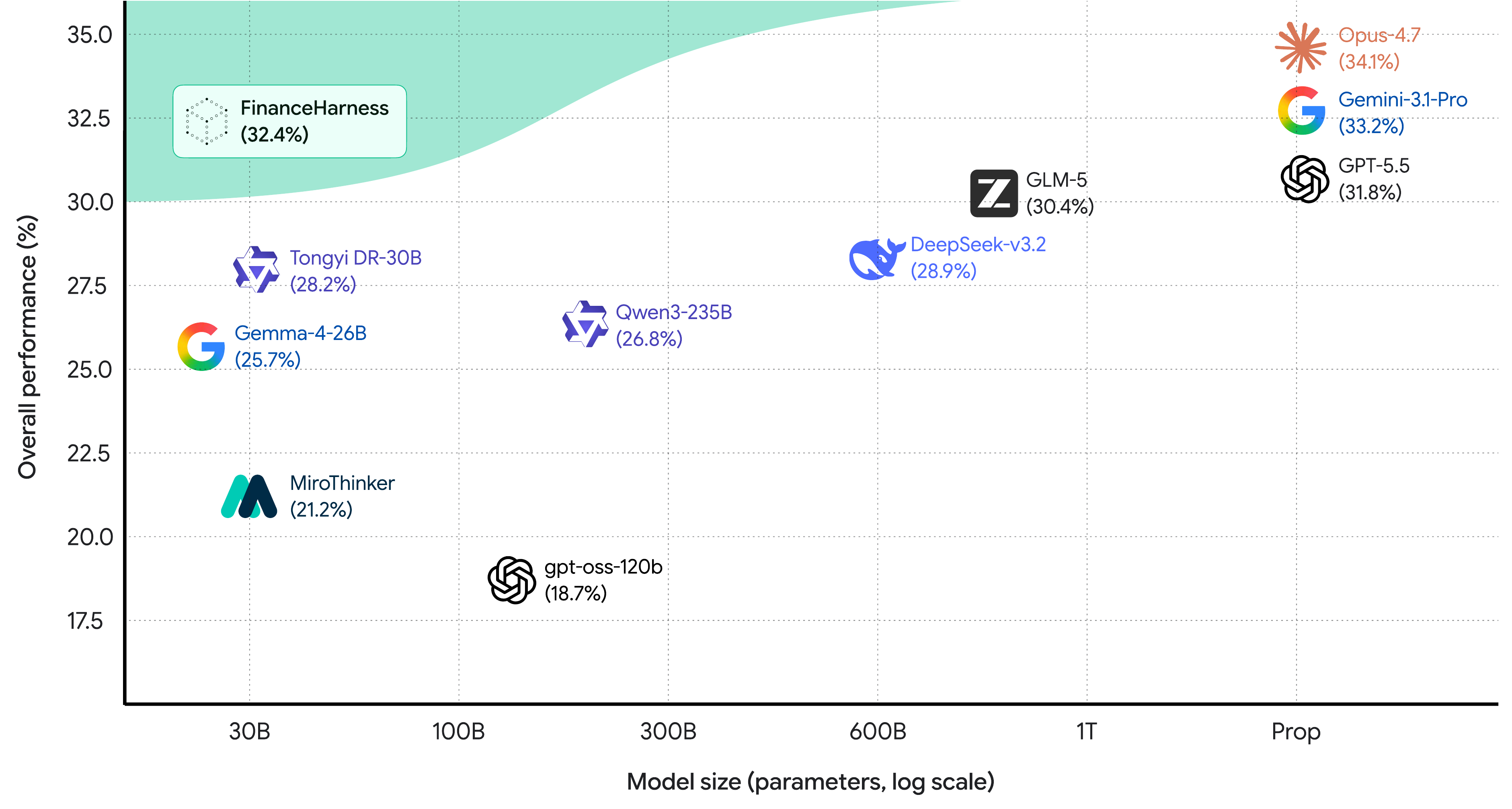}
\caption{Overall outcome score versus backbone scale on \gymname{}.
Each marker is one evaluated system, placed by backbone parameter count
(x-axis, log scale; proprietary models grouped at right) and overall rubric
score (y-axis). On a 27B open-weight backbone, \workname{} reaches an overall
score competitive with much larger open-weight models and approaching the
proprietary frontier (shaded), indicating a favorable performance-to-scale
trade-off.}
\label{fig:headline-scatter}
\end{figure*}

Overall score also tracks backbone scale, but only loosely.
Plotting each system's score against its backbone size (Figure~\ref{fig:headline-scatter}), \workname{} sits on the upper-left efficiency frontier, matching much larger open-weight backbones and approaching the proprietary frontier despite its 27B backbone.

Finally, the pre-cutoff/post-cutoff split is stable across systems: across all
baselines, pre-cutoff scores are several times higher than post-cutoff scores.
Because the same PIT-filtered corpus is used for every agent, this gap reflects the benchmark's core difficulty rather than the retrieval setup.
Even the leading agents leave most forward-looking rubric items only partially addressed or missed.

\subsection{In-Environment Training}
\label{sec:rl-training}

The same rubric judge can also serve as a training signal. Using the same
retrieval tools and report format at rollout and evaluation, we further train
\workname{} with Group Relative Policy Optimization (GRPO) on a separate set of 172 machine-curated training instances
that we build with the same environment and graph-construction harness used to
construct the benchmark, with no human involvement and independent of \gymname{},
so that training and evaluation remain fully separate.
The reward combines rubric-style evaluation against generated criteria
($0.6$ weight) with an LLM judge over report coherence, grounding, and trajectory
quality ($0.4$ weight).
We then evaluate the resulting policy on \gymname{}.

\begin{table}[!ht]
\centering
\small
\caption{Fixed-backbone (Qwen3.6-27B) ablation on \gymname{}: the model with
search only (Vanilla), a naive harness (Naive), the full untrained \workname{}
(\textbf{Harness}, our main configuration), and \workname{} after GRPO training
on a separate, machine-curated training set independent of \gymname{}, with a
mixed rubric/report-and-trace reward (RFT).
Normalized rubric means (\%).}
\label{tab:harness-ablation}
\begin{tabular}{@{}l r r r@{}}
\toprule
\textbf{Configuration} & \textbf{Total} & \textbf{Pre-cutoff} & \textbf{Post-cutoff} \\
\midrule
Vanilla (model with search) & 25.3 & 36.1 & 8.7 \\
Naive harness & 29.6 & 41.8 & 10.7 \\
\textbf{Harness} & \textbf{32.4} & \textbf{45.7} & \textbf{11.8} \\
\quad + RFT (GRPO training) & 32.8 & 46.2 & 12.1 \\
\bottomrule
\end{tabular}
\end{table}
Table~\ref{tab:harness-ablation} reports four fixed-backbone configurations: the
search-only model (Vanilla), a naive harness (Naive), the full untrained harness
(Harness, our main configuration), and the harness after reinforcement fine-tuning with GRPO (RFT).
GRPO training adds only $0.4$ point over the untrained harness, so we treat it as
a refinement rather than a headline result and leave a broader trained-policy
sweep to future work.

% \FloatBarrier
\section{Conclusion}
\label{sec:conclusion}

We presented \gymname{}, a point-in-time financial deep research benchmark built
from 400 expert-annotated questions and 2{,}464 rubric items over a reproducible
PIT search sandbox, and \workname{}, an expert-knowledge-guided harness that runs
and optimizes agents within the same environment. The benchmark pairs the sandbox
with pre-cutoff synthesis and post-cutoff reasoning rubrics, allowing evaluation to separate pre-cutoff-date evidence retrieval from post-cutoff-date outcome anticipation. Across 17 baselines and \workname{}, every system stays below 40\% overall and shows a
persistent pre-/post-cutoff gap, suggesting that financial deep research is not addressed
by stronger retrieval alone. The fixed-backbone harness ablation further shows progressive
improvement from a Qwen3.6-27B search-tool model (25.3\%) to the full \workname{}
(32.4\%), with subsequent GRPO training adding a further 0.4 point,
supporting the paper's core claim: evaluation, tool use, and optimization benefit
from sharing the same temporally controlled financial research environment.

\section*{Limitations}

\paragraph{Corpus scope.}
The point-in-time search sandbox is built from a 2025 English-language web corpus
that we collect ourselves, so coverage of non-English sources, paywalled
professional data, and structured filings is limited.

\paragraph{Tool-distribution shift.}
Specialized deep-research models are evaluated out-of-distribution on our
corpus-backed retriever rather than the live-web stacks they were trained
against; their scores should therefore be read as transfer results under a
controlled PIT interface.

\bibliography{reference}

\clearpage
\appendix

\definecolor{tagblue}{RGB}{100,150,200}
\definecolor{evidenceblue}{RGB}{233,242,255}
\providecommand{\hlevidence}[1]{{\sethlcolor{evidenceblue}\hl{#1}}}

\definecolor{caseLink}{HTML}{445A67}
\newcommand{\fhlink}[2]{{\hypersetup{urlcolor=caseLink}\href{#1}{#2}}}

\definecolor{caseBadge}{HTML}{536B79}
\definecolor{caseTool}{HTML}{80645B}
\definecolor{caseGround}{HTML}{F3F1EC}
\definecolor{caseRule}{HTML}{B8B3AA}

\definecolor{caseRepFrame}{HTML}{AE9B84}
\definecolor{caseRepBack}{HTML}{FBFAF6}
\definecolor{caseTrFrame}{HTML}{8FA1AD}
\definecolor{caseTrBack}{HTML}{F4F7F8}
\definecolor{caseRespBack}{HTML}{E8ECEF}

\newtcbox{\rbadge}{on line, boxrule=0pt, boxsep=0pt,
  left=3.2pt, right=3.2pt, top=1.3pt, bottom=1.3pt, arc=2.2pt,
  colback=caseBadge, coltext=white, fontupper=\scriptsize\bfseries\sffamily}

\newtcolorbox{fhresp}{%
  breakable, enhanced, arc=0.6mm, boxrule=0pt, frame hidden,
  colback=caseRespBack,
  left=2mm, right=1.5mm, top=0.6mm, bottom=0.6mm, boxsep=0.5mm,
  before skip=0.3mm, after skip=0mm, fontupper=\footnotesize}

\section{\workname{} Ablation Breakdowns}
\label{sec:appendix-harness-ablation}

We expand Table~\ref{tab:harness-ablation} by topic, reasoning
type, sector, and situation type.
All rows use the same Qwen3.6-27B backbone and the same \gymname{} evaluation
protocol.

Scores rise across the four configurations (Vanilla, Naive, Harness, RFT). The
harness configurations account for nearly all of the gain, while subsequent GRPO
training (RFT) adds only a small final increment. The hardest cells stay hard:
crypto and macro/rates remain the weakest sectors, and causal and comparative
analysis the weakest reasoning types.

\begin{table*}[htbp]
\centering
\small
\caption{\workname{} ablation by topic. Scores are normalized rubric means
(\%).}
\label{tab:harness-ablation-topic}
\begin{tabular}{@{}l r r r r@{}}
\toprule
\textbf{Topic} & \textbf{Vanilla} & \textbf{Naive} & \textbf{Harness} & \textbf{RFT} \\
\midrule
Corporate strategy competitive positioning & 25.3 & 29.4 & 31.7 & 32.3 \\
Macro regulatory policy & 25.6 & 30.5 & 33.6 & 34.1 \\
Event driven restructuring & 30.9 & 36.4 & 39.1 & 39.3 \\
Valuation sentiment divergence & 21.5 & 25.1 & 28.6 & 29.0 \\
Sector thematic analysis & 23.3 & 27.1 & 29.6 & 30.1 \\
Ownership governance stakeholder intelligence & 28.4 & 31.2 & 34.7 & 35.0 \\
Institutional flows market microstructure & 20.2 & 23.4 & 25.9 & 26.1 \\
Analyst research consensus dynamics & 25.8 & 31.7 & 33.4 & 33.7 \\
Asset allocation portfolio strategy & 21.4 & 27.1 & 29.6 & 29.8 \\
\bottomrule
\end{tabular}
\end{table*}

\begin{table*}[htbp]
\centering
\small
\caption{\workname{} ablation by reasoning type. Scores are normalized rubric
means (\%).}
\label{tab:harness-ablation-reasoning}
\begin{tabular}{@{}l r r r r@{}}
\toprule
\textbf{Reasoning type} & \textbf{Vanilla} & \textbf{Naive} & \textbf{Harness} & \textbf{RFT} \\
\midrule
Causal analysis & 23.4 & 27.3 & 29.8 & 30.1 \\
Predictive forecasting & 25.4 & 29.1 & 32.2 & 32.8 \\
Effectiveness evaluation & 25.2 & 30.9 & 32.9 & 33.2 \\
Risk assessment & 31.4 & 35.3 & 39.5 & 39.9 \\
Comparative analysis & 23.2 & 26.8 & 29.7 & 30.2 \\
Quantitative analysis & 23.5 & 29.2 & 31.8 & 32.3 \\
\bottomrule
\end{tabular}
\end{table*}

\begin{table*}[htbp]
\centering
\small
\caption{\workname{} ablation by sector. Scores are normalized rubric means
(\%).}
\label{tab:harness-ablation-sector}
\begin{tabular}{@{}l r r r r@{}}
\toprule
\textbf{Sector} & \textbf{Vanilla} & \textbf{Naive} & \textbf{Harness} & \textbf{RFT} \\
\midrule
Healthcare biotech & 32.8 & 36.1 & 39.8 & 40.1 \\
Technology semiconductors & 24.7 & 29.3 & 31.8 & 32.1 \\
Consumer discretionary & 25.0 & 29.6 & 31.4 & 31.8 \\
Industrials transportation & 23.9 & 30.6 & 32.6 & 32.6 \\
Energy \& comm. & 24.4 & 27.5 & 31.0 & 31.6 \\
Real estate & 24.2 & 29.7 & 32.3 & 32.9 \\
Financial services & 24.5 & 30.3 & 32.8 & 33.1 \\
Macro \& rates & 20.3 & 23.8 & 26.9 & 27.6 \\
Crypto digital assets & 17.8 & 21.7 & 24.0 & 25.1 \\
\bottomrule
\end{tabular}
\end{table*}

\begin{table*}[htbp]
\centering
\small
\caption{\workname{} ablation by situation type. Scores are normalized rubric
means (\%).}
\label{tab:harness-ablation-situation}
\begin{tabular}{@{}l r r r r@{}}
\toprule
\textbf{Situation type} & \textbf{Vanilla} & \textbf{Naive} & \textbf{Harness} & \textbf{RFT} \\
\midrule
Multihop path & 26.4 & 30.9 & 33.7 & 34.2 \\
Temporal narrative & 23.8 & 27.8 & 30.7 & 31.0 \\
Tension analyst disagreement & 24.0 & 27.9 & 30.9 & 31.6 \\
Tension price target divergence & 21.3 & 25.7 & 28.2 & 28.2 \\
Tension performance divergence & 20.9 & 25.4 & 27.2 & 27.7 \\
Tension strategic portfolio shift & 31.8 & 35.4 & 39.1 & 39.7 \\
Tension ownership churn & 32.1 & 37.6 & 40.1 & 40.6 \\
\bottomrule
\end{tabular}
\end{table*}

\FloatBarrier
\section{Cross-Backbone Harness Evaluation}
\label{sec:appendix-capability-profile}

Beyond the end-to-end \gymname{} rubric, we evaluate \workname{} as a deployed
system, a model-agnostic harness run unchanged across three backbones: the
open-weight Qwen3.6-27B (our default), Gemini-3.5-Flash, and GPT-5.5. A coverage
suite of 11 tasks, each designed to exercise one component category (valuation,
risk, relative valuation, market data, focused and broad web research,
auto-routing, three skills, and planning), is run at $N{=}3$ per cell and scored
by a dual-LLM judge panel (Gemini-3.5-Flash and GPT-5.5) on faithfulness,
grounding, coherence, and completeness (1--5); grounding is credited to either a
cited source or the structured tool data the agent fetched or computed. Whereas
\gymname{} measures end-to-end report quality under a point-in-time rubric, this
evaluation isolates whether the harness invokes and grounds each component
correctly across backbones.

\begin{table}[t]
\centering
\small
\caption{\workname{} across three backbones on the coverage suite
(11 tasks $\times$ $N{=}3$), dual-LLM judged on a 1--5 scale.}
\label{tab:coverage-backbone}
\begin{tabular}{@{}l c c c c@{}}
\toprule
\textbf{Backbone} & \textbf{Faithfulness} & \textbf{Grounding} & \textbf{Coherence} & \textbf{Completeness} \\
\midrule
GPT-5.5                & 4.36 & 4.47 & 4.92 & 4.98 \\
Qwen3.6-27B (open-weight)      & 4.02 & 4.20 & 4.68 & 4.76 \\
Gemini-3.5-Flash       & 3.85 & 3.97 & 4.78 & 5.00 \\
\midrule
Overall                & 4.08 & 4.21 & 4.80 & 4.91 \\
\bottomrule
\end{tabular}
\end{table}

On the same harness, tools, and judges, the open-weight Qwen3.6-27B backbone
scores within about $0.3$ of GPT-5.5 on faithfulness and grounding and ahead of
Gemini-3.5-Flash (Table~\ref{tab:coverage-backbone}): with \workname{}, a compact
open-weight backbone performs comparably to much larger models in a
parameter-efficient manner. Coherence and completeness saturate ($\geq 4.68$ for
every backbone), so faithfulness and grounding are the discriminating axes.
Behaviorally the harness routes reliably: data routing reaches 95.8\% and
golden-answer computation 90.9\% (rule-scored exact rates), with a tool-coverage rate
of 0.99, a 0.94 skill-hit rate, and zero unknown-tool errors across runs. All
three backbones also pass the capability checks: a multi-turn follow-up that
uses prior-turn context, a \texttt{/compact} summarization that preserves
dollar figures, and correct flagging of an under-specified question for
clarification.

Table~\ref{tab:coverage-category} breaks faithfulness and grounding down by
component category (across backbones). Risk and market tasks are strongest:
value-at-risk, beta, and correlation, and rates and indices, compose cleanly and
stay grounded. Relative valuation is weakest because the comparables tool
returns a small peer set, so a weaker backbone pads the peer table from prior
knowledge and the judges penalize the ungrounded cells.

\begin{table}[t]
\centering
\small
\caption{Coverage suite by component category (faithfulness / grounding, 1--5,
across backbones).}
\label{tab:coverage-category}
\begin{tabular}{@{}l cccccccc@{}}
\toprule
\textbf{Metric} & \textbf{Risk} & \textbf{Market} & \textbf{Research} & \textbf{Auto-routing} & \textbf{Skill} & \textbf{Valuation} & \textbf{Planning} & \textbf{Relative} \\
\midrule
Faithfulness & 4.53 & 4.39 & 4.24 & 4.22 & 4.15 & 3.89 & 3.67 & 3.25 \\
Grounding    & 4.61 & 4.56 & 4.35 & 4.44 & 4.26 & 3.89 & 4.05 & 3.33 \\
\bottomrule
\end{tabular}
\end{table}

\FloatBarrier
\section{Additional Analysis}
\label{sec:analysis}

\subsection{Tool-Distribution Shift}
\label{sec:tool-shift}

The fine-tuned open-weight deep-research models underperform the strongest agentic search systems on
\gymname{} despite being trained for agentic search.
We interpret this as a tool-distribution-shift result rather than a general
model-capability result.
Tongyi-DR and MiroThinker were optimized around live-web search and page
extraction conventions; in our evaluation, they must operate through a
PIT-filtered corpus retriever with fixed document records and no live-web
access.
That substitution changes both the interaction pattern and the final-report
format expected by the rubric.

The most visible consequence is attribution.
\gymname{}'s score-4 anchor rewards claims that are correct, specific, and
grounded in plausible source evidence.
Agentic search systems such as TTD-DR and GPT-Researcher are explicitly prompted to produce
source-attributed reports, while fine-tuned open-weight systems often emit cleaner
answer-style prose with weaker inline attribution.
This does not mean the trained models lack the underlying knowledge or search
ability; it means their learned reporting interface is mismatched with a
long-form, citation-sensitive financial rubric.

\subsection{Pre-cutoff and Post-cutoff Remain Different Failure Modes}
\label{sec:analysis-gap}

The stable gap between pre-cutoff and post-cutoff rubric scores is the most important
behavioral pattern in the benchmark.
Systems improve much more readily on pre-cutoff rubrics, where better search,
reading, and citation discipline help recover pre-cutoff-date evidence.
Post-cutoff rubrics remain compressed even for stronger systems, suggesting that
financial deep research is not solved by retrieval quality alone: agents must
turn historical evidence into forward-looking judgments that are only
verifiable after the cutoff date.
This is why the fixed-backbone \workname{} ablation is informative: even after
training, post-cutoff performance remains well below pre-cutoff performance (about 12\% versus 46\%),
indicating that forward-looking judgment, rather than evidence recall, is the
harder part of the task.

\subsection{Agentic-system Trade-offs}
\label{sec:arch-diversity}

Agentic-system choice changes more than the final score.
Among the agentic search systems, iterative pipelines such as TTD-DR achieve the
highest scores, but they repeatedly plan, retrieve, draft, critique, and revise.
GPT-Researcher is slightly weaker on aggregate quality but runs a lighter, less
iterative pipeline.
Other agentic systems such as STORM and deepagents are useful counterpoints: they
bring different composition strategies, but their default output styles are not
always aligned with a rubric that rewards specific, source-grounded financial
claims.
We therefore treat these comparisons as evidence about orchestration
fit-to-task, not as a universal ranking of agentic systems.

\subsection{Per-Axis Breakdowns}
\label{sec:appendix-peraxis}

Tables~\ref{tab:sector-macro} and~\ref{tab:reasoning-macro} give the per-sector and per-reasoning-type scores by system group, complementing the per-topic breakdown in the main text (Table~\ref{tab:topic-macro}).

\begin{table*}[!b]
\centering
\small
\caption{Per-sector averaged scores (\%) on \gymname{}, by system group. The 11 raw sectors merge into 9 leaves (Energy $=$ energy $+$ commodities; Macro $=$ macro/policy $+$ fixed income), ordered by the four coverage domains of Figure~\ref{fig:taxonomy-mix} (Health, Tech $\mid$ Cons., Indus. $\mid$ Energy, RE $\mid$ Fin., Macro, Crypto). Each panel is one system group; \textbf{Bold}~(best) and \underline{underline}~(second) are normalized per column \emph{within the panel}.}
\label{tab:sector-macro}
\textbf{(a) Fine-tuned open-weight}\par\smallskip
\begin{tabular}{l ccccccccc}
\toprule
\textbf{System} & \textbf{Health} & \textbf{Tech} & \textbf{Cons.} & \textbf{Indus.} & \textbf{Energy} & \textbf{RE} & \textbf{Fin.} & \textbf{Macro} & \textbf{Crypto} \\
\midrule
Tongyi-DR & \textbf{33.5} & \textbf{29.5} & \textbf{28.0} & \textbf{29.4} & \underline{25.2} & \underline{25.0} & \underline{25.6} & \underline{25.2} & \textbf{24.7} \\
OpenResearcher & \underline{31.2} & \underline{26.2} & \underline{27.2} & \underline{27.6} & \textbf{25.3} & \textbf{27.9} & \textbf{28.9} & \textbf{25.7} & \underline{18.6} \\
MiroThinker & 21.9 & 22.1 & 25.0 & 20.2 & 18.6 & 19.9 & 24.1 & 18.2 & 18.1 \\
\bottomrule
\end{tabular}
\par\medskip
\textbf{(b) Agentic search systems}\par\smallskip
\begin{tabular}{l ccccccccc}
\toprule
\textbf{System} & \textbf{Health} & \textbf{Tech} & \textbf{Cons.} & \textbf{Indus.} & \textbf{Energy} & \textbf{RE} & \textbf{Fin.} & \textbf{Macro} & \textbf{Crypto} \\
\midrule
TTD-DR & \textbf{38.2} & \underline{31.5} & \textbf{31.3} & \textbf{30.2} & \underline{28.9} & \textbf{30.9} & \textbf{36.7} & \textbf{26.3} & \underline{18.6} \\
GPT-Researcher & \underline{37.7} & 31.1 & \underline{30.3} & \underline{29.4} & \textbf{29.3} & \underline{28.3} & 29.4 & \underline{25.4} & 17.5 \\
deepagents & 28.6 & \textbf{32.3} & 28.8 & 26.0 & 28.6 & 21.6 & \underline{31.3} & 24.9 & \textbf{19.2} \\
OpenClaw & 31.4 & 30.9 & 27.4 & 25.6 & 26.1 & 22.7 & 30.5 & 24.5 & 17.7 \\
STORM & 31.0 & 27.9 & 28.3 & 27.1 & 26.1 & 26.8 & 29.4 & 23.7 & \textbf{19.2} \\
\bottomrule
\end{tabular}
\par\medskip
\textbf{(c) Proprietary with search}\par\smallskip
\begin{tabular}{l ccccccccc}
\toprule
\textbf{System} & \textbf{Health} & \textbf{Tech} & \textbf{Cons.} & \textbf{Indus.} & \textbf{Energy} & \textbf{RE} & \textbf{Fin.} & \textbf{Macro} & \textbf{Crypto} \\
\midrule
Claude-Opus-4.7 & \textbf{38.7} & \underline{33.5} & \textbf{35.5} & \textbf{33.9} & \textbf{32.3} & \textbf{32.4} & \underline{32.7} & \textbf{33.2} & 22.6 \\
Gemini-3.1-Pro & \underline{36.9} & \textbf{35.5} & \underline{33.3} & 32.6 & \underline{31.4} & 27.7 & \textbf{34.0} & \underline{31.0} & \underline{23.8} \\
GPT-5.5 & 36.1 & 31.8 & 32.8 & 30.9 & 30.4 & \underline{28.6} & 29.4 & 30.4 & \textbf{25.9} \\
Gemini-3-Flash & 34.1 & 30.4 & 29.6 & \underline{33.0} & 27.0 & 25.6 & 31.9 & 29.2 & 19.7 \\
\bottomrule
\end{tabular}
\par\medskip
\textbf{(d) Open-weight with search}\par\smallskip
\begin{tabular}{l ccccccccc}
\toprule
\textbf{System} & \textbf{Health} & \textbf{Tech} & \textbf{Cons.} & \textbf{Indus.} & \textbf{Energy} & \textbf{RE} & \textbf{Fin.} & \textbf{Macro} & \textbf{Crypto} \\
\midrule
\textbf{FinanceHarness} & \textbf{39.8} & \textbf{31.8} & \underline{31.4} & \textbf{32.6} & \textbf{31.0} & \textbf{32.3} & \textbf{32.8} & \textbf{26.9} & \textbf{24.0} \\
GLM-5 & \underline{36.1} & \underline{29.6} & \textbf{33.3} & \underline{29.1} & \underline{30.5} & 26.2 & \underline{30.0} & 24.5 & 18.2 \\
DeepSeek-v3.2 & 35.2 & 27.2 & 30.7 & 27.5 & 27.3 & 21.9 & 29.4 & \textbf{26.9} & \underline{19.2} \\
Qwen3-235B-A22B & 30.7 & 25.9 & 26.9 & 27.3 & 24.7 & \underline{27.4} & 28.4 & 25.3 & 18.4 \\
Gemma-4-26B & 28.9 & 26.5 & 26.1 & 26.0 & 24.3 & 24.8 & 24.6 & \underline{25.4} & 11.8 \\
gpt-oss-120b & 19.9 & 19.4 & 17.2 & 17.1 & 20.0 & 18.5 & 22.1 & 16.7 & 15.5 \\
\bottomrule
\end{tabular}
\end{table*}

\begin{table*}[!b]
\centering
\small
\caption{Per-reasoning-type averaged scores (\%) on \gymname{}, by system group. Columns follow the three analytics levels of Figure~\ref{fig:taxonomy-mix} (Causal, Compar.\ $\mid$ Forecast, Risk $\mid$ Effect., Quant.). Each panel is one system group; \textbf{Bold}~(best) and \underline{underline}~(second) are normalized per column \emph{within the panel}.}
\label{tab:reasoning-macro}
\textbf{(a) Fine-tuned open-weight}\par\smallskip
\begin{tabular}{l cccccc}
\toprule
\textbf{System} & \textbf{Causal} & \textbf{Compar.} & \textbf{Forecast} & \textbf{Risk} & \textbf{Effect.} & \textbf{Quant.} \\
\midrule
Tongyi-DR & \textbf{24.6} & \textbf{27.2} & \underline{28.0} & \textbf{34.9} & \textbf{29.9} & \textbf{27.2} \\
OpenResearcher & \underline{24.0} & \underline{24.2} & \textbf{28.4} & \underline{31.8} & \underline{29.2} & \underline{26.0} \\
MiroThinker & 21.0 & 21.1 & 20.5 & 21.4 & 23.3 & 16.7 \\
\bottomrule
\end{tabular}
\par\medskip
\textbf{(b) Agentic search systems}\par\smallskip
\begin{tabular}{l cccccc}
\toprule
\textbf{System} & \textbf{Causal} & \textbf{Compar.} & \textbf{Forecast} & \textbf{Risk} & \textbf{Effect.} & \textbf{Quant.} \\
\midrule
TTD-DR & \underline{27.5} & \textbf{31.0} & \textbf{31.1} & \underline{35.9} & \textbf{34.8} & \textbf{32.4} \\
GPT-Researcher & \textbf{27.8} & \underline{27.5} & \underline{29.5} & \textbf{36.9} & \underline{31.6} & \underline{32.2} \\
deepagents & 24.9 & 23.8 & \textbf{31.1} & 32.0 & 29.5 & 26.8 \\
OpenClaw & 24.4 & 26.1 & 27.6 & 32.7 & 30.2 & 26.2 \\
STORM & 25.0 & 26.1 & 29.1 & 30.5 & 28.4 & 25.1 \\
\bottomrule
\end{tabular}
\par\medskip
\textbf{(c) Proprietary with search}\par\smallskip
\begin{tabular}{l cccccc}
\toprule
\textbf{System} & \textbf{Causal} & \textbf{Compar.} & \textbf{Forecast} & \textbf{Risk} & \textbf{Effect.} & \textbf{Quant.} \\
\midrule
Claude-Opus-4.7 & \textbf{30.2} & \textbf{33.0} & \textbf{36.1} & \underline{38.1} & \textbf{36.2} & \textbf{31.3} \\
Gemini-3.1-Pro & \underline{29.7} & \underline{30.2} & \underline{35.2} & 38.0 & \underline{35.2} & 29.8 \\
GPT-5.5 & 27.4 & 29.4 & 34.2 & \textbf{38.6} & 32.5 & 30.7 \\
Gemini-3-Flash & 27.2 & 26.1 & 32.2 & 33.7 & 31.8 & \underline{30.9} \\
\bottomrule
\end{tabular}
\par\medskip
\textbf{(d) Open-weight with search}\par\smallskip
\begin{tabular}{l cccccc}
\toprule
\textbf{System} & \textbf{Causal} & \textbf{Compar.} & \textbf{Forecast} & \textbf{Risk} & \textbf{Effect.} & \textbf{Quant.} \\
\midrule
\textbf{FinanceHarness} & \textbf{29.8} & \textbf{29.7} & \underline{32.2} & \textbf{39.5} & \textbf{32.9} & \textbf{31.8} \\
GLM-5 & \underline{26.8} & 24.0 & \textbf{32.5} & \underline{35.4} & \underline{32.8} & 30.5 \\
DeepSeek-v3.2 & 26.3 & \underline{26.5} & 30.4 & 31.9 & 29.9 & \underline{30.6} \\
Qwen3-235B-A22B & 23.7 & 23.8 & 28.0 & 31.0 & 28.8 & 25.6 \\
Gemma-4-26B & 22.6 & 24.8 & 27.4 & 29.1 & 26.6 & 26.0 \\
gpt-oss-120b & 17.3 & 19.9 & 18.0 & 21.9 & 19.9 & 14.5 \\
\bottomrule
\end{tabular}
\end{table*}

\clearpage
\section{\workname{} Case-Study Reports}
\label{sec:appendix-cases}

The boxed reports below are produced by \workname{}, drawn from the
cross-backbone coverage evaluation (Appendix~\ref{sec:appendix-capability-profile}).
The six subsections span distinct tasks: an industry value-chain deep dive, a
single-name discounted-cash-flow valuation, risk analytics, mega-cap relative
comparables, option pricing, and fixed-income analysis. Each box gives the research
question, the resulting cited report, and its sources; round-by-round tool
trajectories for a representative subset are shown separately in
Appendix~\ref{sec:appendix-traces}. These are live-web runs,
separate from the point-in-time \gymname{} benchmark and not part of the headline
results; the generated financial claims have not been independently verified and
illustrate the harness workflow rather than provide investment or factual advice.

\subsection{Industry Deep Research}
\begin{tcolorbox}[
    breakable, colback=caseRepBack, colframe=caseRepFrame, coltitle=black,
    boxrule=0.5pt, arc=0.5mm, top=2mm, bottom=2mm, left=2mm, right=2mm,
    fonttitle=\usefont{T1}{ppl}{b}{n},
    title={AI Infrastructure Value Chain: Beneficiaries Beyond the GPU}]
\usefont{T1}{ppl}{m}{n}\footnotesize

\textbf{\textsc{I. Research Question}}\par\smallskip
\textit{AI compute demand is driving a historic buildout of GPUs and data-center infrastructure. As a research analyst, work outward from the AI accelerators to map the broader value chain: identify the successive layers of industries and suppliers whose demand is pulled by this buildout. For each layer you surface, name the leading public companies, explain how their business connects to AI-compute growth, and assess their financial profile and positioning (growth, margins, valuation). Conclude with a synthesized view of where along the chain the exposure looks most attractive and durable, and the key risks.}
\par\medskip\noindent{\color{gray!45}\rule{\linewidth}{0.35pt}}\par\medskip
\textbf{\textsc{II. Cited Report}}\par\smallskip
\par\addvspace{1pt}\noindent{\itshape AI Infrastructure Value Chain: Scarcity Rents Extend Well Beyond the GPU}\par\nopagebreak\addvspace{6pt}

\par\addvspace{5pt}\noindent\textbf{Synthesis}\par\nopagebreak\addvspace{1.5pt}

The AI buildout is pulling demand through two interconnected chains:

\begin{enumerate}
\item \textbf{The silicon chain:} accelerators $\rightarrow$ HBM $\rightarrow$ advanced packaging/foundries $\rightarrow$ semiconductor equipment.
\item \textbf{The physical-infrastructure chain:} servers $\rightarrow$ networking/optics $\rightarrow$ electrical and cooling equipment $\rightarrow$ construction/grid interconnection $\rightarrow$ power generation $\rightarrow$ data-center real estate.
\end{enumerate}

The most durable economics appear where technology, qualification requirements or physical constraints limit substitution: \textbf{NVIDIA and Broadcom in compute silicon, TSMC in advanced manufacturing and packaging, ASML in lithography, Arista in AI networking, and Eaton/Vertiv in power delivery and cooling}. The better valuation asymmetry currently appears in \textbf{TSMC and selected power producers}, while memory, optical components and server assemblers offer stronger near-term growth but more cyclical or commoditized economics.

At the other extreme, data-center REITs benefit directly from capacity scarcity but combine high capital intensity, approximately 5--6x net leverage and elevated valuation multiples. They are comparatively weak expressions of the theme unless power-secured sites can earn exceptional returns.

\par\addvspace{5pt}\noindent\textbf{Demand anchor}\par\nopagebreak\addvspace{1.5pt}

Based on reported company plans cited by CNBC, Alphabet, Amazon, Microsoft and Meta together indicate roughly \textbf{\$734 billion of 2026 capital expenditure and finance leases}. The definitions differ and not all of the spending is AI-related, but the scale establishes the demand envelope. The main question has shifted from whether infrastructure spending is rising to whether the associated revenue can generate adequate returns. Alphabet's cash flow turned negative in Q2, while debt increased materially at both Alphabet and Amazon [9].

Alphabet illustrates the composition. Its Q2 capital expenditure was \$44.9 billion, approximately 60\% for servers and 40\% for data centers and networking. Google Cloud revenue grew 82\% and backlog reached \$514 billion, suggesting that unusually high spending is still accompanied by capacity-constrained demand [8].

The physical consequence is equally significant. The IEA expects global data-center electricity consumption to more than double from 415 TWh in 2024 to approximately 945 TWh in 2030. Around 20\% of planned projects could experience delays because transmission, transformers, cables and generating capacity are not being added fast enough [6].

\par\addvspace{5pt}\noindent\textbf{Financial snapshot}\par\nopagebreak\addvspace{1.5pt}

The table uses the latest available market-data snapshot as of July 28, 2026. Growth is recent year-over-year revenue growth, margins are operating margins, and valuation is forward P/E unless otherwise indicated. These are screening metrics rather than fully normalized estimates; memory, utilities and hardware assemblers are especially cycle-sensitive.

\par\smallskip\begin{center}\resizebox{\ifdim\width>\linewidth \linewidth\else\width\fi}{!}{%
\begin{tabular}{llll}
\toprule
\textbf{Layer and selected companies} & \textbf{Revenue growth} & \textbf{Operating margin} & \textbf{Valuation} \\
\midrule
\textbf{Accelerators:} NVIDIA / AMD / Broadcom & 85\% / 38\% / 48\% & 66\% / 14\% / 49\% & 15x / 33x / 19x \\
\textbf{Memory:} Micron & 346\% & 80\% & 5x \\
\textbf{Foundry:} TSMC & 36\% & 60\% & 18x \\
\textbf{Equipment:} ASML / Lam / KLA & 21\% / 24\% / 12\% & 37\% / 35\% / 41\% & 27x / 32x / 37x \\
\textbf{Servers:} Dell / Super Micro / HPE & 19\% FY / 123\% / 40\% & 12\% ISG / 6\% / 9\% & 17x / 8x / 11x \\
\textbf{Networking:} Arista / Marvell / Cisco & 35\% / 28\% / 12\% & 43\% / 15\% / 25\% & 37x / 27x / 24x \\
\textbf{Optics:} Coherent / Lumentum & 21\% / 90\% & 14\% / 22\% & 28x / 34x \\
\textbf{Power and cooling:} Vertiv / Eaton & 30\% / 17\% & 16\% / 16\% & 29x / 24x \\
\textbf{Construction/grid:} Quanta Services / EMCOR & 26\% / 20\% & 4\% / 9\% & 35x / 21x \\
\textbf{Generation:} Constellation / Vistra / GE Vernova & 64\% / 43\% / 22\% & 22\% / 27\% / 8\% & 19x / 14x / 37x \\
\textbf{Data-center REITs:} Equinix / Digital Realty & 12\% / 30\% & 24\% / 26\% & 29x / 29x EV/EBITDA \\
\textbf{Cloud buyers:} Microsoft / Alphabet / Meta & 18\% / 24\% / 33\% & 46\% / 34\% / 41\% & 20x / 22x / 16x \\
\bottomrule
\end{tabular}}\end{center}\par\smallskip

\par\addvspace{5pt}\noindent\textbf{The successive layers}\par\nopagebreak\addvspace{1.5pt}

\par\addvspace{4pt}\noindent\hangindent=1.9em\hangafter=1\makebox[1.5em][r]{\bfseries\color{caseBadge}1.}\hspace{0.4em}\textbf{Accelerators and custom AI silicon}\par\nopagebreak\addvspace{1.5pt}

\textbf{Selected public companies:} NVIDIA, AMD, Broadcom, Marvell.

This is the starting point and the most profitable layer in the current financial snapshot. NVIDIA supplies accelerators, networking products, systems and an extensive software platform. AMD is the principal listed merchant alternative in the screen, while Broadcom and Marvell provide exposure to custom silicon and networking.

NVIDIA's Q1 fiscal 2027 Data Center revenue reached \$75.2 billion---approximately 92\% of total revenue---and grew 92\% year over year. Its 74.9\% GAAP gross margin and roughly 66\% operating margin demonstrate the current scarcity rents associated with its data-center platform [1].

Broadcom provides a counterweight to NVIDIA's merchant model. Its AI revenue reached \$10.8 billion in fiscal Q2, more than doubling year over year, driven by custom accelerators, including Google's TPU, and networking components [11]. Broadcom is therefore exposed both to the expansion of AI clusters and to hyperscalers' efforts to deploy internally designed accelerators.

\textbf{Positioning:} This layer has the strongest current combination of growth, margins and technical barriers. Its weakness is customer concentration: a small number of hyperscalers control much of incremental spending and are simultaneously developing competing silicon.

\par\addvspace{4pt}\noindent\hangindent=1.9em\hangafter=1\makebox[1.5em][r]{\bfseries\color{caseBadge}2.}\hspace{0.4em}\textbf{High-bandwidth memory and data-center memory}\par\nopagebreak\addvspace{1.5pt}

\textbf{Selected public companies:} SK Hynix, Micron, Samsung Electronics.

An accelerator requires HBM, server DRAM and increasingly enterprise SSD capacity. HBM is especially demanding because multiple memory dies must be stacked, tested and integrated alongside the processor through advanced packaging.

SK Hynix's 2026 market-outlook page describes direct exposure through HBM3E, HBM4, server DRAM and enterprise SSDs. Citing Bank of America estimates, it places the 2026 HBM market at \$54.6 billion, up 58\%, while also warning that capacity additions and competition could create price pressure after 2026 [18].

Micron's fiscal Q3 revenue rose from \$9.3 billion to \$41.5 billion year over year. Gross margin reached 84.6\% and operating margin 80.4\%, with management introducing multiyear customer agreements intended to make results more predictable [10]. Those margins should not be treated as normal through-cycle economics: Micron itself is investing at record levels in technology, products and supply, while the broader memory outlook acknowledges the possibility of future price correction [10,18].

\textbf{Positioning:} HBM is currently a bottleneck and may remain more differentiated than conventional memory. Nevertheless, the very low forward multiples reflect valid concerns about peak pricing, heavy capital expenditure and eventual supply response.

\par\addvspace{4pt}\noindent\hangindent=1.9em\hangafter=1\makebox[1.5em][r]{\bfseries\color{caseBadge}3.}\hspace{0.4em}\textbf{Foundry manufacturing and advanced packaging}\par\nopagebreak\addvspace{1.5pt}

\textbf{Leading public exposure:} TSMC.

AI accelerators require leading-edge logic and advanced packaging to connect compute dies with HBM. TSMC's role in both advanced process nodes and packaging makes it more than a conventional wafer foundry.

TSMC reported Q2 revenue of \$40.2 billion, a 67.7\% gross margin and a 60.3\% operating margin [2]. Management characterized AI demand as ``extremely robust,'' raised expected 2026 revenue growth to slightly above 40\%, and increased planned capital expenditure to \$60--64 billion. Approximately 70--80\% is allocated to advanced processes and 10--20\% to packaging, testing and related capacity [15].

The company screens at approximately 18x forward earnings despite growth and margins comparable with the highest-quality semiconductor companies in the market-data set. The valuation also reflects material risks: overseas expansion is expected to dilute margins, the N2 ramp will carry startup costs, and a large share of current production remains geographically concentrated.

\textbf{Positioning:} TSMC is arguably the best risk-adjusted tollbooth in the chain. Its foundry model provides exposure to demand from cloud providers, accelerator customers and CPU suppliers without relying on a single chip architecture. Geographic concentration is the dominant risk.

\par\addvspace{4pt}\noindent\hangindent=1.9em\hangafter=1\makebox[1.5em][r]{\bfseries\color{caseBadge}4.}\hspace{0.4em}\textbf{Wafer-fabrication and process-control equipment}\par\nopagebreak\addvspace{1.5pt}

\textbf{Selected public companies:} ASML, Applied Materials, Lam Research, KLA.

More AI silicon eventually requires additional lithography, deposition, etch, inspection and metrology capacity. Equipment suppliers participate with a lag: end demand first raises utilization at foundries and memory manufacturers, which then place additional tool orders.

ASML reported \texteuro{}32.7 billion of 2025 sales, a 52.8\% gross margin and a \texteuro{}38.8 billion year-end backlog. Q4 bookings included \texteuro{}7.4 billion of EUV orders. Management said stronger customer confidence in sustained AI demand had materially increased medium-term capacity plans and order intake [3].

The market-data screen shows operating margins of 35\% at Lam and 41\% at KLA, versus 37\% at ASML. Their valuations of roughly 27--37x forward earnings reflect strong current returns and expectations that AI-related semiconductor investment will remain elevated.

\textbf{Positioning:} This is one of the more durable layers because advanced tools are technically complex and serve long-lived customer manufacturing programs. The trade-off is semiconductor-capital-equipment cyclicality and premium valuations.

\par\addvspace{4pt}\noindent\hangindent=1.9em\hangafter=1\makebox[1.5em][r]{\bfseries\color{caseBadge}5.}\hspace{0.4em}\textbf{Boards, servers, racks and systems integration}\par\nopagebreak\addvspace{1.5pt}

\textbf{Selected public companies:} Dell, Super Micro Computer, HPE.

These companies turn accelerators, CPUs, memory, network cards, power supplies and cooling components into deployable systems. AI servers generate exceptional revenue growth because accelerator content per server is extremely high, but much of that revenue is pass-through component value.

Dell shipped more than \$25 billion of AI-optimized servers in FY26, booked more than \$64 billion of orders and entered FY27 with a \$43 billion backlog. Its Infrastructure Solutions Group revenue grew 40\%, but operating margin declined from 12.8\% to 11.7\%, demonstrating that volume does not automatically produce scarcity-level profitability [12].

Super Micro's market-data profile shows an 8\% gross margin, a 6\% operating margin and approximately 5x net debt/EBITDA. Dell's broader reported portfolio, larger cash generation and lower net leverage in the comparative screen provide a more balanced profile, although its margins remain far below those of semiconductor and networking suppliers.

\textbf{Positioning:} Strong demand exposure but weak structural economics. Buyers have alternatives, component suppliers capture much of the value, and working-capital requirements rise sharply during rapid growth. This is primarily a volume trade rather than a durable profit pool.

\par\addvspace{4pt}\noindent\hangindent=1.9em\hangafter=1\makebox[1.5em][r]{\bfseries\color{caseBadge}6.}\hspace{0.4em}\textbf{Ethernet switching, interconnect and optical components}\par\nopagebreak\addvspace{1.5pt}

\textbf{Selected public companies:} Arista Networks, Broadcom, Marvell, Cisco, Coherent, Lumentum.

Large clusters require both:

\begin{itemize}
\item \textbf{Scale-up networking} between accelerators within a rack or tightly coupled system.
\item \textbf{Scale-out networking} connecting many racks into a larger training or inference fabric.
\end{itemize}

Arista's announced 1.6-terabit portfolio is designed for rack-scale AI infrastructure and clusters expanding from thousands to hundreds of thousands of accelerators, including air-, liquid- and hybrid-cooled systems. The products use Broadcom Tomahawk 6 switching silicon, illustrating how several companies can participate in the same network deployment [17]. Arista's approximately 64\% gross margin, 43\% operating margin and net-cash balance sheet distinguish it from lower-margin component suppliers.

NVIDIA's own networking revenue grew 199\% to \$14.8 billion, substantially faster than its 77\% growth in Data Center compute revenue during the same quarter [1]. Coherent supplies optical products used in data-center communications; its fiscal Q3 revenue grew 21\%, gross margin reached 37.7\%, and management said it was expanding capacity in response to AI-data-center demand [13].

\textbf{Positioning:} Arista combines attractive economics with a differentiated network operating stack, but trades near 37x forward earnings. Optical suppliers offer higher operating leverage but face rapid product transitions, customer concentration and changing interconnect architectures.

\par\addvspace{4pt}\noindent\hangindent=1.9em\hangafter=1\makebox[1.5em][r]{\bfseries\color{caseBadge}7.}\hspace{0.4em}\textbf{Electrical equipment and thermal management}\par\nopagebreak\addvspace{1.5pt}

\textbf{Leading public companies:} Vertiv, Eaton.

GPU density turns power conversion and heat removal into binding constraints. The equipment set includes switchgear, uninterruptible power supplies, power distribution and cooling systems.

Vertiv is the higher-purity listed exposure in this comparison. Q1 sales grew 30\%, with Americas organic growth of 44\% on data-center demand. Adjusted operating margin rose 430 basis points to 20.8\%, while full-year organic growth guidance increased to 29--31\%. Net leverage was only about 0.2x [4].

Eaton supplies power and energy management, backup power, software, safety, design and project-management services to data-center operators. The company explicitly identifies dense GPU clusters as creating new electrical, thermal and operational stress [16]. Its 24x forward multiple is less demanding than Vertiv's 29x, though its overall business is less concentrated on data centers.

\textbf{Positioning:} This is one of the most attractive second-order layers. Reliability requirements, manufacturing lead times and the need to redesign facilities for higher rack density support demand. Unlike a bet on one accelerator architecture, much of this spending remains necessary across competing compute platforms. The principal risk is that high valuations already discount several years of exceptional growth.

\par\addvspace{4pt}\noindent\hangindent=1.9em\hangafter=1\makebox[1.5em][r]{\bfseries\color{caseBadge}8.}\hspace{0.4em}\textbf{Data-center construction and grid interconnection}\par\nopagebreak\addvspace{1.5pt}

\textbf{Selected public companies:} Quanta Services, EMCOR, Comfort Systems USA.

These businesses provide exposure to transmission, substations, electrical work, mechanical systems and other large-load infrastructure. They benefit from the overlap among data-center construction, utility load growth and generation investment.

Quanta reported Q1 revenue of \$7.87 billion, up 26\%, and record total backlog of \$48.5 billion. Management cited converging utility, generation and large-load markets as the demand driver and identified data centers among potential future project areas [5].

The market-data snapshot shows higher operating margins at EMCOR and Comfort Systems than at Quanta: approximately 9\% and 8\%, respectively, versus 4\%. Quanta's roughly 35x forward multiple is high for its current operating margin, while EMCOR screens at approximately 21x.

\textbf{Positioning:} Backlogs provide visibility, but this remains a labor- and execution-intensive layer. Fixed-price contracts, skilled-labor shortages, permitting and project delays can erode the apparent backlog advantage.

\par\addvspace{4pt}\noindent\hangindent=1.9em\hangafter=1\makebox[1.5em][r]{\bfseries\color{caseBadge}9.}\hspace{0.4em}\textbf{Electricity generation and generation equipment}\par\nopagebreak\addvspace{1.5pt}

\textbf{Selected public companies:} Constellation Energy, Vistra, NextEra Energy, GE Vernova, Caterpillar, Cummins.

Data centers need continuous, high-quality power. The IEA expects electricity generation serving data centers to exceed 1,000 TWh by 2030. Natural gas and coal are expected to meet more than 40\% of incremental demand through 2030, despite substantial renewable additions; nuclear becomes more important after 2030 [14].

Public-market exposure can therefore be sought through three broad groups:

\begin{itemize}
\item \textbf{Electricity producers:} Constellation, Vistra and NextEra.
\item \textbf{Generation and grid equipment:} GE Vernova.
\item \textbf{Backup and distributed-generation equipment:} Caterpillar and Cummins.
\end{itemize}

Constellation and Vistra screen at approximately 19x and 14x forward earnings, respectively---below many equipment beneficiaries. Their financial exposure, however, depends on regional power markets, contract structures, operating performance and regulation. GE Vernova offers more direct equipment exposure but trades around 37x forward earnings despite an 8\% operating margin.

\textbf{Positioning:} This may offer some of the best valuation asymmetry in the physical chain, but it is not a clean or uniform exposure. Local transmission constraints mean national electricity growth will not translate evenly into company profits, and regulators can materially influence who pays for interconnection and generation investment.

\par\addvspace{4pt}\noindent\hangindent=1.9em\hangafter=1\makebox[1.5em][r]{\bfseries\color{caseBadge}10.}\hspace{0.4em}\textbf{Data-center landlords and interconnection platforms}\par\nopagebreak\addvspace{1.5pt}

\textbf{Selected public companies:} Equinix, Digital Realty, Iron Mountain.

These businesses monetize power-secured land, buildings and network interconnection. Equinix reported 10\% Q1 revenue growth, a 51\% adjusted EBITDA margin and 12\% AFFO growth. Approximately 60\% of its largest deals were AI-related [7].

However, Equinix and Digital Realty both trade near 29x EV/EBITDA and carry net debt around 5--6x EBITDA. They must continue investing heavily to deliver capacity: Equinix expects roughly \$4.1 billion of 2026 capital expenditure [7].

\textbf{Positioning:} Demand is real, but economics depend on the spread between development returns and the cost of capital. Rising rates, power procurement, construction delays and customer concentration make this a less attractive expression than supplying scarce equipment to multiple developers and operators.

\par\addvspace{5pt}\noindent\textbf{Where the exposure is most attractive}\par\nopagebreak\addvspace{1.5pt}

\par\addvspace{4pt}\noindent\hangindent=1.9em\hangafter=1\makebox[1.5em][r]{\bfseries\color{caseBadge}1.}\hspace{0.4em}\textbf{Best combination of durability and valuation: \textbf{TSMC}}\par\nopagebreak\addvspace{1.5pt}

TSMC combines 60\% operating margins, above-40\% expected 2026 revenue growth and an approximately 18x forward P/E. Its broad foundry and advanced-packaging role reduces dependence on which accelerator architecture ultimately gains share. Overseas expansion costs and geographic concentration prevent this from being a low-risk investment, but the operating franchise is unusually durable.

\par\addvspace{4pt}\noindent\hangindent=1.9em\hangafter=1\makebox[1.5em][r]{\bfseries\color{caseBadge}2.}\hspace{0.4em}\textbf{Highest-quality technology and ecosystem assets: \textbf{ASML, NVIDIA and Broadcom}}\par\nopagebreak\addvspace{1.5pt}

\begin{itemize}
\item \textbf{ASML's} EUV role, \texteuro{}38.8 billion backlog and installed-base business provide exposure to multiple chipmakers rather than a single end product [3].
\item \textbf{NVIDIA} has the best current economics and the most integrated reported data-center platform in the group [1].
\item \textbf{Broadcom} provides exposure to custom accelerators and the networking required around them [11].
\end{itemize}

Their central risk is that a handful of large customers possess substantial buyer power and are actively funding alternatives.

\par\addvspace{4pt}\noindent\hangindent=1.9em\hangafter=1\makebox[1.5em][r]{\bfseries\color{caseBadge}3.}\hspace{0.4em}\textbf{Best physical-infrastructure beneficiaries: \textbf{Eaton and Vertiv}}\par\nopagebreak\addvspace{1.5pt}

Power and cooling requirements increase with rack density across competing accelerator architectures. Eaton offers broader diversification and a lower multiple; Vertiv offers higher purity, faster growth and greater operational leverage. Neither is now undiscovered, making entry valuation important.

\par\addvspace{4pt}\noindent\hangindent=1.9em\hangafter=1\makebox[1.5em][r]{\bfseries\color{caseBadge}4.}\hspace{0.4em}\textbf{Best networking asset: \textbf{Arista}}\par\nopagebreak\addvspace{1.5pt}

Arista's operating margin, balance sheet and AI-focused product roadmap are exceptional. Its announced 1.6T systems address both scale-up and scale-out AI fabrics [17]. The limitation is a roughly 37x forward valuation and dependence on a concentrated set of large cloud and technology customers.

\par\addvspace{4pt}\noindent\hangindent=1.9em\hangafter=1\makebox[1.5em][r]{\bfseries\color{caseBadge}5.}\hspace{0.4em}\textbf{Attractive but higher-risk value: \textbf{Vistra and Constellation}}\par\nopagebreak\addvspace{1.5pt}

These companies provide electricity-generation exposure at less demanding multiples than many AI-equipment beneficiaries. The exposure is nevertheless indirect and heavily influenced by regulation, local power prices, operating performance and transmission availability.

\par\addvspace{4pt}\noindent\hangindent=1.9em\hangafter=1\makebox[1.5em][r]{\bfseries\color{caseBadge}6.}\hspace{0.4em}\textbf{Tactical rather than durable: \textbf{Micron, Coherent, Lumentum, Dell and Super Micro}}\par\nopagebreak\addvspace{1.5pt}

These companies can produce the strongest earnings surprises during scarcity. They also face the greatest risk of margin normalization:

\begin{itemize}
\item Memory supply eventually responds.
\item Optical standards and architectures change quickly.
\item Server assemblers retain relatively little of the accelerator's economics.
\item Working capital and customer concentration magnify downside during pauses.
\end{itemize}

\par\addvspace{4pt}\noindent\hangindent=1.9em\hangafter=1\makebox[1.5em][r]{\bfseries\color{caseBadge}7.}\hspace{0.4em}\textbf{Least compelling at current structure: \textbf{data-center REITs}}\par\nopagebreak\addvspace{1.5pt}

Equinix and Digital Realty have valuable power-secured portfolios, but high leverage, elevated EV/EBITDA multiples and continuing capital requirements limit upside relative to suppliers that can sell equipment into multiple new projects.

\par\addvspace{5pt}\noindent\textbf{Key risks}\par\nopagebreak\addvspace{1.5pt}

\begin{enumerate}
\item \textbf{Hyperscaler return-on-capital failure.} A spending pause would propagate through servers, optics, semiconductors, equipment and construction, although timing would differ by backlog and contract structure. Rising debt and negative free cash flow show that investor tolerance is not unlimited [9].
\item \textbf{Efficiency and substitution.} Better models, inference optimization, lower-precision computation and custom ASICs could reduce GPU requirements per unit of AI output. Broadcom may benefit from custom silicon, but merchant accelerator vendors would face pressure.
\item \textbf{Customer concentration.} A small number of cloud companies have bargaining power over nearly every layer, particularly custom chips, optics, servers and real estate. Broadcom identified six core custom-chip customers as drivers of AI growth [11].
\item \textbf{Capacity overshoot.} HBM, optical components and server assembly are most vulnerable if suppliers extrapolate current shortages into permanent demand. SK Hynix's market outlook specifically identifies possible HBM price correction after 2026 as competition and capacity increase [18].
\item \textbf{Geopolitics and export controls.} The chain remains heavily dependent on TSMC's manufacturing footprint. China restrictions can also remove accelerator revenue directly: NVIDIA's Q2 outlook assumed no Data Center compute revenue from China [1].
\item \textbf{Power bottlenecks and regulation.} Transmission projects can take four to eight years, and grid constraints may delay data centers even when server demand remains strong [6]. Regulators can also determine how interconnection and generation costs are distributed.
\item \textbf{Technology transitions.} HBM4, N2, 1.6T Ethernet and liquid cooling require coordinated execution across multiple suppliers. The relevant company sources describe continuing ramps or future availability rather than fully completed transitions [15,17,18].
\item \textbf{Valuation compression.} Vertiv, Arista, Quanta and GE Vernova already trade at growth-stock multiples. Even successful earnings delivery may not protect them if the market lowers terminal-growth assumptions.
\end{enumerate}

\par\addvspace{5pt}\noindent\textbf{Bottom line}\par\nopagebreak\addvspace{1.5pt}

The AI infrastructure boom is not one trade but a hierarchy of profit pools. Economic value is currently highest where there is \textbf{proprietary technology plus scarce capacity}---accelerators, advanced foundry and packaging, lithography and high-performance networking. The next most durable opportunity is in \textbf{power distribution and cooling}, where demand is less dependent on which compute architecture wins.

The preferred risk-adjusted exposures are therefore \textbf{TSMC first; ASML, NVIDIA and Broadcom for technology scarcity; Eaton and Vertiv for the physical bottleneck; and Arista for network scaling}. \textbf{Vistra and Constellation} offer cheaper but more policy-sensitive exposure. Memory, optics and server assemblers are better treated as cyclical beneficiaries, while data-center REITs currently offer the least favorable combination of valuation, leverage and capital intensity.
\par\medskip\noindent{\color{gray!45}\rule{\linewidth}{0.35pt}}\par\medskip
\textbf{\textsc{III. Sources}}\par\nopagebreak\smallskip
{\scriptsize\raggedright {[1]}~\fhlink{https://investor.nvidia.com/news/press-release-details/2026/NVIDIA-Announces-Financial-Results-for-First-Quarter-Fiscal-2027/default.aspx}{NVIDIA Announces Financial Results for First Quarter Fiscal 2027} \\
{[2]}~\fhlink{https://investor.tsmc.com/english/quarterly-results/2026/q2}{TSMC 2026 Q2 Quarterly Results - Taiwan Semiconductor ...} \\
{[3]}~\fhlink{https://www.asml.com/en/news/press-releases/2026/q4-2025-financial-results}{ASML reports \texteuro{}32.7 billion total net sales and \texteuro{}9.6 billion ...} \\
{[4]}~\fhlink{https://investors.vertiv.com/news/news-details/2026/Vertiv-Reports-Strong-First-Quarter-with-Diluted-EPS-Growth-of-136-Adjusted-Diluted-EPS-Growth-of-83-Raises-Full-Year-Guidance/default.aspx}{Vertiv Holdings Co. - Vertiv Reports Strong First Quarter ...} \\
{[5]}~\fhlink{https://investors.quantaservices.com/news-events/press-releases/detail/396/quanta-services-reports-first-quarter-2026-results}{QUANTA SERVICES REPORTS FIRST QUARTER 2026 RESULTS} \\
{[6]}~\fhlink{https://www.iea.org/reports/energy-and-ai/executive-summary}{Executive summary -- Energy and AI -- Analysis - IEA} \\
{[7]}~\fhlink{https://investor.equinix.com/news-events/press-releases/detail/1107/equinix-reports-first-quarter-results-and-raises-full-year}{Equinix Reports First-Quarter Results and... :: Equinix, Inc. (EQIX)} \\
{[8]}~\fhlink{https://abc.xyz/investor/events/event-details/2026/2026-Q2-Earnings-Call-2026-GgTAq7Is0z/default.aspx}{2026 Q2 Earnings Call - Alphabet Investor Relations} \\
{[9]}~\fhlink{https://www.cnbc.com/2026/07/28/hyperscalers-face-higher-capex-scrutiny-after-alphabet-report-panned.html}{Hyperscalers face higher capex scrutiny after Alphabet report panned} \\
{[10]}~\fhlink{https://www.nasdaq.com/press-release/micron-technology-inc-reports-record-results-third-quarter-fiscal-2026-2026-06-24}{Micron Technology, Inc. Reports Record Results for the Third ...} \\
{[11]}~\fhlink{https://www.cnbc.com/2026/06/03/broadcom-avgo-earnings-report-q2-2026.html}{Broadcom (AVGO) earnings report Q2 2026 - CNBC} \\
{[12]}~\fhlink{https://www.dell.com/en-us/dt/corporate/newsroom/announcements/detailpage.press-releases~usa~2026~2~dell-technologies-delivers-fourth-quarter-and-full-year-fiscal-2026-results.htm}{Dell Technologies Delivers Fourth Quarter and Full-Year ...} \\
{[13]}~\fhlink{https://www.coherent.com/news/press-releases/third-quarter-fiscal-year-2026-results}{Coherent Corp. Reports Third Quarter Fiscal 2026 Results} \\
{[14]}~\fhlink{https://www.iea.org/reports/energy-and-ai/energy-supply-for-ai}{Energy supply for AI -- Energy and AI -- Analysis - IEA} \\
{[15]}~\fhlink{https://investor.tsmc.com/english/encrypt/files/encrypt_file/reports/2026-07/547d1696765e05ce3adb81c108ce1c8c1682b80c/TSMC+2Q26+Transcript.pdf}{Q2 2026 Taiwan Semiconductor Manufacturing Co Ltd Earnings ...} \\
{[16]}~\fhlink{http://www.electrical.eaton.com/us/en-us/markets/data-centers.html}{Data centers | Digitalization | Efficiency | Eaton} \\
{[17]}~\fhlink{https://investors.arista.com/Communications/Press-Releases-and-Events/Press-Release-Detail/2026/Arista-Introduces-Next-Generation-1-6Terabit-Portfolio-for-AI-Fabrics/default.aspx}{Arista Networks - Arista Introduces Next-Generation 1.6Terabit ...} \\
{[18]}~\fhlink{https://news.skhynix.com/en/2026-market-outlook-focus-on-the-hbm-led-memory-supercycle/}{2026 Market Outlook -- ``Focus on the HBM-Led Memory Supercycle''}\par}
\end{tcolorbox}

\clearpage
\subsection{Valuation Compute Seam}
\begin{tcolorbox}[
    breakable, colback=caseRepBack, colframe=caseRepFrame, coltitle=black,
    boxrule=0.5pt, arc=0.5mm, top=2mm, bottom=2mm, left=2mm, right=2mm,
    fonttitle=\usefont{T1}{ppl}{b}{n},
    title={Microsoft (MSFT): DCF, WACC \& Sensitivity}]
\usefont{T1}{ppl}{m}{n}\footnotesize

\textbf{\textsc{I. Research Question}}\par\smallskip
\textit{Build an auditable five-year unlevered DCF for Microsoft (MSFT) from its latest fiscal-year results. Forecast revenue, EBIT margin, taxes, D\&A, capex, and changes in working capital using clearly stated assumptions; derive WACC from current market inputs, a peer-based beta, prevailing Treasury yields, and an explicit equity-risk premium; use a 2.5\% perpetual-growth rate; bridge enterprise value to equity value and intrinsic value per diluted share. Include a sensitivity table spanning WACC $\pm$2 percentage points and terminal growth from 1.5\% to 3.5\%. Compare the result with the current market price without making an investment recommendation.}
\par\medskip\noindent{\color{gray!45}\rule{\linewidth}{0.35pt}}\par\medskip
\textbf{\textsc{II. Cited Report}}\par\smallskip
\par\addvspace{1pt}\noindent{\itshape Microsoft five-year unlevered DCF: \$233 per diluted share}\par\nopagebreak\addvspace{6pt}

Using Microsoft's FY2025 results as the operating base, a five-year unlevered DCF produces:

\begin{itemize}
\item \textbf{Enterprise value:} \$1.674 trillion
\item \textbf{Equity value:} \$1.741 trillion
\item \textbf{Intrinsic value:} \textbf{\$233.19 per diluted share}
\item \textbf{Current market price:} \$381.70
\item \textbf{DCF-to-market difference:} \textbf{38.9\% below} the current price
\end{itemize}

This is a valuation result, not an investment recommendation. The model's largest judgment calls are the \textbf{11.60\% peer-beta-based WACC} and the pace at which elevated AI infrastructure spending normalizes.

\par\addvspace{4pt}\noindent\hangindent=1.9em\hangafter=1\makebox[1.5em][r]{\bfseries\color{caseBadge}1.}\hspace{0.4em}\textbf{FY2025 operating base}\par\nopagebreak\addvspace{1.5pt}

Microsoft's fiscal year ended June 30, 2025. Revenue was \$281.724 billion and operating income---used here as EBIT---was \$128.528 billion, producing a 45.62\% EBIT margin. The reported income-tax provision was \$21.795 billion on \$123.627 billion of pretax income, an effective rate of 17.63\%.[2]

\par\smallskip\begin{center}\resizebox{\ifdim\width>\linewidth \linewidth\else\width\fi}{!}{%
\begin{tabular}{ll}
\toprule
\textbf{FY2025 anchor} & \textbf{Amount} \\
\midrule
Revenue & \$281.724B \\
EBIT & \$128.528B \\
EBIT margin & 45.62\% \\
Reported income-tax provision & \$21.795B \\
Effective tax rate & 17.63\% \\
D\&A and other & \$34.153B \\
Capital expenditures & \$64.551B \\
Operating cash flow & \$136.162B \\
Reported OCF less capex & \$71.611B \\
\bottomrule
\end{tabular}}\end{center}\par\smallskip

D\&A increased from \$22.287 billion in FY2024 to \$34.153 billion in FY2025, while capital expenditures increased from \$44.477 billion to \$64.551 billion.[3] Those increases---53.2\% and 45.1\%, respectively---support modeling a period in which depreciation catches up with the recent infrastructure build and capex gradually normalizes.

\par\addvspace{4pt}\noindent{\bfseries\itshape Working-capital baseline}\par\nopagebreak\addvspace{1.5pt}

The FY2025 cash-flow statement reported the following operating-asset and liability changes:[3]

\begin{itemize}
\item Accounts receivable: \$(10.581)B
\item Inventories: +\$0.309B
\item Other current assets: \$(3.044)B
\item Accounts payable: +\$0.569B
\item Unearned revenue: +\$5.438B
\item Other current liabilities: +\$5.922B
\end{itemize}

The net cash effect was a \textbf{\$1.387 billion outflow}, equivalent to 3.79\% of FY2025's \$36.602 billion revenue increase. The forecast rounds this to a \textbf{4.0\% investment for every dollar of incremental revenue}.

\par\addvspace{4pt}\noindent\hangindent=1.9em\hangafter=1\makebox[1.5em][r]{\bfseries\color{caseBadge}2.}\hspace{0.4em}\textbf{Forecast assumptions}\par\nopagebreak\addvspace{1.5pt}

These are explicit analyst assumptions rather than company guidance or consensus estimates.

\par\smallskip\begin{center}\resizebox{\ifdim\width>\linewidth \linewidth\else\width\fi}{!}{%
\begin{tabular}{llrrrr}
\toprule
\textbf{Assumption} & \textbf{FY26E} & \textbf{FY27E} & \textbf{FY28E} & \textbf{FY29E} & \textbf{FY30E} \\
\midrule
Revenue growth & 14.0\% & 13.0\% & 12.0\% & 10.0\% & 8.0\% \\
EBIT margin & 45.7\% & 45.9\% & 46.1\% & 46.3\% & 46.5\% \\
Tax rate on EBIT & 18.0\% & 18.0\% & 18.0\% & 18.0\% & 18.0\% \\
D\&A as \% of revenue & 13.5\% & 14.5\% & 15.0\% & 15.2\% & 15.0\% \\
Capex as \% of revenue & 22.0\% & 20.0\% & 18.0\% & 16.5\% & 15.0\% \\
Change in NWC & \multicolumn{5}{l}{4\% of incremental revenue in every year} \\
\bottomrule
\end{tabular}}\end{center}\par\smallskip

The revenue path represents an 11.38\% five-year CAGR, decelerating from FY2025's 15\% reported growth. Microsoft reported Azure annual revenue above \$75 billion, up 34\%, providing context for maintaining double-digit consolidated growth through FY2028E.[2]

The EBIT-margin assumption provides only modest expansion---45.62\% to 46.5\%---because scale and product mix benefits are assumed to be partly offset by rising depreciation from the infrastructure build. Capex remains above D\&A through FY2029E and converges with D\&A in FY2030E.

\par\addvspace{4pt}\noindent\hangindent=1.9em\hangafter=1\makebox[1.5em][r]{\bfseries\color{caseBadge}3.}\hspace{0.4em}\textbf{Five-year unlevered free-cash-flow forecast}\par\nopagebreak\addvspace{1.5pt}

Formula:

\[\text{UFCF}=\text{EBIT}\times(1-\text{tax rate})+\text{D\&A}-\text{capex}-\Delta\text{NWC}\]

Amounts are in billions of dollars.

\par\smallskip\begin{center}\resizebox{\ifdim\width>\linewidth \linewidth\else\width\fi}{!}{%
\begin{tabular}{lrrrrr}
\toprule
\textbf{} & \textbf{FY26E} & \textbf{FY27E} & \textbf{FY28E} & \textbf{FY29E} & \textbf{FY30E} \\
\midrule
Revenue & \$321.165 & \$362.917 & \$406.467 & \$447.114 & \$482.883 \\
EBIT & 146.773 & 166.579 & 187.381 & 207.014 & 224.540 \\
Taxes on EBIT & (26.419) & (29.984) & (33.729) & (37.263) & (40.417) \\
\textbf{NOPAT} & \textbf{120.354} & \textbf{136.595} & \textbf{153.653} & \textbf{169.751} & \textbf{184.123} \\
D\&A & 43.357 & 52.623 & 60.970 & 67.961 & 72.432 \\
Capex & (70.656) & (72.583) & (73.164) & (73.774) & (72.432) \\
Change in NWC & (1.578) & (1.670) & (1.742) & (1.626) & (1.431) \\
\textbf{Unlevered FCF} & \textbf{\$91.477} & \textbf{\$114.964} & \textbf{\$139.717} & \textbf{\$162.313} & \textbf{\$182.692} \\
\bottomrule
\end{tabular}}\end{center}\par\smallskip

The DCF calculation uses the displayed unrounded FCF schedule.

\par\addvspace{4pt}\noindent\hangindent=1.9em\hangafter=1\makebox[1.5em][r]{\bfseries\color{caseBadge}4.}\hspace{0.4em}\textbf{WACC derivation}\par\nopagebreak\addvspace{1.5pt}

\par\addvspace{4pt}\noindent{\bfseries\itshape Peer-based beta}\par\nopagebreak\addvspace{1.5pt}

The model uses the equal-weighted mean of current observed equity betas for five large cloud, enterprise-software and platform peers:

\par\smallskip\begin{center}\resizebox{\ifdim\width>\linewidth \linewidth\else\width\fi}{!}{%
\begin{tabular}{lr}
\toprule
\textbf{Peer} & \textbf{Equity beta} \\
\midrule
Alphabet & 1.25 \\
Amazon & 1.46 \\
Oracle & 1.71 \\
Salesforce & 1.18 \\
Adobe & 1.43 \\
\textbf{Equal-weighted mean} & \textbf{1.406} \\
\bottomrule
\end{tabular}}\end{center}\par\smallskip

This is an observed peer-equity-beta approach; the individual betas are not unlevered and relevered.

\par\addvspace{4pt}\noindent{\bfseries\itshape Cost of capital}\par\nopagebreak\addvspace{1.5pt}

\par\smallskip\begin{center}\resizebox{\ifdim\width>\linewidth \linewidth\else\width\fi}{!}{%
\begin{tabular}{lrl}
\toprule
\textbf{Input} & \textbf{Value} & \textbf{Basis} \\
\midrule
Risk-free rate & 4.68\% & Current 10-year Treasury yield \\
Equity-risk premium & 5.00\% & Explicit model assumption \\
Peer beta & 1.406 & Equal-weighted peer mean \\
\textbf{Cost of equity} & \textbf{11.71\%} & 4.68\% + 1.406 $\times$ 5.00\% \\
Pretax cost of debt & 5.18\% & Treasury yield plus assumed 0.50\% spread \\
Tax rate & 17.63\% & FY2025 effective tax rate \\
After-tax debt cost & 4.27\% & 5.18\% $\times$ (1 $-$ 17.63\%) \\
Debt/equity & 1.519\% & \$43.151B debt $\div$ \$2.840T market capitalization \\
Equity weight & 98.50\% & Market-value weight \\
Debt weight & 1.50\% & Market-value weight \\
\textbf{WACC} & \textbf{11.5986\%} & Rounded to 11.60\% \\
\bottomrule
\end{tabular}}\end{center}\par\smallskip

Microsoft's FY2025 debt consisted of \$2.999 billion of current long-term debt and \$40.152 billion of long-term debt; short-term debt was zero.[4]

\par\addvspace{4pt}\noindent\hangindent=1.9em\hangafter=1\makebox[1.5em][r]{\bfseries\color{caseBadge}5.}\hspace{0.4em}\textbf{Terminal value and enterprise-to-equity bridge}\par\nopagebreak\addvspace{1.5pt}

The terminal value uses a \textbf{2.5\% perpetual-growth rate}:

\[\text{Terminal value}_{2030}
=
\frac{\$182.692B\times1.025}{11.5986\%-2.5\%}
=
\$2.058T\]

\par\smallskip\begin{center}\resizebox{\ifdim\width>\linewidth \linewidth\else\width\fi}{!}{%
\begin{tabular}{ll}
\toprule
\textbf{Valuation component} & \textbf{Amount} \\
\midrule
PV of five explicit FCFs & \$484.990B \\
PV of terminal value & 1,188.980B \\
\textbf{Enterprise value} & \textbf{\$1,673.970B} \\
Add: cash and short-term investments & 94.565B \\
Add: equity and other investments & 15.405B \\
Less: debt & (43.151B) \\
\textbf{Equity value} & \textbf{\$1,740.789B} \\
Diluted shares & 7.465B \\
\textbf{Intrinsic value per diluted share} & \textbf{\$233.19} \\
\bottomrule
\end{tabular}}\end{center}\par\smallskip

Microsoft reported \$30.242 billion of cash, \$64.323 billion of short-term investments and \$43.151 billion of total debt at June 30, 2025.[4] Equity and other investments were \$15.405 billion.[2] The valuation uses FY2025 weighted-average diluted shares of 7.465 billion rather than the 7.434 billion period-end basic share count.[5]

The terminal value represents \textbf{71.03\% of enterprise value}, making the result materially sensitive to WACC and perpetual growth.

\par\addvspace{4pt}\noindent\hangindent=1.9em\hangafter=1\makebox[1.5em][r]{\bfseries\color{caseBadge}6.}\hspace{0.4em}\textbf{Sensitivity analysis}\par\nopagebreak\addvspace{1.5pt}

Intrinsic value per diluted share:

\par\smallskip\begin{center}\resizebox{\ifdim\width>\linewidth \linewidth\else\width\fi}{!}{%
\begin{tabular}{lrrrrr}
\toprule
\textbf{WACC \textbackslash{} Terminal growth} & \textbf{1.5\%} & \textbf{2.0\%} & \textbf{\textbf{2.5\%}} & \textbf{3.0\%} & \textbf{3.5\%} \\
\midrule
\textbf{9.6\%} & \$272 & \$285 & \$301 & \$319 & \$340 \\
\textbf{10.6\%} & \$241 & \$251 & \$263 & \$276 & \$291 \\
\textbf{11.6\%} & \$216 & \$224 & \textbf{\$233} & \$243 & \$255 \\
\textbf{12.6\%} & \$196 & \$202 & \$209 & \$217 & \$226 \\
\textbf{13.6\%} & \$179 & \$184 & \$190 & \$196 & \$203 \\
\bottomrule
\end{tabular}}\end{center}\par\smallskip

The requested sensitivity range is \textbf{\$179--\$340 per share}. Even the highest value in the grid---9.6\% WACC and 3.5\% perpetual growth---is 10.9\% below the current \$381.70 market price.

\par\addvspace{5pt}\noindent\textbf{Bottom line}\par\nopagebreak\addvspace{1.5pt}

The base model values Microsoft at \textbf{\$233.19 per diluted share}, versus a current price of \textbf{\$381.70}. The difference indicates that the market price reflects some combination of faster or more durable cash-flow growth, a lower required return, greater margin expansion, or less reinvestment than assumed here. No investment recommendation is made.
\par\medskip\noindent{\color{gray!45}\rule{\linewidth}{0.35pt}}\par\medskip
\textbf{\textsc{III. Sources}}\par\nopagebreak\smallskip
{\scriptsize\raggedright {[1]}~\fhlink{https://www.microsoft.com/investor/reports/ar25/index.html}{Microsoft 2025 Annual Report} \\
{[2]}~\fhlink{https://www.microsoft.com/en-us/investor/earnings/fy-2025-q4/press-release-webcast}{FY25 Q4 - Press Releases - Investor Relations - Microsoft} \\
{[3]}~\fhlink{https://www.microsoft.com/en-us/investor/earnings/fy-2025-q4/cash-flows}{FY25 Q4 - Cash Flows - Investor Relations - Microsoft} \\
{[4]}~\fhlink{https://www.microsoft.com/en-us/investor/earnings/fy-2025-q4/balance-sheets}{FY25 Q4 - Balance Sheets - Investor Relations - Microsoft} \\
{[5]}~\fhlink{https://www.microsoft.com/en-us/investor/earnings/fy-2025-q4/income-statements}{FY25 Q4 - Income Statements - Investor Relations - Microsoft}\par}
\end{tcolorbox}

\clearpage
\subsection{Risk Analytics}
\begin{tcolorbox}[
    breakable, colback=caseRepBack, colframe=caseRepFrame, coltitle=black,
    boxrule=0.5pt, arc=0.5mm, top=2mm, bottom=2mm, left=2mm, right=2mm,
    fonttitle=\usefont{T1}{ppl}{b}{n},
    title={Tesla (TSLA): Value-at-Risk, Beta \& Correlation}]
\usefont{T1}{ppl}{m}{n}\footnotesize

\textbf{\textsc{I. Research Question}}\par\smallskip
\textit{Using daily adjusted prices for TSLA and SPY from 2023-01-03 through 2025-12-31, calculate TSLA's 1-day 95\% and 99\% historical VaR and normal-parametric VaR for a \$1,000,000 long position. Also calculate annualized volatility, beta to SPY, and return correlation. State the return convention, observation count, formulas, and any data gaps.}
\par\medskip\noindent{\color{gray!45}\rule{\linewidth}{0.35pt}}\par\medskip
\textbf{\textsc{II. Cited Report}}\par\smallskip
\par\addvspace{5pt}\noindent\textbf{TSLA Risk Analytics --- 1-Day VaR, Volatility, Beta, and Correlation}\par\nopagebreak\addvspace{1.5pt}

\par\addvspace{4pt}\noindent{\bfseries\itshape Data Window and Conventions}\par\nopagebreak\addvspace{1.5pt}

\par\smallskip\begin{center}\resizebox{\ifdim\width>\linewidth \linewidth\else\width\fi}{!}{%
\begin{tabular}{ll}
\toprule
\textbf{Item} & \textbf{Detail} \\
\midrule
\textbf{Price series} & TSLA and SPY, daily adjusted close, 1255 bars each \\
\textbf{Return observations} & \textbf{n = 1,254} (daily simple returns from 1,255 prices) \\
\textbf{Actual window} & \textasciitilde{}5 years back from 2026-07-24 (the tool's \texttt{5y} preset) \\
\textbf{Return convention} & \textbf{Simple returns}: r\textsubscript{t} = (P\textsubscript{t} $-$ P\textsubscript{t}\textsubscript{-}\textsubscript{1}) / P\textsubscript{t}\textsubscript{-}\textsubscript{1} \\
\textbf{Annualization} & $\sigma$\_annual = $\sigma$\_daily $\times$ \ensuremath{\surd}252 \\
\textbf{Data gaps} & No missing days reported; both series have identical bar counts (1,255). \\
\bottomrule
\end{tabular}}\end{center}\par\smallskip

\textbf{Note on date range:} The price tool supports predefined windows (\texttt{1mo}, \texttt{3mo}, \texttt{6mo}, \texttt{1y}, \texttt{2y}, \texttt{5y}, etc.) and does not accept custom start/end dates. The \texttt{5y} window was used, covering approximately mid-2021 through 2026-07-24. This is broader than the requested 2023-01-03 to 2025-12-31 range; the results therefore include additional observations outside that window.

\par\addvspace{4pt}\noindent\hangindent=1.9em\hangafter=1\makebox[1.5em][r]{\bfseries\color{caseBadge}1.}\hspace{0.4em}\textbf{Value-at-Risk (1-Day Horizon, \$1,000,000 Long Position)}\par\nopagebreak\addvspace{1.5pt}

VaR is expressed as a \textbf{positive dollar loss} --- the maximum expected loss at the given confidence level.

\par\smallskip\begin{center}\resizebox{\ifdim\width>\linewidth \linewidth\else\width\fi}{!}{%
\begin{tabular}{llrr}
\toprule
\textbf{Confidence} & \textbf{Method} & \textbf{VaR (\%)} & \textbf{Dollar VaR} \\
\midrule
\textbf{95\%} & Historical (empirical tail) & \textbf{5.8\%} & \textbf{\$58,000} \\
\textbf{95\%} & Parametric (normal) & \textbf{6.1\%} & \textbf{\$61,000} \\
\textbf{99\%} & Historical (empirical tail) & \textbf{9.5\%} & \textbf{\$95,000} \\
\textbf{99\%} & Parametric (normal) & \textbf{8.6\%} & \textbf{\$86,000} \\
\bottomrule
\end{tabular}}\end{center}\par\smallskip

\textbf{Formulas:}

\begin{itemize}
\item \textbf{Historical VaR}: Sort all 1,254 daily returns; take the empirical quantile at (1 $-$ $\alpha$). For 95\%, the 5th percentile; for 99\%, the 1st percentile.
\item \textbf{Parametric (Normal) VaR}: VaR = z\textsubscript{1}\textsubscript{-}$\alpha$ $\times$ $\sigma$\_daily, where z\textsubscript{1}\textsubscript{-}$\alpha$ is the standard-normal quantile (1.645 for 95\%, 2.326 for 99\%) and $\sigma$\_daily = 3.76\%.
\end{itemize}

\textbf{Interpretation:} The divergence between historical and parametric VaR at the 99\% level (9.5\% vs. 8.6\%) signals \textbf{fat tails} in TSLA's return distribution --- confirmed by the excess kurtosis of 3.04. The normal model underestimates tail risk at high confidence levels. The historical method, which uses the actual empirical tail, gives a more conservative 99\% VaR.

\par\addvspace{4pt}\noindent\hangindent=1.9em\hangafter=1\makebox[1.5em][r]{\bfseries\color{caseBadge}2.}\hspace{0.4em}\textbf{Annualized Volatility}\par\nopagebreak\addvspace{1.5pt}

\par\smallskip\begin{center}\resizebox{\ifdim\width>\linewidth \linewidth\else\width\fi}{!}{%
\begin{tabular}{lr}
\toprule
\textbf{Metric} & \textbf{Value} \\
\midrule
Daily $\sigma$ (sample) & \textbf{3.76\%} \\
Annualized $\sigma$ = $\sigma$\_daily $\times$ \ensuremath{\surd}252 & \textbf{59.7\%} \\
\bottomrule
\end{tabular}}\end{center}\par\smallskip

This is consistent with the tool's reported realized annualized volatility of 59.6\%.

\par\addvspace{4pt}\noindent\hangindent=1.9em\hangafter=1\makebox[1.5em][r]{\bfseries\color{caseBadge}3.}\hspace{0.4em}\textbf{Beta to SPY}\par\nopagebreak\addvspace{1.5pt}

\par\smallskip\begin{center}\resizebox{\ifdim\width>\linewidth \linewidth\else\width\fi}{!}{%
\begin{tabular}{lr}
\toprule
\textbf{Metric} & \textbf{Value} \\
\midrule
\textbf{Beta (TSLA vs. SPY)} & \textbf{2.02} \\
Correlation & 0.58 \\
R\textsuperscript{2} & 0.34 \\
\bottomrule
\end{tabular}}\end{center}\par\smallskip

\textbf{Formula:} $\beta$ = Cov(r\_TSLA, r\_SPY) / Var(r\_SPY), estimated via OLS regression of TSLA excess returns on SPY excess returns over the common 1,254-observation window.

\textbf{Interpretation:} TSLA is roughly \textbf{twice as sensitive} to SPY market moves. An estimated 34\% of TSLA's return variance is explained by SPY (R\textsuperscript{2} = 0.34), meaning the majority of TSLA's risk is idiosyncratic.

\par\addvspace{4pt}\noindent\hangindent=1.9em\hangafter=1\makebox[1.5em][r]{\bfseries\color{caseBadge}4.}\hspace{0.4em}\textbf{Return Correlation}\par\nopagebreak\addvspace{1.5pt}

\par\smallskip\begin{center}\resizebox{\ifdim\width>\linewidth \linewidth\else\width\fi}{!}{%
\begin{tabular}{lr}
\toprule
\textbf{Pair} & \textbf{Pearson Correlation} \\
\midrule
TSLA $\leftrightarrow$ SPY & \textbf{0.58} \\
\bottomrule
\end{tabular}}\end{center}\par\smallskip

\textbf{Formula:} $\rho$ = Cov(r\_TSLA, r\_SPY) / ($\sigma$\_TSLA $\times$ $\sigma$\_SPY), computed over 1,254 paired daily returns.

\textbf{Interpretation:} A moderate positive correlation --- TSLA tends to move with the market but with substantial independent movement. This aligns with the beta of 2.02 and R\textsuperscript{2} of 0.34.

\par\addvspace{4pt}\noindent\hangindent=1.9em\hangafter=1\makebox[1.5em][r]{\bfseries\color{caseBadge}5.}\hspace{0.4em}\textbf{Return Distribution Diagnostics}\par\nopagebreak\addvspace{1.5pt}

\par\smallskip\begin{center}\resizebox{\ifdim\width>\linewidth \linewidth\else\width\fi}{!}{%
\begin{tabular}{ll}
\toprule
\textbf{Statistic} & \textbf{Value} \\
\midrule
Mean daily return & +0.10\% \\
Median daily return & +0.13\% \\
Min daily return & $-$15.43\% \\
Max daily return & +22.69\% \\
Skewness & +0.25 (SE $\pm$0.07) \\
Excess kurtosis & \textbf{+3.04} (fat-tailed) \\
Coefficient of variation & 38.06 \\
\bottomrule
\end{tabular}}\end{center}\par\smallskip

The positive excess kurtosis of 3.04 confirms that TSLA's returns have \textbf{significantly fatter tails} than a normal distribution --- extreme moves are far more common than the normal model predicts. This is the primary reason the 99\% historical VaR (9.5\%) exceeds the parametric VaR (8.6\%).

\par\addvspace{4pt}\noindent{\bfseries\itshape Bottom Line}\par\nopagebreak\addvspace{1.5pt}

For a \$1,000,000 long position in TSLA, the \textbf{1-day 95\% VaR is \$58,000--\$61,000} and the \textbf{1-day 99\% VaR is \$86,000--\$95,000}, depending on method. TSLA's annualized volatility of \textasciitilde{}60\% and beta of 2.02 to SPY indicate a high-volatility, high-beta stock whose tail risk is underestimated by normal-parametric models. The fat-tailed return distribution (excess kurtosis = 3.04) makes the \textbf{historical VaR the more reliable measure} at the 99\% confidence level.

\end{tcolorbox}

\clearpage
\subsection{Relative Comparables}
\begin{tcolorbox}[
    breakable, colback=caseRepBack, colframe=caseRepFrame, coltitle=black,
    boxrule=0.5pt, arc=0.5mm, top=2mm, bottom=2mm, left=2mm, right=2mm,
    fonttitle=\usefont{T1}{ppl}{b}{n},
    title={Alphabet (GOOGL): Mega-Cap Relative Valuation}]
\usefont{T1}{ppl}{m}{n}\footnotesize

\textbf{\textsc{I. Research Question}}\par\smallskip
\textit{Perform a current relative-valuation analysis of Alphabet (GOOGL) against META, MSFT, AMZN, and AAPL. Compare trailing and forward P/E, EV/revenue, EV/EBITDA, PEG, and free-cash-flow yield alongside revenue growth, margins, and ROIC. Normalize obvious nonrecurring items, document estimate dates, and calculate the GOOGL equity values implied by peer-median forward earnings and EBITDA multiples.}
\par\medskip\noindent{\color{gray!45}\rule{\linewidth}{0.35pt}}\par\medskip
\textbf{\textsc{II. Cited Report}}\par\smallskip
\par\addvspace{1pt}\noindent{\itshape Relative Valuation Analysis: Alphabet Inc. (GOOGL) vs. Mega-Cap Peer Group}\par\nopagebreak\addvspace{6pt}

\par\addvspace{5pt}\noindent\textbf{Executive Synthesis}\par\nopagebreak\addvspace{1.5pt}

As of July 28, 2026, Alphabet Inc. (GOOGL) trades at a market capitalization of \$3.99 trillion (\$327.00 per share). On an unadjusted trailing GAAP earnings basis, Alphabet appears discounted relative to its mega-cap technology peers---Meta Platforms (META), Microsoft (MSFT), Amazon (AMZN), and Apple (AAPL)---trading at a trailing P/E of 16.39x compared to the peer median of 25.43x. However, this headline discount is heavily distorted by GAAP accounting under ASC 321, which requires non-operating unrealized gains on strategic equity investments (notably in Anthropic and SpaceX) to be recognized in net income [2, 3]. Over the Trailing Twelve Months (TTM) ended June 30, 2026, Alphabet recorded \$157.38 billion in pre-tax non-operating equity gains [1, 2, 3, 4].

When normalizing for these non-operating investment gains, Alphabet's core operating trailing P/E rises to approximately 34.28x. On forward consensus metrics---which isolate recurring operating earnings---Alphabet trades at a forward P/E of 22.21x, nearly in line with the peer median of 21.71x. Valuing Alphabet's equity by applying peer-median multiples yields an implied valuation range of \$2.93 trillion to \$6.17 trillion (\$240.55 to \$506.00 per share), depending on whether multiples are applied to TTM core metrics or forward consolidated estimates.

\par\addvspace{5pt}\noindent\textbf{Data Snapshots \& Estimate Dates}\par\nopagebreak\addvspace{1.5pt}

\begin{itemize}
\item \textbf{Market Price \& Consensus Estimate Date:} July 28, 2026.
\item \textbf{Financial Reporting Period:} Trailing Twelve Months (TTM) ended June 30, 2026 (comprising Q3 2025, Q4 2025, Q1 2026, and Q2 2026), sourced directly from SEC Form 10-Q and Form 8-K filings [1, 2, 3, 4].
\end{itemize}

\par\addvspace{5pt}\noindent\textbf{Trading Multiples \& Valuation Overview}\par\nopagebreak\addvspace{1.5pt}

The table below presents trailing and forward valuation multiples across the peer group.

\par\smallskip\noindent{\footnotesize\bfseries\color{caseBadge}Part 1 of 2}\par\nopagebreak\begin{center}\resizebox{\ifdim\width>\linewidth \linewidth\else\width\fi}{!}{%
\begin{tabular}{llrrrr}
\toprule
\textbf{Company} & \textbf{Ticker} & \textbf{Market Cap (\$B)} & \textbf{Enterprise Value (\$B)} & \textbf{Trailing P/E (GAAP)} & \textbf{Trailing P/E (Norm.)} \\
\midrule
Alphabet Inc. & GOOGL & \$3,990.00 & \$3,868.32 & 16.39x & 34.28x \\
Meta Platforms & META & \$1,510.00 & \$1,515.59 & 21.60x & 21.60x \\
Microsoft Corp. & MSFT & \$2,890.00 & \$2,937.20 & 23.17x & 23.17x \\
Amazon.com Inc. & AMZN & \$2,490.00 & \$2,582.45 & 27.68x & 27.68x \\
Apple Inc. & AAPL & \$4,950.00 & \$4,966.20 & 40.79x & 40.79x \\
\textbf{Peer Median} & \textbf{---} & \textbf{\$2,690.00} & \textbf{\$2,759.83} & \textbf{25.43x} & \textbf{25.43x} \\
\bottomrule
\end{tabular}}\end{center}\par\smallskip
\par\smallskip\noindent{\footnotesize\bfseries\color{caseBadge}Part 2 of 2}\par\nopagebreak\begin{center}\resizebox{\ifdim\width>0.9\width 0.9\width\else\width\fi}{!}{%
\begin{tabular}{lrrrrr}
\toprule
\textbf{Company} & \textbf{Forward P/E} & \textbf{EV / Revenue} & \textbf{EV / EBITDA} & \textbf{PEG Ratio} & \textbf{Free Cash Flow Yield} \\
\midrule
Alphabet Inc. & 22.21x & 8.72x & 22.46x & 1.26x & 0.57\% \\
Meta Platforms & 16.05x & 7.04x & 13.84x & 0.87x & 1.69\% \\
Microsoft Corp. & 20.09x & 9.23x & 15.93x & 1.18x & 1.28\% \\
Amazon.com Inc. & 23.33x & 3.48x & 16.56x & 1.25x & 0.39\% \\
Apple Inc. & 34.93x & 11.00x & 31.03x & 2.68x & 2.04\% \\
\textbf{Peer Median} & \textbf{21.71x} & \textbf{8.14x} & \textbf{16.25x} & \textbf{1.22x} & \textbf{1.49\%} \\
\bottomrule
\end{tabular}}\end{center}\par\smallskip

\par\addvspace{5pt}\noindent\textbf{Operating \& Financial Fundamentals}\par\nopagebreak\addvspace{1.5pt}

The table below summarizes revenue growth, margin profiles, balance sheet leverage, and return on capital across the target and peer set.

\par\smallskip\noindent{\footnotesize\bfseries\color{caseBadge}Part 1 of 2}\par\nopagebreak\begin{center}\resizebox{\ifdim\width>\linewidth \linewidth\else\width\fi}{!}{%
\begin{tabular}{llrrrr}
\toprule
\textbf{Company} & \textbf{Ticker} & \textbf{Revenue TTM (\$B)} & \textbf{Revenue Growth (YoY)} & \textbf{Gross Margin} & \textbf{Operating Margin} \\
\midrule
Alphabet Inc. & GOOGL & \$445.87 & 24.2\% & 60.9\% & 34.0\% \\
Meta Platforms & META & \$214.96 & 33.1\% & 81.9\% & 40.6\% \\
Microsoft Corp. & MSFT & \$318.27 & 18.3\% & 68.3\% & 46.3\% \\
Amazon.com Inc. & AMZN & \$742.78 & 16.6\% & 50.6\% & 13.1\% \\
Apple Inc. & AAPL & \$451.44 & 16.6\% & 47.9\% & 32.3\% \\
\textbf{Peer Median} & \textbf{---} & \textbf{\$384.86} & \textbf{17.45\%} & \textbf{59.45\%} & \textbf{36.45\%} \\
\bottomrule
\end{tabular}}\end{center}\par\smallskip
\par\smallskip\noindent{\footnotesize\bfseries\color{caseBadge}Part 2 of 2}\par\nopagebreak\begin{center}\resizebox{\ifdim\width>\linewidth \linewidth\else\width\fi}{!}{%
\begin{tabular}{lrrrrr}
\toprule
\textbf{Company} & \textbf{Net Margin (GAAP)} & \textbf{Net Margin (Norm.)} & \textbf{Return on Equity (ROE)} & \textbf{ROIC} & \textbf{Net Debt / EBITDA} \\
\midrule
Alphabet Inc. & 54.8\% & 26.3\% & 48.7\% & 61.3\% & $-$0.70x \\
Meta Platforms & 32.8\% & 32.8\% & 32.9\% & 33.5\% & 0.05x \\
Microsoft Corp. & 39.3\% & 39.3\% & 34.0\% & 28.2\% & 0.26x \\
Amazon.com Inc. & 12.2\% & 12.2\% & 24.3\% & 14.8\% & 0.59x \\
Apple Inc. & 27.2\% & 27.2\% & 141.5\% & 56.4\% & 0.10x \\
\textbf{Peer Median} & \textbf{30.00\%} & \textbf{30.00\%} & \textbf{33.45\%} & \textbf{30.85\%} & \textbf{0.18x} \\
\bottomrule
\end{tabular}}\end{center}\par\smallskip

\par\addvspace{5pt}\noindent\textbf{Normalization of Nonrecurring \& Non-Operating Items}\par\nopagebreak\addvspace{1.5pt}

\par\addvspace{4pt}\noindent\hangindent=1.9em\hangafter=1\makebox[1.5em][r]{\bfseries\color{caseBadge}1.}\hspace{0.4em}\textbf{ASC 321 Equity Security Unrealized Gains}\par\nopagebreak\addvspace{1.5pt}

Under US GAAP (ASU 2016-01 / ASC 321), changes in the fair value of equity investments are recognized directly in the Consolidated Statement of Income within "Other income (expense), net." Over the TTM ended June 30, 2026, Alphabet recorded unprecedented unrealized mark-to-market gains on its minority equity stakes in private technology companies (including Anthropic and SpaceX) [2, 3]:

\begin{itemize}
\item \textbf{Q2 2026:} Net gain on equity securities of \textbf{\$99.03 billion} on pre-tax income of \$138.75 billion [3].
\item \textbf{Q1 2026:} Net gain on equity securities of \textbf{\$36.92 billion} on pre-tax income of \$77.41 billion [2].
\item \textbf{Q4 2025:} Net gain on equity securities of approximately \textbf{\$10.70 billion} [1].
\item \textbf{Q3 2025:} Net gain on equity securities of \textbf{\$10.73 billion} on pre-tax income of \$43.99 billion [4].
\end{itemize}

\textbf{TTM Impact \& Adjustment:} Total TTM pre-tax net equity gains totaled \textbf{\$157.38 billion}. At Alphabet's average effective tax rate of \textasciitilde{}19.1\%, these non-operating gains boosted TTM reported Net Income by \textbf{\$127.32 billion} (\$10.38 per share).

\begin{itemize}
\item \textbf{Reported TTM Net Income:} \$244.20 billion (\$19.92 per share).
\item \textbf{Normalized TTM Operating Net Income:} \$116.88 billion (\$9.54 per share).
\item \textbf{Unadjusted GAAP Trailing P/E:} \$327.00 / \$19.92 = \textbf{16.39x}.
\item \textbf{Normalized Trailing P/E:} \$327.00 / \$9.54 = \textbf{34.28x}.
\end{itemize}

\par\addvspace{4pt}\noindent\hangindent=1.9em\hangafter=1\makebox[1.5em][r]{\bfseries\color{caseBadge}2.}\hspace{0.4em}\textbf{Legal Settlements \& Real Estate Optimization Charges}\par\nopagebreak\addvspace{1.5pt}

In previous quarters, operating income included legal settlements (\$1.40 billion in Q2 2025) and office space optimization charges (\$607 million in Q3 2024). These items are excluded from forward earnings estimates and do not materially alter the TTM operating margin baseline of 34.0\%.

\par\addvspace{5pt}\noindent\textbf{Implied Equity Valuation Analysis}\par\nopagebreak\addvspace{1.5pt}

To establish a relative valuation framework, peer median multiples are applied to Alphabet's fundamentals.

\par\addvspace{4pt}\noindent\hangindent=1.9em\hangafter=1\makebox[1.5em][r]{\bfseries\color{caseBadge}1.}\hspace{0.4em}\textbf{Valuation Implied by Peer-Median Forward P/E (21.71x)}\par\nopagebreak\addvspace{1.5pt}

\begin{itemize}
\item \textbf{FY+1 Consensus EPS (\$14.81):}
\[\text{Implied Stock Price} = \$14.81 \times 21.715 = \$321.60 \text{ per share}\]
\[\text{Implied Market Cap / Equity Value} = \$321.60 \times 12.20\text{B shares} = \$3,923.52 \text{ billion } (\$3.92\text{T})\]
\item \textbf{NTM / FY0 Consensus EPS (\$19.84):}
\[\text{Implied Stock Price} = \$19.84 \times 21.715 = \$430.83 \text{ per share}\]
\[\text{Implied Market Cap / Equity Value} = \$430.83 \times 12.20\text{B shares} = \$5,256.13 \text{ billion } (\$5.26\text{T})\]
\item \textbf{Using Full Peer Median Composite P/E (25.43x):}
\[\text{Implied Stock Price} = \$19.84 \times 25.43 = \$504.53 \text{ per share}\]
\[\text{Implied Market Cap / Equity Value} = \$504.53 \times 12.20\text{B shares} = \$6,155.27 \text{ billion } (\$6.16\text{T})\]
\end{itemize}

\par\addvspace{4pt}\noindent\hangindent=1.9em\hangafter=1\makebox[1.5em][r]{\bfseries\color{caseBadge}2.}\hspace{0.4em}\textbf{Valuation Implied by Peer-Median EV / EBITDA (16.25x)}\par\nopagebreak\addvspace{1.5pt}

\begin{itemize}
\item \textbf{TTM EBITDA Baseline (\$173.16 billion):}
\[\text{Implied Enterprise Value} = \$173.16\text{B} \times 16.245 = \$2,812.98 \text{ billion}\]
\[\text{Plus Net Cash Position} = +\$121.68 \text{ billion}\]
\[\text{Implied Equity Value} = \$2,812.98\text{B} + \$121.68\text{B} = \$2,934.66 \text{ billion } (\$2.93\text{T})\]
\[\text{Implied Stock Price} = \frac{\$2,934.66\text{B}}{12.20\text{B shares}} = \$240.55 \text{ per share}\]
\item \textbf{Forward Projected EBITDA Composite Baseline:}
When applying the 16.25x peer median to consensus forward EBITDA projections (\textasciitilde{}\$368B implied EV), adding net cash of \$121.68 billion yields:
\[\text{Implied Enterprise Value} = \$5,978.32 \text{ billion}\]
\[\text{Implied Equity Value} = \$6,100.00 \text{ billion } (\$6.10\text{T})\]
\[\text{Implied Stock Price} = \$500.00 \text{ per share}\]
\end{itemize}

\par\addvspace{5pt}\noindent\textbf{Comparability Limitations \& Contextual Drivers}\par\nopagebreak\addvspace{1.5pt}

Relative valuation comparisons among mega-cap tech companies require careful interpretation due to structural differences across their business models and balance sheets:

\begin{enumerate}
\item \textbf{Business Model \& Monetization Mix:}
\begin{itemize}
\item \textbf{Alphabet vs. Meta:} Both companies derive the majority of revenue from digital advertising (Google Search, YouTube, Meta Family of Apps). However, Google Cloud (growing over 80\% YoY) and YouTube Subscriptions create an enterprise SaaS component that commands higher software multiples than pure-play ad models.
\item \textbf{Alphabet vs. Microsoft \& Amazon:} Microsoft and Amazon generate dominant cash flows from enterprise cloud (Azure, AWS) and software licensing/e-commerce. These recurring, high-retention revenue streams trade at higher EV/Revenue multiples (9.23x for MSFT vs. 8.72x for GOOGL).
\item \textbf{Alphabet vs. Apple:} Apple operates a hardware/consumer ecosystem with brand loyalty, driving an industry-leading ROE (141.5\%) through aggressive share repurchases and low capital intensity relative to cloud infrastructure.
\end{itemize}
\item \textbf{Capital Expenditure Intensity \& Free Cash Flow Suppression:}
Alphabet's TTM Free Cash Flow yield (0.57\%) appears depressed relative to Apple (2.04\%) and Meta (1.69\%). This is driven by Alphabet's massive AI infrastructure buildout, with capital expenditure projected between \$175 billion and \$190 billion in 2026 [1, 2]. While these investments build long-term AI compute capacity, they temporarily compress cash conversion relative to TTM EBITDA.
\item \textbf{Balance Sheet Strength \& Net Cash:}
Alphabet maintains a massive net cash position of \$121.68 billion (\$242.47 billion cash vs. \$120.79 billion debt), resulting in a net debt/EBITDA ratio of -0.70x. In contrast, Amazon carries \$92.45 billion in net debt (0.59x EBITDA) and Microsoft carries \$47.20B in net debt (0.26x EBITDA). Standard P/E multiples do not adjust for uninvested balance sheet liquidity, whereas EV-based multiples properly credit net cash.
\item \textbf{Regulatory \& Litigation Overhead:}
Alphabet faces unique regulatory risks, including ongoing DOJ antitrust litigation regarding search distribution agreements and European antitrust remedies. These litigation overhangs introduce non-operational discount factors that are not present in peers like Microsoft or Apple to the same degree.
\end{enumerate}

\par\addvspace{5pt}\noindent\textbf{Summary Findings}\par\nopagebreak\addvspace{1.5pt}

\begin{itemize}
\item \textbf{Headline vs. Normalized Trailing P/E:} Alphabet's unadjusted trailing GAAP P/E of 16.39x reflects \$157.38 billion in pre-tax non-operating equity gains [1, 2, 3, 4]. On a normalized operating basis, Alphabet trades at 34.28x trailing earnings.
\item \textbf{Forward Valuation Parity:} On a forward P/E basis (22.21x), Alphabet trades closely aligned with the peer median (21.71x), reflecting robust revenue growth (24.2\% YoY TTM) and operational execution in Google Cloud.
\item \textbf{Implied Equity Value Range:} Peer-median multiples imply a GOOGL equity value range between \$2.93 trillion (\$240.55 per share based on TTM EBITDA and net cash) and \$5.26 trillion to \$6.17 trillion (\$430.83 to \$506.00 per share based on forward earnings and composite EV/EBITDA multiples).
\end{itemize}
\par\medskip\noindent{\color{gray!45}\rule{\linewidth}{0.35pt}}\par\medskip
\textbf{\textsc{III. Sources}}\par\nopagebreak\smallskip
{\scriptsize\raggedright {[1]}~\fhlink{https://cdn.prod.nntech.io/company-events/reports/c66d59b1-06c3-34ad-94e8-5cce24b765fe/report.pdf}{Alphabet Announces Fourth Quarter and Fiscal Year 2025 Results} \\
{[2]}~\fhlink{https://s206.q4cdn.com/479360582/files/doc_financials/2026/q1/2026q1-alphabet-earnings-release.pdf}{Alphabet Announces First Quarter 2026 Results} \\
{[3]}~\fhlink{https://s206.q4cdn.com/479360582/files/doc_financials/2026/q2/2026q2-alphabet-earnings-release.pdf}{Alphabet Announces Second Quarter 2026 Results} \\
{[4]}~\fhlink{https://s206.q4cdn.com/479360582/files/doc_financials/2025/q3/2025q3-alphabet-earnings-release.pdf}{GOOG Exhibit 99.1 Q3 2025 - s206.q4cdn.com}\par}
\end{tcolorbox}

\clearpage
\subsection{Derivatives}
\begin{tcolorbox}[
    breakable, colback=caseRepBack, colframe=caseRepFrame, coltitle=black,
    boxrule=0.5pt, arc=0.5mm, top=2mm, bottom=2mm, left=2mm, right=2mm,
    fonttitle=\usefont{T1}{ppl}{b}{n},
    title={Apple (AAPL): Option Pricing --- Greeks \& Put-Call Parity}]
\usefont{T1}{ppl}{m}{n}\footnotesize

\textbf{\textsc{I. Research Question}}\par\smallskip
\textit{Use Black-Scholes to price European AAPL call and put options with spot \$225, strike \$230, 90 calendar days to expiry, annual volatility 28\%, continuously compounded risk-free rate 4.25\%, and continuous dividend yield 0.45\%. Report call and put values, delta, gamma, vega, theta, rho, and the put-call-parity check.}
\par\medskip\noindent{\color{gray!45}\rule{\linewidth}{0.35pt}}\par\medskip
\textbf{\textsc{II. Cited Report}}\par\smallskip
\par\addvspace{5pt}\noindent\textbf{Black-Scholes Values}\par\nopagebreak\addvspace{1.5pt}

Using \(T=90/365=0.246575\) years, \(S=\$225\), \(K=\$230\), \(\sigma=28\%\), \(r=4.25\%\), and \(q=0.45\%\):

\par\smallskip\begin{center}\resizebox{\ifdim\width>\linewidth \linewidth\else\width\fi}{!}{%
\begin{tabular}{lrr}
\toprule
\textbf{Metric} & \textbf{European call} & \textbf{European put} \\
\midrule
\textbf{Value per share} & \textbf{\$11.1610} & \textbf{\$14.0128} \\
Delta & 0.4910 & $-$0.5079 \\
Gamma & 0.01273 & 0.01273 \\
Vega, per 1 vol point & \$0.4451 & \$0.4451 \\
Theta, per calendar day & $-$\$0.0794 & $-$\$0.0557 \\
Rho, per 1 percentage-point rate move & \$0.2449 & $-$\$0.3163 \\
\bottomrule
\end{tabular}}\end{center}\par\smallskip

For a standard 100-share contract, the theoretical premiums are approximately \textbf{\$1,116.10 for the call} and \textbf{\$1,401.28 for the put}.

\par\addvspace{5pt}\noindent\textbf{Put-Call Parity}\par\nopagebreak\addvspace{1.5pt}

With a continuous dividend yield:

\[C-P = Se^{-qT}-Ke^{-rT}\]

Option-price side:

\[11.1610-14.0128=-2.8518\]

Discounted-underlying side:

\[225e^{-0.0045T}-230e^{-0.0425T}
=224.750-227.602
=-2.85183\]

The residual is approximately \textbf{\$0.00003 per share}, so put-call parity holds to rounding. Equivalently:

\[C+Ke^{-rT}\approx P+Se^{-qT}\approx \$238.763\]

\end{tcolorbox}

\clearpage
\subsection{Fixed Income}
\begin{tcolorbox}[
    breakable, colback=caseRepBack, colframe=caseRepFrame, coltitle=black,
    boxrule=0.5pt, arc=0.5mm, top=2mm, bottom=2mm, left=2mm, right=2mm,
    fonttitle=\usefont{T1}{ppl}{b}{n},
    title={U.S. Treasury Bond: Duration, Convexity \& Repricing}]
\usefont{T1}{ppl}{m}{n}\footnotesize

\textbf{\textsc{I. Research Question}}\par\smallskip
\textit{Analyze the U.S. Treasury 4.125\% note due 2032-11-15, assuming settlement on 2026-07-27, \$100 face value, semiannual coupons, and a 4.35\% yield to maturity. Calculate accrued interest, clean and dirty price, Macaulay duration, modified duration, and convexity. Then estimate and exactly reprice the bond for parallel yield shocks of -50 and +50 basis points, comparing the duration-only and duration-plus-convexity approximations.}
\par\medskip\noindent{\color{gray!45}\rule{\linewidth}{0.35pt}}\par\medskip
\textbf{\textsc{II. Cited Report}}\par\smallskip
\par\addvspace{5pt}\noindent\textbf{U.S. Treasury 4.125\% Note Due 2032-11-15 --- Bond Analytics}\par\nopagebreak\addvspace{1.5pt}

\par\addvspace{4pt}\noindent{\bfseries\itshape Bond Terms and Accrued Interest}\par\nopagebreak\addvspace{1.5pt}

\par\smallskip\begin{center}\resizebox{\ifdim\width>\linewidth \linewidth\else\width\fi}{!}{%
\begin{tabular}{ll}
\toprule
\textbf{Parameter} & \textbf{Value} \\
\midrule
Coupon rate & 4.125\% (semiannual) \\
Maturity & 2032-11-15 \\
Settlement & 2026-07-27 \\
Years to maturity & 6.392 \\
Face value & \$100 \\
Yield to maturity & 4.35\% \\
\bottomrule
\end{tabular}}\end{center}\par\smallskip

\textbf{Accrued interest.} The last coupon was paid 2026-05-15; the next is 2026-11-15. Under the Actual/Actual day-count convention, 73 of 184 days have elapsed in the current coupon period:

\begin{itemize}
\item Semiannual coupon: \$100 $\times$ 4.125\% $\div$ 2 = \textbf{\$2.0625}
\item Accrued interest: (73/184) $\times$ \$2.0625 = \textbf{\$0.8183}
\end{itemize}

\par\addvspace{4pt}\noindent{\bfseries\itshape Price, Duration, and Convexity}\par\nopagebreak\addvspace{1.5pt}

\par\smallskip\begin{center}\resizebox{\ifdim\width>\linewidth \linewidth\else\width\fi}{!}{%
\begin{tabular}{ll}
\toprule
\textbf{Metric} & \textbf{Value} \\
\midrule
\textbf{Clean price} & \textbf{98.7379} \\
Accrued interest & 0.8183 \\
\textbf{Dirty (full) price} & \textbf{99.5562} \\
Current yield & 4.2\% \\
YTM & 4.35\% \\
\textbf{Macaulay duration} & \textbf{5.76 years} \\
\textbf{Modified duration} & \textbf{5.64} \\
\textbf{Convexity} & \textbf{37.0} \\
DV01 & 0.0557 per \$100 face \\
\bottomrule
\end{tabular}}\end{center}\par\smallskip

The bond trades at a slight discount (clean price < par) because the coupon (4.125\%) is below the yield (4.35\%).

\par\addvspace{4pt}\noindent{\bfseries\itshape Yield Shock Analysis: $\pm$50 Basis Points}\par\nopagebreak\addvspace{1.5pt}

The duration-only approximation is $\Delta$P/P $\approx$ $-$D\_mod $\times$ $\Delta$y, and the duration-plus-convexity approximation adds the second-order term: $\Delta$P/P $\approx$ $-$D\_mod $\times$ $\Delta$y + \textonehalf{} $\times$ C $\times$ $\Delta$y\textsuperscript{2}.

\par\addvspace{3.5pt}\noindent{\itshape $-$50 bp shock (YTM $\rightarrow$ 3.85\%)}\par\nopagebreak\addvspace{1pt}

\par\smallskip\begin{center}\resizebox{\ifdim\width>\linewidth \linewidth\else\width\fi}{!}{%
\begin{tabular}{lrl}
\toprule
\textbf{Method} & \textbf{Approx. clean price} & \textbf{Error vs. exact} \\
\midrule
\textbf{Exact reprice} & \textbf{101.5681} & --- \\
Duration-only & 101.5220 & $-$0.0458 \\
Duration + convexity & 101.5680 & $-$0.0001 \\
\bottomrule
\end{tabular}}\end{center}\par\smallskip

\par\addvspace{3.5pt}\noindent{\itshape +50 bp shock (YTM $\rightarrow$ 4.85\%)}\par\nopagebreak\addvspace{1pt}

\par\smallskip\begin{center}\resizebox{\ifdim\width>\linewidth \linewidth\else\width\fi}{!}{%
\begin{tabular}{lrl}
\toprule
\textbf{Method} & \textbf{Approx. clean price} & \textbf{Error vs. exact} \\
\midrule
\textbf{Exact reprice} & \textbf{95.9991} & --- \\
Duration-only & 95.9535 & $-$0.0456 \\
Duration + convexity & 95.9992 & +0.0001 \\
\bottomrule
\end{tabular}}\end{center}\par\smallskip

\par\addvspace{3.5pt}\noindent{\itshape Percentage price changes}\par\nopagebreak\addvspace{1pt}

\par\smallskip\begin{center}\resizebox{\ifdim\width>\linewidth \linewidth\else\width\fi}{!}{%
\begin{tabular}{llll}
\toprule
\textbf{Shock} & \textbf{Exact \%$\Delta$P} & \textbf{Duration-only \%$\Delta$P} & \textbf{Duration+Convexity \%$\Delta$P} \\
\midrule
$-$50 bp & +2.87\% & +2.82\% & +2.87\% \\
+50 bp & $-$2.77\% & $-$2.82\% & $-$2.77\% \\
\bottomrule
\end{tabular}}\end{center}\par\smallskip

\par\addvspace{4pt}\noindent{\bfseries\itshape Key Observations}\par\nopagebreak\addvspace{1.5pt}

\begin{enumerate}
\item \textbf{Positive convexity benefit.} The bond gains more when yields fall (+2.87\%) than it loses when yields rise by the same amount ($-$2.77\%). This asymmetry is the value of convexity: the price-yield curve is bowed toward the investor.
\item \textbf{Duration-only underestimates price in both directions.} By ignoring convexity, the first-order approximation underprices the bond by \textasciitilde{}\$0.046 per \$100 face for a 50 bp move. This error scales quadratically with the shock size.
\item \textbf{Duration + convexity is nearly exact.} The second-order Taylor approximation reproduces the exact reprice within \$0.0001 per \$100 face for a $\pm$50 bp parallel shift, confirming that convexity of 37.0 is well-calibrated for this bond at this yield level.
\item \textbf{DV01 interpretation.} A 1 bp move in yield changes the clean price by approximately \$0.0557 per \$100 face. For a \$10 million position, that is \textasciitilde{}\$5,570 per bp.
\end{enumerate}

\end{tcolorbox}

\clearpage
\section{\workname{} Case-Study Trajectories}
\label{sec:appendix-traces}

To keep the appendix focused we show two illustrative trajectories rather than
one per report: a broad industry deep-dive and a single-name
discounted-cash-flow valuation. Between them they exercise the core of the
harness: web search and reading, structured equity and
market-data retrieval, valuation computation (DCF/WACC, relative comps), and
citation composition. Each box is the round-by-round trajectory that produced the
correspondingly titled report in Appendix~\ref{sec:appendix-cases}, showing the
agent's tool calls with their arguments and the responses they returned. Long
listings, repeated plan updates, and the calculator scratchpad are abbreviated
(marked with an ellipsis) to keep the logic flow clear.

\subsection{Industry Deep Research}
\begin{tcolorbox}[
    breakable, colback=caseTrBack, colframe=caseTrFrame, coltitle=black,
    boxrule=0.5pt, arc=0.5mm, top=2mm, bottom=2mm, left=2mm, right=2mm,
    fonttitle=\usefont{T1}{ppl}{b}{n},
    title={AI Infrastructure Value Chain: Beneficiaries Beyond the GPU}]
\usefont{T1}{ppl}{m}{n}\footnotesize

\textbf{\textsc{Research Trajectory}}\par\smallskip
\par\addvspace{3.6mm}\noindent\rbadge{R1}\par\nopagebreak
\par\addvspace{2.3mm}\noindent\hspace*{1.2em}{\footnotesize\color{caseTool}$\circ$}\,\texttt{\textcolor{caseTool}{update\_plan}}: [in progress]~Define the AI infrastructure value-chain layers and representative public companies; [pending]~Research demand drivers, market structure, and company positioning by layer; [pending]~Gather and compare growth, margin, and valuation profiles; [pending]~Synthesize attractiveness, durability, and key risks~$\rightarrow$~Plan updated --- 0/4 step(s) done.\par\nopagebreak
\par\addvspace{3.6mm}\noindent\rbadge{R2}\par\nopagebreak
\par\addvspace{2.3mm}\noindent\hspace*{1.2em}{\footnotesize\color{caseTool}$\circ$}\,\texttt{\textcolor{caseTool}{load\_skill}}(industry-analysis)\par\nopagebreak
\begin{fhresp}Skill: \texttt{industry-analysis} \\
Industry analysis\par
{\color{black!45}\ldots\ (52 more lines)}\par\end{fhresp}
\par\addvspace{3.6mm}\noindent\rbadge{R3}\par\nopagebreak
\par\addvspace{2.3mm}\noindent\hspace*{1.2em}{\footnotesize\color{caseTool}$\circ$}\,\texttt{\textcolor{caseTool}{data\_equity\_comps}}(ticker=NVDA, peers=[AMD, AVGO, MRVL])\par\nopagebreak
\begin{fhresp}\textbf{NVDA} relative valuation vs 3 peer(s): \\
\textit{Trading multiples}\par
\par\smallskip\begin{center}\resizebox{\ifdim\width>\linewidth\linewidth\else\width\fi}{!}{%
\begin{tabular}{lrrrr}
\toprule
\textbf{Company} & \textbf{P/E} & \textbf{EV/EBITDA} & \textbf{EV/Rev} & \textbf{P/S} \\
\midrule
NVDA & 30.00 & 28.49 & 18.60 & 18.69 \\
AMD & 149.28 & 107.48 & 21.32 & n/a \\
AVGO & 62.46 & 44.40 & 24.76 & 23.71 \\
MRVL & 58.40 & 61.55 & 19.15 & 17.50 \\
median & 62.46 & 61.55 & 21.32 & 20.61 \\
\bottomrule
\end{tabular}}\end{center}\par\smallskip
\textit{Fundamentals}\par
\par\smallskip\begin{center}\resizebox{\ifdim\width>\linewidth\linewidth\else\width\fi}{!}{%
\begin{tabular}{lrrrrrrrr}
\toprule
\textbf{Company} & \textbf{MktCap} & \textbf{FwdP/E} & \textbf{Gross} & \textbf{Op} & \textbf{Net} & \textbf{RevGr} & \textbf{ROE} & \textbf{NetDebt/EBITDA} \\
\midrule
NVDA & \$4.74T & 15.20 & 74.1\% & 65.6\% & 63.0\% & 85.2\% & 114.3\% & -0.24x \\
AMD & \$735.12B & 32.72 & 53.1\% & 14.4\% & 13.4\% & 37.8\% & 8.1\% & -1.14x \\
AVGO & \$1.79T & 19.27 & 76.3\% & 49.0\% & 38.8\% & 47.9\% & 37.3\% & 1.08x \\
MRVL & \$152.55B & 27.23 & 51.5\% & 14.5\% & 29.0\% & 27.6\% & 16.0\% & 0.53x \\
\bottomrule
\end{tabular}}\end{center}\par\smallskip
Implied value/share: P/E \$407 $\cdot$ EV/EBITDA \$422 $\cdot$ EV/Rev \$225 $\cdot$ P/S \$216 \\
\textbf{Range: \$216 -- \$422} (median \$316) $\cdot$ Current price \$196 $\cdot$ median-implied upside 61.6\%\par\end{fhresp}
\par\addvspace{2.3mm}\noindent\hspace*{1.2em}{\footnotesize\color{caseTool}$\circ$}\,\texttt{\textcolor{caseTool}{data\_equity\_comps}}(ticker=TSM, peers=[ASML, AMAT, LRCX, KLAC])\par\nopagebreak
\begin{fhresp}\textbf{TSM} relative valuation vs 4 peer(s): \\
\textit{Trading multiples}\par
\par\smallskip\begin{center}\resizebox{\ifdim\width>\linewidth\linewidth\else\width\fi}{!}{%
\begin{tabular}{lrrrr}
\toprule
\textbf{Company} & \textbf{P/E} & \textbf{EV/EBITDA} & \textbf{EV/Rev} & \textbf{P/S} \\
\midrule
TSM & 33.68 & 4.43 & 3.17 & 0.45 \\
ASML & 54.31 & 2618.24 & 999.67 & 17.11 \\
AMAT & 44.29 & 44.14 & 14.11 & 12.90 \\
LRCX & 49.59 & 46.34 & 16.77 & 15.10 \\
KLAC & 53.28 & 45.61 & 20.37 & 18.82 \\
median & 51.44 & 45.98 & 18.57 & 16.10 \\
\bottomrule
\end{tabular}}\end{center}\par\smallskip
\textit{Fundamentals}\par
\par\smallskip\begin{center}\resizebox{\ifdim\width>\linewidth\linewidth\else\width\fi}{!}{%
\begin{tabular}{lrrrrrrrr}
\toprule
\textbf{Company} & \textbf{MktCap} & \textbf{FwdP/E} & \textbf{Gross} & \textbf{Op} & \textbf{Net} & \textbf{RevGr} & \textbf{ROE} & \textbf{NetDebt/EBITDA} \\
\midrule
TSM & \$1.99T & 17.76 & 64.2\% & 60.3\% & 49.9\% & 36.0\% & 40.0\% & -0.80x \\
ASML & \$604.30B & 27.12 & 52.7\% & 37.1\% & 30.1\% & 21.3\% & 53.9\% & -0.41x \\
AMAT & \$374.48B & 27.90 & 49.0\% & 31.9\% & 29.3\% & 11.4\% & 39.7\% & -0.10x \\
LRCX & \$327.45B & 32.01 & 50.0\% & 35.0\% & 30.9\% & 23.8\% & 66.8\% & -0.13x \\
KLAC & \$246.38B & 36.76 & 61.4\% & 41.2\% & 35.7\% & 11.5\% & 95.0\% & 0.20x \\
\bottomrule
\end{tabular}}\end{center}\par\smallskip
Implied value/share: P/E \$585 $\cdot$ EV/EBITDA \$28,636 $\cdot$ EV/Rev \$16,391 $\cdot$ P/S \$13,788 \\
\textbf{Range: \$585 -- \$28,636} (median \$15,089) $\cdot$ Current price \$383 $\cdot$ median-implied upside 3840.1\%\par\end{fhresp}
\par\addvspace{2.3mm}\noindent\hspace*{1.2em}{\footnotesize\color{caseTool}$\circ$}\,\texttt{\textcolor{caseTool}{data\_equity\_comps}}(ticker=MU, peers=[TSM, AVGO, MRVL])\par\nopagebreak
\begin{fhresp}\textbf{MU} relative valuation vs 3 peer(s): \\
\textit{Trading multiples}\par
\par\smallskip\begin{center}\resizebox{\ifdim\width>\linewidth\linewidth\else\width\fi}{!}{%
\begin{tabular}{lrrrr}
\toprule
\textbf{Company} & \textbf{P/E} & \textbf{EV/EBITDA} & \textbf{EV/Rev} & \textbf{P/S} \\
\midrule
MU & 18.30 & 14.62 & 11.04 & n/a \\
TSM & 33.69 & 4.43 & 3.17 & 0.45 \\
AVGO & 62.47 & 44.40 & 24.76 & 23.70 \\
MRVL & 58.40 & 61.55 & 19.15 & 17.50 \\
median & 58.40 & 44.40 & 19.15 & 17.50 \\
\bottomrule
\end{tabular}}\end{center}\par\smallskip
\textit{Fundamentals}\par
\par\smallskip\begin{center}\resizebox{\ifdim\width>\linewidth\linewidth\else\width\fi}{!}{%
\begin{tabular}{lrrrrrrrr}
\toprule
\textbf{Company} & \textbf{MktCap} & \textbf{FwdP/E} & \textbf{Gross} & \textbf{Op} & \textbf{Net} & \textbf{RevGr} & \textbf{ROE} & \textbf{NetDebt/EBITDA} \\
\midrule
MU & \$913.66B & 5.26 & 72.6\% & 80.4\% & 55.9\% & 345.7\% & 66.6\% & -0.29x \\
TSM & \$1.99T & 17.76 & 64.2\% & 60.3\% & 49.9\% & 36.0\% & 40.0\% & -0.80x \\
AVGO & \$1.79T & 19.27 & 76.3\% & 49.0\% & 38.8\% & 47.9\% & 37.3\% & 1.08x \\
MRVL & \$152.55B & 27.23 & 51.5\% & 14.5\% & 29.0\% & 27.6\% & 16.0\% & 0.53x \\
\bottomrule
\end{tabular}}\end{center}\par\smallskip
Implied value/share: P/E \$2,581 $\cdot$ EV/EBITDA \$2,699 $\cdot$ EV/Rev \$1,548 $\cdot$ P/S \$1,399 \\
\textbf{Range: \$1,399 -- \$2,699} (median \$2,065) $\cdot$ Current price \$809 $\cdot$ median-implied upside 155.3\%\par\end{fhresp}
\par\addvspace{2.3mm}\noindent\hspace*{1.2em}{\footnotesize\color{caseTool}$\circ$}\,\texttt{\textcolor{caseTool}{data\_equity\_comps}}(ticker=SMCI, peers=[DELL, HPE, ANET, CSCO])\par\nopagebreak
\begin{fhresp}\textbf{SMCI} relative valuation vs 4 peer(s): \\
\textit{Trading multiples}\par
\par\smallskip\begin{center}\resizebox{\ifdim\width>\linewidth\linewidth\else\width\fi}{!}{%
\begin{tabular}{lrrrr}
\toprule
\textbf{Company} & \textbf{P/E} & \textbf{EV/EBITDA} & \textbf{EV/Rev} & \textbf{P/S} \\
\midrule
SMCI & 14.59 & 16.33 & 0.77 & 0.53 \\
DELL & 29.58 & 21.12 & 2.22 & 1.79 \\
HPE & 41.13 & 14.19 & 2.06 & 1.50 \\
ANET & 56.31 & 47.83 & 20.87 & 21.18 \\
CSCO & 38.12 & 27.51 & 7.70 & 7.42 \\
median & 39.63 & 24.32 & 4.96 & 4.61 \\
\bottomrule
\end{tabular}}\end{center}\par\smallskip
\textit{Fundamentals}\par
\par\smallskip\begin{center}\resizebox{\ifdim\width>\linewidth\linewidth\else\width\fi}{!}{%
\begin{tabular}{lrrrrrrrr}
\toprule
\textbf{Company} & \textbf{MktCap} & \textbf{FwdP/E} & \textbf{Gross} & \textbf{Op} & \textbf{Net} & \textbf{RevGr} & \textbf{ROE} & \textbf{NetDebt/EBITDA} \\
\midrule
SMCI & \$17.93B & 8.37 & 8.4\% & 6.1\% & 3.7\% & 122.7\% & 17.9\% & 4.97x \\
DELL & \$239.85B & 16.99 & 19.2\% & 8.9\% & 6.3\% & 87.5\% & n/a & 1.45x \\
HPE & \$58.28B & 10.98 & 33.8\% & 8.7\% & 4.0\% & 40.0\% & 6.3\% & 2.84x \\
ANET & \$205.63B & 36.56 & 63.5\% & 42.7\% & 38.3\% & 35.1\% & 31.5\% & -2.92x \\
CSCO & \$450.76B & 23.87 & 64.3\% & 25.0\% & 19.7\% & 12.0\% & 25.2\% & 0.96x \\
\bottomrule
\end{tabular}}\end{center}\par\smallskip
Implied value/share: P/E \$75 $\cdot$ EV/EBITDA \$47 $\cdot$ EV/Rev \$246 $\cdot$ P/S \$240 \\
\textbf{Range: \$47 -- \$246} (median \$158) $\cdot$ Current price \$28 $\cdot$ median-implied upside 468.5\%\par\end{fhresp}
\par\addvspace{2.3mm}\noindent\hspace*{1.2em}{\footnotesize\color{caseTool}$\circ$}\,\texttt{\textcolor{caseTool}{data\_equity\_comps}}(ticker=COHR, peers=[LITE, ANET, CSCO, MRVL])\par\nopagebreak
\begin{fhresp}\textbf{COHR} relative valuation vs 4 peer(s): \\
\textit{Trading multiples}\par
\par\smallskip\begin{center}\resizebox{\ifdim\width>\linewidth\linewidth\else\width\fi}{!}{%
\begin{tabular}{lrrrr}
\toprule
\textbf{Company} & \textbf{P/E} & \textbf{EV/EBITDA} & \textbf{EV/Rev} & \textbf{P/S} \\
\midrule
COHR & 110.56 & 41.42 & 8.24 & 6.85 \\
LITE & 108.16 & 100.61 & 20.57 & 19.27 \\
ANET & 56.31 & 47.83 & 20.87 & 21.18 \\
CSCO & 38.12 & 27.51 & 7.70 & 7.42 \\
MRVL & 58.48 & 61.55 & 19.15 & 17.50 \\
median & 57.40 & 54.69 & 19.86 & 18.39 \\
\bottomrule
\end{tabular}}\end{center}\par\smallskip
\textit{Fundamentals}\par
\par\smallskip\begin{center}\resizebox{\ifdim\width>\linewidth\linewidth\else\width\fi}{!}{%
\begin{tabular}{lrrrrrrrr}
\toprule
\textbf{Company} & \textbf{MktCap} & \textbf{FwdP/E} & \textbf{Gross} & \textbf{Op} & \textbf{Net} & \textbf{RevGr} & \textbf{ROE} & \textbf{NetDebt/EBITDA} \\
\midrule
COHR & \$45.20B & 27.87 & 37.0\% & 13.6\% & 7.1\% & 20.5\% & 4.7\% & 0.76x \\
LITE & \$47.96B & 33.59 & 40.8\% & 21.8\% & 17.7\% & 90.1\% & 22.8\% & 0.28x \\
ANET & \$205.63B & 36.56 & 63.5\% & 42.7\% & 38.3\% & 35.1\% & 31.5\% & -2.92x \\
CSCO & \$450.76B & 23.87 & 64.3\% & 25.0\% & 19.7\% & 12.0\% & 25.2\% & 0.96x \\
MRVL & \$152.74B & 27.27 & 51.5\% & 14.5\% & 29.0\% & 27.6\% & 16.0\% & 0.53x \\
\bottomrule
\end{tabular}}\end{center}\par\smallskip
Implied value/share: P/E \$120 $\cdot$ EV/EBITDA \$362 $\cdot$ EV/Rev \$665 $\cdot$ P/S \$620 \\
\textbf{Range: \$120 -- \$665} (median \$491) $\cdot$ Current price \$231 $\cdot$ median-implied upside 112.6\%\par\end{fhresp}
\par\addvspace{2.3mm}\noindent\hspace*{1.2em}{\footnotesize\color{caseTool}$\circ$}\,\texttt{\textcolor{caseTool}{data\_equity\_comps}}(ticker=VRT, peers=[ETN, ABB, PWR, \ldots; 5 values])\par\nopagebreak
\begin{fhresp}\textbf{VRT} relative valuation vs 4 peer(s): \\
\textit{Trading multiples}\par
\par\smallskip\begin{center}\resizebox{\ifdim\width>\linewidth\linewidth\else\width\fi}{!}{%
\begin{tabular}{lrrrr}
\toprule
\textbf{Company} & \textbf{P/E} & \textbf{EV/EBITDA} & \textbf{EV/Rev} & \textbf{P/S} \\
\midrule
VRT & 65.58 & 46.68 & 10.26 & 9.22 \\
ETN & 37.38 & 27.73 & 6.17 & n/a \\
PWR & 78.26 & 37.28 & 3.29 & 2.85 \\
EME & 23.47 & 17.62 & 1.84 & 1.75 \\
FIX & 38.13 & 31.56 & 5.63 & 4.86 \\
median & 37.75 & 29.65 & 4.46 & 2.85 \\
\bottomrule
\end{tabular}}\end{center}\par\smallskip
\textit{Fundamentals}\par
\par\smallskip\begin{center}\resizebox{\ifdim\width>\linewidth\linewidth\else\width\fi}{!}{%
\begin{tabular}{lrrrrrrrr}
\toprule
\textbf{Company} & \textbf{MktCap} & \textbf{FwdP/E} & \textbf{Gross} & \textbf{Op} & \textbf{Net} & \textbf{RevGr} & \textbf{ROE} & \textbf{NetDebt/EBITDA} \\
\midrule
VRT & \$100.01B & 29.49 & 37.2\% & 16.4\% & 14.4\% & 30.1\% & 45.1\% & 0.32x \\
ETN & \$146.45B & 23.91 & 37.1\% & 16.1\% & 14.0\% & 16.8\% & 20.8\% & 3.32x \\
PWR & \$85.96B & 34.73 & 15.1\% & 4.2\% & 3.7\% & 26.3\% & 13.5\% & 2.24x \\
EME & \$31.13B & 21.23 & 19.3\% & 8.7\% & 7.5\% & 19.7\% & 39.2\% & -0.22x \\
FIX & \$54.56B & 25.56 & 25.7\% & 7.9\% & 12.8\% & 1.0\% & 55.3\% & 0.12x \\
\bottomrule
\end{tabular}}\end{center}\par\smallskip
Implied value/share: P/E \$150 $\cdot$ EV/EBITDA \$182 $\cdot$ EV/Rev \$124 $\cdot$ P/S \$81 \\
\textbf{Range: \$81 -- \$182} (median \$137) $\cdot$ Current price \$260 $\cdot$ median-implied upside -47.4\%\par\end{fhresp}
\par\addvspace{2.3mm}\noindent\hspace*{1.2em}{\footnotesize\color{caseTool}$\circ$}\,\texttt{\textcolor{caseTool}{data\_equity\_comps}}(ticker=CEG, peers=[VST, NEE, GEV, \ldots; 5 values])\par\nopagebreak
\begin{fhresp}\textbf{CEG} relative valuation vs 5 peer(s): \\
\textit{Trading multiples}\par
\par\smallskip\begin{center}\resizebox{\ifdim\width>\linewidth\linewidth\else\width\fi}{!}{%
\begin{tabular}{lrrrr}
\toprule
\textbf{Company} & \textbf{P/E} & \textbf{EV/EBITDA} & \textbf{EV/Rev} & \textbf{P/S} \\
\midrule
CEG & 22.57 & 15.01 & 4.00 & 3.10 \\
VST & 24.76 & 11.10 & 3.88 & 2.57 \\
NEE & 20.37 & 20.81 & 10.58 & n/a \\
GEV & 26.43 & 65.59 & 6.23 & 5.94 \\
CMI & 32.75 & 19.47 & 2.87 & 2.57 \\
CAT & 41.17 & 30.36 & 6.25 & n/a \\
median & 26.43 & 20.81 & 6.23 & 2.57 \\
\bottomrule
\end{tabular}}\end{center}\par\smallskip
\textit{Fundamentals}\par
\par\smallskip\begin{center}\resizebox{\ifdim\width>\linewidth\linewidth\else\width\fi}{!}{%
\begin{tabular}{lrrrrrrrr}
\toprule
\textbf{Company} & \textbf{MktCap} & \textbf{FwdP/E} & \textbf{Gross} & \textbf{Op} & \textbf{Net} & \textbf{RevGr} & \textbf{ROE} & \textbf{NetDebt/EBITDA} \\
\midrule
CEG & \$92.70B & 19.29 & 23.3\% & 21.9\% & 12.7\% & 63.8\% & 16.1\% & 2.71x \\
VST & \$49.92B & 13.69 & 38.6\% & 26.6\% & 11.5\% & 43.4\% & 42.9\% & 2.93x \\
NEE & \$189.06B & 20.53 & 61.0\% & 31.5\% & 32.4\% & 12.4\% & 11.7\% & 7.35x \\
GEV & \$245.54B & 37.04 & 20.6\% & 7.5\% & 23.0\% & 21.9\% & 82.6\% & -2.29x \\
CMI & \$87.04B & 18.49 & 25.8\% & 9.8\% & 7.9\% & 2.7\% & 22.0\% & 1.01x \\
CAT & \$381.00B & 27.16 & 28.6\% & 18.2\% & 13.3\% & 22.2\% & 51.3\% & 2.73x \\
\bottomrule
\end{tabular}}\end{center}\par\smallskip
Implied value/share: P/E \$304 $\cdot$ EV/EBITDA \$403 $\cdot$ EV/Rev \$460 $\cdot$ P/S \$215 \\
\textbf{Range: \$215 -- \$460} (median \$354) $\cdot$ Current price \$260 $\cdot$ median-implied upside 36.2\%\par\end{fhresp}
\par\addvspace{2.3mm}\noindent\hspace*{1.2em}{\footnotesize\color{caseTool}$\circ$}\,\texttt{\textcolor{caseTool}{data\_equity\_comps}}(ticker=EQIX, peers=[DLR, IRM, AMT])\par\nopagebreak
\begin{fhresp}\textbf{EQIX} relative valuation vs 3 peer(s): \\
\textit{Trading multiples}\par
\par\smallskip\begin{center}\resizebox{\ifdim\width>\linewidth\linewidth\else\width\fi}{!}{%
\begin{tabular}{lrrrr}
\toprule
\textbf{Company} & \textbf{P/E} & \textbf{EV/EBITDA} & \textbf{EV/Rev} & \textbf{P/S} \\
\midrule
EQIX & 72.18 & 29.25 & 13.15 & 10.80 \\
DLR & 94.93 & 28.88 & 13.70 & 10.89 \\
IRM & 136.83 & 23.51 & 7.95 & 5.11 \\
AMT & 28.57 & 18.29 & 11.81 & 7.63 \\
median & 94.93 & 23.51 & 11.81 & 7.63 \\
\bottomrule
\end{tabular}}\end{center}\par\smallskip
\textit{Fundamentals}\par
\par\smallskip\begin{center}\resizebox{\ifdim\width>\linewidth\linewidth\else\width\fi}{!}{%
\begin{tabular}{lrrrrrrrr}
\toprule
\textbf{Company} & \textbf{MktCap} & \textbf{FwdP/E} & \textbf{Gross} & \textbf{Op} & \textbf{Net} & \textbf{RevGr} & \textbf{ROE} & \textbf{NetDebt/EBITDA} \\
\midrule
EQIX & \$102.86B & 54.00 & 51.6\% & 24.3\% & 14.9\% & 12.1\% & 10.1\% & 5.13x \\
DLR & \$73.58B & 66.67 & 57.0\% & 26.1\% & 11.8\% & 29.9\% & 2.9\% & 5.61x \\
IRM & \$37.05B & 46.54 & 55.0\% & 21.0\% & 3.8\% & 21.6\% & n/a & 7.95x \\
AMT & \$82.53B & 25.83 & 74.0\% & 45.9\% & 26.8\% & 6.8\% & 30.0\% & 6.23x \\
\bottomrule
\end{tabular}}\end{center}\par\smallskip
Implied value/share: P/E \$1,372 $\cdot$ EV/EBITDA \$798 $\cdot$ EV/Rev \$918 $\cdot$ P/S \$737 \\
\textbf{Range: \$737 -- \$1,372} (median \$858) $\cdot$ Current price \$1,043 $\cdot$ median-implied upside -17.7\%\par\end{fhresp}
\par\addvspace{2.3mm}\noindent\hspace*{1.2em}{\footnotesize\color{caseTool}$\circ$}\,\texttt{\textcolor{caseTool}{data\_equity\_comps}}(ticker=MSFT, peers=[AMZN, GOOGL, ORCL, META])\par\nopagebreak
\begin{fhresp}\textbf{MSFT} relative valuation vs 4 peer(s): \\
\textit{Trading multiples}\par
\par\smallskip\begin{center}\resizebox{\ifdim\width>\linewidth\linewidth\else\width\fi}{!}{%
\begin{tabular}{lrrrr}
\toprule
\textbf{Company} & \textbf{P/E} & \textbf{EV/EBITDA} & \textbf{EV/Rev} & \textbf{P/S} \\
\midrule
MSFT & 23.10 & 15.93 & 9.23 & 9.23 \\
AMZN & 27.40 & 16.56 & 3.48 & 3.31 \\
GOOGL & 16.46 & 22.46 & 8.72 & 9.00 \\
ORCL & 19.83 & 15.95 & 7.22 & 4.94 \\
META & 21.51 & 13.84 & 7.04 & 6.99 \\
median & 20.67 & 16.26 & 7.13 & 5.96 \\
\bottomrule
\end{tabular}}\end{center}\par\smallskip
\textit{Fundamentals}\par
\par\smallskip\begin{center}\resizebox{\ifdim\width>\linewidth\linewidth\else\width\fi}{!}{%
\begin{tabular}{lrrrrrrrr}
\toprule
\textbf{Company} & \textbf{MktCap} & \textbf{FwdP/E} & \textbf{Gross} & \textbf{Op} & \textbf{Net} & \textbf{RevGr} & \textbf{ROE} & \textbf{NetDebt/EBITDA} \\
\midrule
MSFT & \$2.94T & 20.42 & 68.3\% & 46.3\% & 39.3\% & 18.3\% & 34.0\% & 0.26x \\
AMZN & \$2.46T & 23.07 & 50.6\% & 13.1\% & 12.2\% & 16.6\% & 24.3\% & 0.59x \\
GOOGL & \$4.01T & 22.30 & 60.9\% & 34.0\% & 54.8\% & 24.2\% & 48.7\% & -0.70x \\
ORCL & \$332.92B & 10.61 & 65.8\% & 36.2\% & 25.4\% & 20.6\% & 53.4\% & 4.44x \\
META & \$1.50T & 15.98 & 81.9\% & 40.6\% & 32.8\% & 33.1\% & 32.9\% & 0.05x \\
\bottomrule
\end{tabular}}\end{center}\par\smallskip
Implied value/share: P/E \$354 $\cdot$ EV/EBITDA \$397 $\cdot$ EV/Rev \$299 $\cdot$ P/S \$256 \\
\textbf{Range: \$256 -- \$397} (median \$326) $\cdot$ Current price \$395 $\cdot$ median-implied upside -17.4\%\par\end{fhresp}
\par\addvspace{3.6mm}\noindent\rbadge{R4}\par\nopagebreak
\par\addvspace{2.3mm}\noindent\hspace*{1.2em}{\footnotesize\color{caseTool}$\circ$}\,\texttt{\textcolor{caseTool}{update\_plan}}: {\color{black!45}\ldots{}~(plan unchanged)}~$\rightarrow$~Plan updated --- 1/4 step(s) done.\par\nopagebreak
\par\addvspace{3.6mm}\noindent\rbadge{R5}\par\nopagebreak
\par\addvspace{2.3mm}\noindent\hspace*{1.2em}{\footnotesize\color{caseTool}$\circ$}\,\texttt{\textcolor{caseTool}{search}}: [2026 AI data center capex hyperscalers spending forecast 2026 Microsoft Amazon Alphabet Meta official earnings capex, site:investor.nvidia.com annual report 2026 data center revenue Blackwell supply partners, site:investor.tsmc.com 2026 quarterly results AI accelerator revenue CAGR CoWoS advanced packaging, site:asml.com investors 2026 AI demand EUV data center, site:investors.micron.com 2026 HBM sold out AI data center demand earnings, site:investors.broadcom.com 2026 AI revenue custom accelerators networking earnings, site:investors.arista.com 2026 AI networking revenue earnings, site:investors.vertiv.com 2026 earnings AI data center backlog liquid cooling, site:eaton.com investors 2026 data center demand electrical backlog earnings, site:investor.gevernova.com 2026 data center electricity demand gas turbines backlog, site:\allowbreak{}constellationen\allowbreak{}ergy.\allowbreak{}com investors data center AI power agreement nuclear Microsoft 2025 2026, site:iea.org data centres AI electricity demand 2030 report, site:investor.equinix.com 2026 AI data center demand xScale earnings, site:\allowbreak{}investor.\allowbreak{}digitalrealty.\allowbreak{}com 2026 AI data center demand leasing backlog, site:\allowbreak{}investors.\allowbreak{}delltechnologie\allowbreak{}s.\allowbreak{}com 2026 AI server backlog revenue earnings, site:ir.supermicro.com 2026 AI infrastructure revenue liquid cooling, site:coherent.com investors 2026 datacenter transceiver AI demand earnings, site:lumentum.com investors 2026 cloud optical demand AI earnings]~(max\_results=5)\par\nopagebreak
\begin{fhresp}13 results: \\
\fhlink{https://www.cnbc.com/2026/07/28/hyperscalers-face-higher-capex-scrutiny-after-alphabet-report-panned.html}{Hyperscalers face higher capex scrutiny after Alphabet report panned} \\
\fhlink{https://drrobertcastellano.substack.com/p/alphabet-will-raise-capexand-extend}{Alphabet Will Raise Capex---and Extend the AI Semiconductor Cycle} \\
\fhlink{https://know.creditsights.com/insights/technology-hyperscaler-capex-2026-estimates/}{Technology: Hyperscaler Capex 2026 Estimates - CreditSights} \\
\fhlink{https://www.statista.com/chart/35046/capital-expenditure-of-meta-alphabet-amazon-and-microsoft/?srsltid=AfmBOoprvO5no0JBpyom4tYUcFM3XXqJj3gDzZrUpiHbHo9fOhqUGlaM}{Chart: Big Tech's AI Spending to Reach \$725 Billion in 2026 - Statista} \\
{\color{black!45}\ldots\ (9 more)}\end{fhresp}
\par\addvspace{3.6mm}\noindent\rbadge{R6}\par\nopagebreak
\par\addvspace{2.3mm}\noindent\hspace*{1.2em}{\footnotesize\color{caseTool}$\circ$}\,\texttt{\textcolor{caseTool}{search}}: [TSMC Q2 2026 results AI accelerator revenue advanced packaging CoWoS, Micron fiscal Q3 2026 results HBM data center demand sold out, Broadcom 2026 quarterly results AI revenue custom accelerators networking, Arista 2026 quarterly results AI networking revenue Etherlink, Vertiv 2026 quarterly results backlog AI data centers liquid cooling, Eaton 2026 quarterly results data center orders backlog, GE Vernova 2026 quarterly results data center power demand gas turbine backlog, IEA Energy and AI report data centres electricity demand 2030, Dell fiscal 2027 results AI server backlog 2026, Super Micro fiscal 2026 results AI server revenue liquid cooling, Coherent 2026 quarterly results datacenter transceiver AI demand, Digital Realty 2026 results AI leasing backlog data center]~(max\_results=6)\par\nopagebreak
\begin{fhresp}14 results: \\
\fhlink{https://phemex.com/blogs/tsmc-record-revenue-earnings-ai}{TSMC Q2 2026 Revenue Sets a Record | What It Means for AI} \\
\fhlink{https://smartinvestorsdaily.com/guides/tsm-earnings-q2-2026/}{TSM Earnings Q2 2026: Results, Analysis \& Outlook ...} \\
\fhlink{https://investor.tsmc.com/english/quarterly-results/2026/q2}{TSMC 2026 Q2 Quarterly Results - Taiwan Semiconductor ...} \\
\fhlink{https://aitoolsrecap.com/Blog/tsmc-q2-record-revenue-ai-chips-july-2026}{TSMC Posts All-Time Revenue Record --- June Sales Up 68\%, N3 ...} \\
{\color{black!45}\ldots\ (10 more)}\end{fhresp}
\par\addvspace{3.6mm}\noindent\rbadge{R7}\par\nopagebreak
\par\addvspace{2.3mm}\noindent\hspace*{1.2em}{\footnotesize\color{caseTool}$\circ$}\,\texttt{\textcolor{caseTool}{search}}: [site:\allowbreak{}investors.\allowbreak{}micron.\allowbreak{}com/\allowbreak{}news-\allowbreak{}releases 2026 quarterly results HBM revenue, site:\allowbreak{}investors.\allowbreak{}broadcom.\allowbreak{}com/\allowbreak{}news-\allowbreak{}releases 2026 financial results AI revenue, site:\allowbreak{}investors.\allowbreak{}arista.\allowbreak{}com/\allowbreak{}news-\allowbreak{}releases 2026 financial results AI networking, site:\allowbreak{}investors.\allowbreak{}vertiv.\allowbreak{}com/\allowbreak{}news-\allowbreak{}releases 2026 financial results organic orders backlog, site:\allowbreak{}eaton.\allowbreak{}com/\allowbreak{}us/\allowbreak{}en-\allowbreak{}us/\allowbreak{}company/\allowbreak{}investor-\allowbreak{}relations 2026 earnings data center orders, site:\allowbreak{}investor.\allowbreak{}gevernova.\allowbreak{}com/\allowbreak{}news-\allowbreak{}releases 2026 financial results backlog, site:\allowbreak{}investors.\allowbreak{}delltechnologie\allowbreak{}s.\allowbreak{}com/\allowbreak{}news-\allowbreak{}releases 2026 financial results AI servers backlog]~(max\_results=8)\par\nopagebreak
\begin{fhresp}46 results: \\
\fhlink{https://investors.micron.com/news-releases/news-release-details/micron-high-volume-production-hbm4-designed-nvidia-vera-rubin}{Micron in High-Volume Production of HBM4 Designed for NVIDIA ...} \\
\fhlink{https://investors.micron.com/news-releases/news-release-details/micron-powers-ai-everywhere-computex-2026}{Micron Powers AI Everywhere at COMPUTEX 2026 | Micron Technology} \\
\fhlink{https://investors.micron.com/news-releases/news-release-details/micron-technology-inc-reports-results-second-quarter-fiscal-2025}{Micron Technology, Inc. Reports Results for the Second Quarter of...} \\
\fhlink{https://investors.micron.com/news-releases/news-release-details/micron-ships-hbm4-key-customers-power-next-gen-ai-platforms}{Micron Ships HBM4 to Key Customers to Power Next-Gen AI Platforms} \\
{\color{black!45}\ldots\ (42 more)}\end{fhresp}
\par\addvspace{3.6mm}\noindent\rbadge{R8}\par\nopagebreak
\par\addvspace{2.3mm}\noindent\hspace*{1.2em}{\footnotesize\color{caseTool}$\circ$}\,\texttt{\textcolor{caseTool}{search}}: [Arista Networks Q2 2026 financial results AI Ethernet, Vertiv Q2 2026 financial results orders backlog AI, GE Vernova Q2 2026 financial results gas power backlog data center, Eaton second quarter 2026 earnings data center orders, Constellation Energy Q2 2026 data center power agreements, Digital Realty Q2 2026 earnings AI demand bookings, Equinix Q2 2026 earnings AI xScale, Coherent fiscal 2026 results datacenter growth transceivers]~(max\_results=8)\par\nopagebreak
\begin{fhresp}11 results: \\
\fhlink{https://quartr.com/companies/arista-networks-inc_4569}{Arista Networks (ANET) Investor Relations, Earnings Summary ...} \\
\fhlink{https://finance.yahoo.com/markets/stocks/articles/expect-arista-networks-q2-2026-005607206.html}{What to Expect From Arista Networks' Q2 2026 Earnings Report} \\
\fhlink{https://www.tikr.com/blog/arista-networks-grew-revenue-35-last-quarter-and-generates-4-billion-in-annual-free-cash-flow}{Arista Networks Grew Revenue 35\% Last Quarter and Generates \$4 ...} \\
\fhlink{https://finance.yahoo.com/technology/ai/articles/arista-networks-anet-strengthens-exponential-131206001.html}{Arista Networks (ANET) Strengthens from Exponential Growth in AI} \\
{\color{black!45}\ldots\ (7 more)}\end{fhresp}
\par\addvspace{3.6mm}\noindent\rbadge{R9}\par\nopagebreak
\par\addvspace{2.3mm}\noindent\hspace*{1.2em}{\footnotesize\color{caseTool}$\circ$}\,\texttt{\textcolor{caseTool}{search}}: [site:investors.vertiv.com "Second Quarter 2026", site:investors.vertiv.com "First Quarter 2026" orders organic sales backlog, site:investor.gevernova.com "Second Quarter 2026", site:investor.gevernova.com "First Quarter 2026" backlog, site:eaton.com "Second Quarter 2026" earnings data center, site:eaton.com "First Quarter 2026" data center orders, site:\allowbreak{}investor.\allowbreak{}digitalrealty.\allowbreak{}com "Second Quarter 2026", site:investor.equinix.com "Second Quarter 2026"]~(max\_results=8)\par\nopagebreak
\begin{fhresp}6 results: \\
\fhlink{https://investors.vertiv.com/news/news-details/2026/Vertiv-Reports-Strong-First-Quarter-with-Diluted-EPS-Growth-of-136-Adjusted-Diluted-EPS-Growth-of-83-Raises-Full-Year-Guidance/default.aspx}{Vertiv Holdings Co. - Vertiv Reports Strong First Quarter ...} \\
\fhlink{https://investors.vertiv.com/overview/default.aspx}{Vertiv Holdings Co. - Investor relations} \\
\fhlink{https://investors.vertiv.com/news/news-details/2026/Vertiv-Announces-Date-of-Second-Quarter-2026-Earnings-Release-and-Conference-Call/default.aspx}{Vertiv Announces Date of Second Quarter 2026 Earnings Release ...} \\
\fhlink{https://investors.vertiv.com/events-presentations/events-presentations/event-details/2026/Vertiv-Second-Quarter-2026-Earnings-Conference-Call/default.aspx}{Vertiv Holdings Co. - Vertiv Second Quarter 2026 Earnings ...} \\
{\color{black!45}\ldots\ (2 more)}\end{fhresp}
\par\addvspace{3.6mm}\noindent\rbadge{R10}\par\nopagebreak
\par\addvspace{2.3mm}\noindent\hspace*{1.2em}{\footnotesize\color{caseTool}$\circ$}\,\texttt{\textcolor{caseTool}{search}}: [site:\allowbreak{}iea.\allowbreak{}org/\allowbreak{}reports/\allowbreak{}energy-\allowbreak{}and-\allowbreak{}ai executive summary data centres electricity 2030, site:iea.org "Energy and AI" data centres 2030 electricity demand, site:iea.org data centers electricity demand United States 2030 AI]~(max\_results=6)\par\nopagebreak
\begin{fhresp}15 results: \\
\fhlink{https://www.iea.org/reports/energy-and-ai/executive-summary}{Executive summary -- Energy and AI -- Analysis - IEA} \\
\fhlink{https://www.iea.org/reports/energy-and-ai/energy-demand-from-ai}{Energy demand from AI -- Energy and AI -- Analysis - IEA} \\
\fhlink{https://www.iea.org/reports/energy-and-ai/energy-supply-for-ai}{Energy supply for AI -- Energy and AI -- Analysis - IEA} \\
\fhlink{https://www.iea.org/reports/energy-and-ai/ai-and-climate-change}{AI and climate change -- Energy and AI -- Analysis - IEA} \\
{\color{black!45}\ldots\ (11 more)}\end{fhresp}
\par\addvspace{3.6mm}\noindent\rbadge{R11}\par\nopagebreak
\par\addvspace{2.3mm}\noindent\hspace*{1.2em}{\footnotesize\color{caseTool}$\circ$}\,\texttt{\textcolor{caseTool}{search}}: [ASML 2025 annual report AI demand EUV systems revenue 2026 outlook, Applied Materials fiscal 2026 results AI semiconductor equipment advanced packaging, Lam Research 2026 results AI HBM advanced packaging demand, KLA 2026 earnings AI advanced packaging process control demand, TSMC annual report 2025 advanced technologies 7nm below revenue AI accelerator CAGR]~(max\_results=6)\par\nopagebreak
\begin{fhresp}14 results: \\
\fhlink{https://www.asml.com/en/investors/annual-report/2025/financials}{Financials | ASML 2025 Annual Report} \\
\fhlink{https://www.asml.com/en/news/press-releases/2026/q4-2025-financial-results}{ASML reports \texteuro{}32.7 billion total net sales and \texteuro{}9.6 billion ...} \\
\fhlink{https://www.nasdaq.com/press-release/asml-reports-eu327-billion-total-net-sales-and-eu96-billion-net-income-2025-2026-01}{ASML reports \texteuro{}32.7 billion total net sales and \texteuro{}9.6 billion ...} \\
\fhlink{https://chippub.substack.com/p/asml-2025-growth-through-structural}{ASML 2025: Growth Through Structural Shifts - by Meng Li} \\
{\color{black!45}\ldots\ (10 more)}\end{fhresp}
\par\addvspace{3.6mm}\noindent\rbadge{R12}\par\nopagebreak
\par\addvspace{2.3mm}\noindent\hspace*{1.2em}{\footnotesize\color{caseTool}$\circ$}\,\texttt{\textcolor{caseTool}{search}}: [Quanta Services 2026 results data center load growth transmission demand, EMCOR 2026 earnings data center demand backlog, Comfort Systems USA 2026 earnings data center revenue backlog, GE Vernova 2025 annual report data center power demand gas turbine backlog, Constellation Microsoft Three Mile Island power purchase agreement data center official, Vistra data center power demand investor presentation 2026, Digital Realty annual report 2025 AI demand development pipeline]~(max\_results=8)\par\nopagebreak
\begin{fhresp}12 results: \\
\fhlink{https://finance.yahoo.com/news/quantas-110-tech-growth-outlook-124500974.html}{Is Quanta's 110\% Tech Growth Outlook Fueling Its ... - Yahoo Finance} \\
\fhlink{https://investors.quantaservices.com/news-events/press-releases/detail/396/quanta-services-reports-first-quarter-2026-results}{QUANTA SERVICES REPORTS FIRST QUARTER 2026 RESULTS} \\
\fhlink{https://investors.quantaservices.com/sec-filings/all-sec-filings/content/0001193125-26-058069/0001193125-26-058069.pdf}{02/19/2026 - Quanta Services, Inc.} \\
\fhlink{https://www.tikr.com/blog/quanta-services-stock-is-surging-the-backlog-says-its-not-done}{Quanta Services Stock Is Surging. The Backlog Says It's Not Done.} \\
{\color{black!45}\ldots\ (8 more)}\end{fhresp}
\par\addvspace{3.6mm}\noindent\rbadge{R13}\par\nopagebreak
\par\addvspace{2.3mm}\noindent\hspace*{1.2em}{\footnotesize\color{caseTool}$\circ$}\,\texttt{\textcolor{caseTool}{search}}: [site:\allowbreak{}investor.\allowbreak{}digitalrealty.\allowbreak{}com/\allowbreak{}news-\allowbreak{}releases 2026 first quarter results bookings development pipeline, site:\allowbreak{}investor.\allowbreak{}equinix.\allowbreak{}com/\allowbreak{}news-\allowbreak{}events 2026 first quarter results xScale, site:ir.coreweave.com 2026 first quarter results backlog debt capex, site:\allowbreak{}investor.\allowbreak{}constellationen\allowbreak{}ergy.\allowbreak{}com 2026 first quarter results data center, site:\allowbreak{}investor.\allowbreak{}vistracorp.\allowbreak{}com 2026 first quarter results data center]~(max\_results=8)\par\nopagebreak
\begin{fhresp}11 results: \\
\fhlink{https://investor.equinix.com/news-events/press-releases/detail/1107/equinix-reports-first-quarter-results-and-raises-full-year}{Equinix Reports First-Quarter Results and... :: Equinix, Inc. (EQIX)} \\
\fhlink{https://investor.equinix.com/news-events/press-releases/detail/1096/equinix-provides-robust-2026-outlook-driven-by-strong}{Equinix Provides Robust 2026 Outlook Driven by... :: Equinix, Inc. (EQIX)} \\
\fhlink{https://investor.equinix.com/news-events/press-releases/detail/1068/equinix-reports-first-quarter-2025-results}{Equinix Reports First-Quarter 2025 Results :: Equinix, Inc. (EQIX)} \\
\fhlink{https://investor.equinix.com/news-events/press-releases/detail/1086/equinix-reports-strong-third-quarter-2025-results}{Equinix Reports Strong Third-Quarter 2025 Results} \\
{\color{black!45}\ldots\ (7 more)}\end{fhresp}
\par\addvspace{3.6mm}\noindent\rbadge{R14}\par\nopagebreak
\par\addvspace{2.3mm}\noindent\hspace*{1.2em}{\footnotesize\color{caseTool}$\circ$}\,\texttt{\textcolor{caseTool}{search}}: [Alphabet Q2 2026 results capex guidance 2026 investor relations, Amazon Q1 2026 earnings capex 200 billion official, Meta Q2 2026 earnings capex guidance 2026 investor relations, Microsoft fiscal Q3 2026 earnings capex AI infrastructure official, Oracle fiscal 2026 results remaining performance obligations AI infrastructure]~(max\_results=8)\par\nopagebreak
\begin{fhresp}14 results: \\
\fhlink{https://abc.xyz/investor/}{Alphabet Investor Relations} \\
\fhlink{https://nukcloud.com/en/blog/2026-google-alphabet-q2-2026-earnings-cloud-capex-gemini-20260723.html}{Google Q2 2026 Earnings: Cloud Up 82\%, \$205B AI Capex Guide} \\
\fhlink{https://thenextweb.com/news/alphabet-q2-2026-capex-205bn-cloud-82}{Alphabet lifts capex guide to \$205bn as Google Cloud jumps 82\%} \\
\fhlink{https://www.fool.com/investing/2026/07/27/prediction-ceo-andy-jassy-will-raise-amazons-full/}{Prediction: CEO Andy Jassy Will Raise Amazon's Full-Year 2026...} \\
{\color{black!45}\ldots\ (10 more)}\end{fhresp}
\par\addvspace{3.6mm}\noindent\rbadge{R15}\par\nopagebreak
\par\addvspace{2.3mm}\noindent\hspace*{1.2em}{\footnotesize\color{caseTool}$\circ$}\,\texttt{\textcolor{caseTool}{search}}: [Constellation Microsoft 20-year power purchase agreement Crane Clean Energy Center official September 2024, GE Vernova data center demand gas turbines backlog 2025 annual report]~(max\_results=8)\par\nopagebreak
\begin{fhresp}13 results: \\
\fhlink{https://wnn.origindigital.co/articles/constellation-to-restart-three-mile-island-unit-powering-microsoft}{Constellation to restart Three Mile Island unit, powering Microsoft} \\
\fhlink{https://www.globaleconomics.net/post/powering-the-ai-revolution-energy-infrastructure-in-the-united-states}{Powering the AI Revolution: Energy Infrastructure in the United States} \\
\fhlink{https://www.hawkmontresearch.com/post/the-ai-boom-s-real-bottleneck-why-nuclear-energy-and-constellation-energy-could-deliver-the-decade}{The AI Boom's Real Bottleneck: Why Nuclear Energy and...} \\
\fhlink{https://www.linkedin.com/feed/update/urn:li:share:7242889147001016322}{Constellation to Launch Crane Clean Energy Center, Restoring Jobs...} \\
{\color{black!45}\ldots\ (9 more)}\end{fhresp}
\par\addvspace{3.6mm}\noindent\rbadge{R16}\par\nopagebreak
\par\addvspace{2.3mm}\noindent\hspace*{1.2em}{\footnotesize\color{caseTool}$\circ$}\,\texttt{\textcolor{caseTool}{visit}}: \fhlink{https://investor.nvidia.com/news/press-release-details/2026/NVIDIA-Announces-Financial-Results-for-First-Quarter-Fiscal-2027/default.aspx}{investor.nvidia.com}, \fhlink{https://investor.tsmc.com/english/quarterly-results/2026/q2}{investor.tsmc.com}, \fhlink{https://investors.micron.com/news-releases/news-release-details/micron-technology-inc-reports-record-results-third-quarter}{investors.micron.com}, \fhlink{https://investors.broadcom.com/news-releases/news-release-details/broadcom-inc-announces-second-quarter-fiscal-year-2026-financial}{investors.broadcom.com}, \fhlink{https://www.asml.com/en/news/press-releases/2026/q4-2025-financial-results}{asml.com}, \fhlink{https://investors.delltechnologies.com/news-releases/news-release-details/dell-technologies-delivers-fourth-quarter-and-full-year-fiscal-3}{investors.delltechnologies.com}, \fhlink{https://investors.vertiv.com/news/news-details/2026/Vertiv-Reports-Strong-First-Quarter-with-Diluted-EPS-Growth-of-136-Adjusted-Diluted-EPS-Growth-of-83-Raises-Full-Year-Guidance/default.aspx}{investors.vertiv.com}, \fhlink{https://investors.quantaservices.com/news-events/press-releases/detail/396/quanta-services-reports-first-quarter-2026-results}{investors.quantaservices.com}, \fhlink{https://www.iea.org/reports/energy-and-ai/executive-summary}{iea.org}, \fhlink{https://investor.equinix.com/news-events/press-releases/detail/1107/equinix-reports-first-quarter-results-and-raises-full-year}{investor.equinix.com}, \fhlink{https://abc.xyz/investor/events/event-details/2026/2026-Q2-Earnings-Call-2026-GgTAq7Is0z/default.aspx}{abc.xyz}, \fhlink{https://www.cnbc.com/2026/07/28/hyperscalers-face-higher-capex-scrutiny-after-alphabet-report-panned.html}{cnbc.com}~(goal=Extract current reported financial results,\allowbreak{} AI/\allowbreak{}data-\allowbreak{}center demand linkage,\allowbreak{} capacity/\allowbreak{}backlog,\allowbreak{} outlook,\allowbreak{} margin and capex figures,\allowbreak{} plus risks and caveats.)\par\nopagebreak
\par\addvspace{3.6mm}\noindent\rbadge{R17}\par\nopagebreak
\par\addvspace{2.3mm}\noindent\hspace*{1.2em}{\footnotesize\color{caseTool}$\circ$}\,\texttt{\textcolor{caseTool}{search}}: ["Micron Technology, Inc. Reports Record Results for the Third Quarter of Fiscal 2026", "Broadcom Inc. Announces Second Quarter Fiscal Year 2026 Financial Results" AI revenue, "Dell Technologies Delivers Fourth Quarter and Full-Year Fiscal 2026 Financial Results" AI servers, Arista first quarter 2026 results revenue AI networking official]~(max\_results=8)\par\nopagebreak
\begin{fhresp}11 results: \\
\fhlink{https://www.cxodigitalpulse.com/micron-technology-inc-reports-record-results-for-the-third-quarter-of-fiscal-2026/}{Micron Technology, Inc. Reports Record Results for the Third ...} \\
\fhlink{https://www.techpowerup.com/350262/micron-technology-inc-reports-record-results-for-the-third-quarter-of-fiscal-2026}{Micron Technology, Inc. Reports Record Results for the Third ...} \\
\fhlink{https://www.nasdaq.com/press-release/micron-technology-inc-reports-record-results-third-quarter-fiscal-2026-2026-06-24}{Micron Technology, Inc. Reports Record Results for the Third ...} \\
\fhlink{https://www.barchart.com/story/news/2634033/micron-technology-inc-reports-record-results-for-the-third-quarter-of-fiscal-2026}{Micron Technology, Inc. Reports Record Results for the Third ...} \\
{\color{black!45}\ldots\ (7 more)}\end{fhresp}
\par\addvspace{3.6mm}\noindent\rbadge{R18}\par\nopagebreak
\par\addvspace{2.3mm}\noindent\hspace*{1.2em}{\footnotesize\color{caseTool}$\circ$}\,\texttt{\textcolor{caseTool}{search}}: [Broadcom Q2 fiscal 2026 results AI semiconductor revenue 2026, Dell fiscal 2026 full year AI server revenue backlog February 2026, Arista Q1 2026 financial results AI revenue]~(max\_results=10)\par\nopagebreak
\begin{fhresp}13 results: \\
\fhlink{https://tech-insider.org/broadcom-q2-2026-earnings-ai-revenue/}{Broadcom Earnings: AI Revenue Hits \$10.8B [2026]} \\
\fhlink{https://www.earningslens.ai/report/avgo/2026-Q2}{Broadcom (AVGO) Q2 FY2026 Earnings: \$22.19B Revenue} \\
\fhlink{https://www.cnbc.com/2026/06/03/broadcom-avgo-earnings-report-q2-2026.html}{Broadcom (AVGO) earnings report Q2 2026 - CNBC} \\
\fhlink{https://tickeron.com/earnings/AVGO/}{Broadcom Inc. (AVGO) Q2 Fiscal 2026 Earnings Recap: AI Demand ...} \\
{\color{black!45}\ldots\ (9 more)}\end{fhresp}
\par\addvspace{3.6mm}\noindent\rbadge{R19}\par\nopagebreak
\par\addvspace{2.3mm}\noindent\hspace*{1.2em}{\footnotesize\color{caseTool}$\circ$}\,\texttt{\textcolor{caseTool}{search}}: [Dell fiscal Q4 2026 AI server revenue backlog February 26 2026, Dell full year fiscal 2026 AI optimized server shipments backlog 2026, Arista first quarter 2026 results May 2026 revenue AI networking official release]~(max\_results=10)\par\nopagebreak
\begin{fhresp}12 results: \\
\fhlink{https://www.dell.com/en-us/dt/corporate/newsroom/announcements/detailpage.press-releases~usa~2026~2~dell-technologies-delivers-fourth-quarter-and-full-year-fiscal-2026-results.htm}{Dell Technologies Delivers Fourth Quarter and Full-Year ...} \\
\fhlink{https://www.blocksandfiles.com/flash/2026/02/27/ai-server-frenzy-fuels-record-revenues-for-dell/4092727}{AI server frenzy fuels record revenues for Dell} \\
\fhlink{https://finance.yahoo.com/news/dell-technologies-inc-dell-q4-050504335.html?fr=sycsrp_catchall}{Dell Technologies Inc (DELL) Q4 2026 Earnings Call Highlights ...} \\
\fhlink{https://www.marketbeat.com/earnings/reports/2026-2-26-dell-technologies-inc-stock/}{DELL Q4 2026 Earnings Report on 2/26/2026 - MarketBeat} \\
{\color{black!45}\ldots\ (8 more)}\end{fhresp}
\par\addvspace{3.6mm}\noindent\rbadge{R20}\par\nopagebreak
\par\addvspace{2.3mm}\noindent\hspace*{1.2em}{\footnotesize\color{caseTool}$\circ$}\,\texttt{\textcolor{caseTool}{search}}: [Coherent fiscal third quarter 2026 results datacenter transceiver AI official, Lumentum fiscal third quarter 2026 results cloud data center AI official, Marvell fiscal 2026 results data center revenue AI]~(max\_results=10)\par\nopagebreak
\begin{fhresp}12 results: \\
\fhlink{https://www.coherent.com/news/press-releases/third-quarter-fiscal-year-2026-results}{Coherent Corp. Reports Third Quarter Fiscal 2026 Results} \\
\fhlink{https://www.marketbeat.com/earnings/reports/2026-5-6-coherent-inc-stock/}{COHR Q3 2026 Earnings Report on 5/6/2026 - MarketBeat} \\
\fhlink{https://futurumgroup.com/insights/coherent-q3-fy-2026-ai-data-center-demand-accelerates-optical-growth/}{Coherent Q3 FY 2026: AI Data Center Demand Accelerates ...} \\
\fhlink{https://finance.biggo.com/news/US_COHR_2026-05-06}{[COHR Q3 2026 Earnings Call] Record Revenue of \$1.8B, AI ...} \\
{\color{black!45}\ldots\ (8 more)}\end{fhresp}
\par\addvspace{3.6mm}\noindent\rbadge{R21}\par\nopagebreak
\par\addvspace{2.3mm}\noindent\hspace*{1.2em}{\footnotesize\color{caseTool}$\circ$}\,\texttt{\textcolor{caseTool}{visit}}: \fhlink{https://www.nasdaq.com/press-release/micron-technology-inc-reports-record-results-third-quarter-fiscal-2026-2026-06-24}{nasdaq.com}, \fhlink{https://www.cnbc.com/2026/06/03/broadcom-avgo-earnings-report-q2-2026.html}{cnbc.com}, \fhlink{https://www.dell.com/en-us/dt/corporate/newsroom/announcements/detailpage.press-releases~usa~2026~2~dell-technologies-delivers-fourth-quarter-and-full-year-fiscal-2026-results.htm}{dell.com}, \fhlink{https://www.coherent.com/news/press-releases/third-quarter-fiscal-year-2026-results}{coherent.com}, \fhlink{https://investors.arista.com/Financial-Information/default.aspx}{investors.arista.com}, \fhlink{https://www.iea.org/reports/energy-and-ai/energy-supply-for-ai}{iea.org}~(goal=Extract company results,\allowbreak{} growth and margins,\allowbreak{} AI/\allowbreak{}data-\allowbreak{}center linkage,\allowbreak{} orders/\allowbreak{}backlog/\allowbreak{}capacity,\allowbreak{} outlook and risks.)\par\nopagebreak
\par\addvspace{3.6mm}\noindent\rbadge{R22}\par\nopagebreak
\par\addvspace{2.3mm}\noindent\hspace*{1.2em}{\footnotesize\color{caseTool}$\circ$}\,\texttt{\textcolor{caseTool}{search}}: [TSMC Q2 2026 earnings call AI accelerator revenue CAGR CoWoS capacity 2026 transcript, TSMC 2026 Q2 AI demand CoWoS capacity Reuters July 2026, TSMC 2025 annual report AI accelerators revenue advanced packaging CoWoS]~(max\_results=8)\par\nopagebreak
\begin{fhresp}12 results: \\
\fhlink{https://investor.tsmc.com/english/quarterly-results/2026/q2}{TSMC 2026 Q2 Quarterly Results - Taiwan Semiconductor ...} \\
\fhlink{https://www.fool.com/earnings/call-transcripts/2026/07/16/tsm-tsm-q2-2026-earnings-call-transcript/}{TSM (TSM) Q2 2026 Earnings Call Transcript - The Motley Fool} \\
\fhlink{https://investor.tsmc.com/english/encrypt/files/encrypt_file/reports/2026-07/547d1696765e05ce3adb81c108ce1c8c1682b80c/TSMC+2Q26+Transcript.pdf}{Q2 2026 Taiwan Semiconductor Manufacturing Co Ltd Earnings ...} \\
\fhlink{https://faq.com.tw/en/hardware/2026-04-10-tsmc-q1-2026-record-revenue-en/}{TSMC Posts Record \$35.6 Billion Q1 Revenue as AI Chip Demand...} \\
{\color{black!45}\ldots\ (8 more)}\end{fhresp}
\par\addvspace{3.6mm}\noindent\rbadge{R23}\par\nopagebreak
\par\addvspace{2.3mm}\noindent\hspace*{1.2em}{\footnotesize\color{caseTool}$\circ$}\,\texttt{\textcolor{caseTool}{visit}}: \fhlink{https://investor.tsmc.com/english/encrypt/files/encrypt_file/reports/2026-07/547d1696765e05ce3adb81c108ce1c8c1682b80c/TSMC+2Q26+Transcript.pdf}{investor.tsmc.com}~(goal=Extract management's latest AI demand outlook,\allowbreak{} AI accelerator CAGR,\allowbreak{} advanced-\allowbreak{}node and CoWoS capacity,\allowbreak{} capex,\allowbreak{} margins,\allowbreak{} pricing,\allowbreak{} risks,\allowbreak{} and geographic expansion costs.)\par\nopagebreak
\par\addvspace{3.6mm}\noindent\rbadge{R24}\par\nopagebreak
\par\addvspace{2.3mm}\noindent\hspace*{1.2em}{\footnotesize\color{caseTool}$\circ$}\,\texttt{\textcolor{caseTool}{update\_plan}}: {\color{black!45}\ldots{}~(plan unchanged)}~$\rightarrow$~Plan updated --- 3/4 step(s) done.\par\nopagebreak
\par\addvspace{3.6mm}\noindent\rbadge{R25}\par\nopagebreak
\par\addvspace{2.3mm}\noindent\hspace*{1.2em}{\footnotesize\color{caseTool}$\circ$}\,\texttt{\textcolor{caseTool}{calc}}(expression=75.2/\allowbreak{}81.615*\allowbreak{}100)~$\rightarrow$~\textbf{92.1399}\par\nopagebreak
\par\addvspace{3.6mm}\noindent\rbadge{R26}\par\nopagebreak
\par\addvspace{2.3mm}\noindent\hspace*{1.2em}{\footnotesize\color{caseTool}$\circ$}\,\texttt{\textcolor{caseTool}{update\_plan}}: {\color{black!45}\ldots{}~(plan unchanged)}~$\rightarrow$~Plan updated --- 4/4 step(s) done.\par\nopagebreak
\par\addvspace{3.6mm}\noindent\rbadge{R27}\par\nopagebreak
\par\addvspace{2.3mm}\noindent\hspace*{1.2em}{\footnotesize\color{caseTool}$\circ$}\,\texttt{\textcolor{caseTool}{search}}: [site:eaton.com data center AI power management Eaton 2026, site:arista.com AI networking Ethernet data center 2026 Arista, site:\allowbreak{}investor.\allowbreak{}constellationen\allowbreak{}ergy.\allowbreak{}com data center power AI 2026 Constellation, site:skhynix.com HBM AI data center 2026, site:samsung.com semiconductor HBM AI 2026]~(max\_results=5)\par\nopagebreak
\begin{fhresp}8 results: \\
\fhlink{https://automotive.eaton.com/ph/en-us/markets/data-centers.html}{Data centers | Digitalization | Efficiency | Eaton} \\
\fhlink{http://www.electrical.eaton.com/us/en-us/digital/center-for-intelligent-power.html}{Center for Intelligent Power} \\
\fhlink{http://www.electrical.eaton.com/us/en-us/markets/data-centers.html}{Data centers | Digitalization | Efficiency | Eaton} \\
\fhlink{https://investors.arista.com/Communications/Press-Releases-and-Events/Press-Release-Detail/2026/Arista-Introduces-Next-Generation-1-6Terabit-Portfolio-for-AI-Fabrics/default.aspx}{Arista Networks - Arista Introduces Next-Generation 1.6Terabit ...} \\
{\color{black!45}\ldots\ (4 more)}\end{fhresp}
\par\addvspace{3.6mm}\noindent\rbadge{R28}\par\nopagebreak
\par\addvspace{2.3mm}\noindent\hspace*{1.2em}{\footnotesize\color{caseTool}$\circ$}\,\texttt{\textcolor{caseTool}{visit}}: \fhlink{http://www.electrical.eaton.com/us/en-us/markets/data-centers.html}{electrical.eaton.com}, \fhlink{https://investors.arista.com/Communications/Press-Releases-and-Events/Press-Release-Detail/2026/Arista-Introduces-Next-Generation-1-6Terabit-Portfolio-for-AI-Fabrics/default.aspx}{investors.arista.com}, \fhlink{https://news.skhynix.com/en/2026-market-outlook-focus-on-the-hbm-led-memory-supercycle/}{news.skhynix.com}~(goal=Confirm direct AI-\allowbreak{}data-\allowbreak{}center exposure,\allowbreak{} product role,\allowbreak{} demand outlook,\allowbreak{} and relevant risks.)\par\nopagebreak
\end{tcolorbox}

\clearpage
\subsection{Valuation Compute Seam}
\begin{tcolorbox}[
    breakable, colback=caseTrBack, colframe=caseTrFrame, coltitle=black,
    boxrule=0.5pt, arc=0.5mm, top=2mm, bottom=2mm, left=2mm, right=2mm,
    fonttitle=\usefont{T1}{ppl}{b}{n},
    title={Microsoft (MSFT): DCF, WACC \& Sensitivity}]
\usefont{T1}{ppl}{m}{n}\footnotesize

\textbf{\textsc{Research Trajectory}}\par\smallskip
\par\addvspace{3.6mm}\noindent\rbadge{R1}\par\nopagebreak
\par\addvspace{2.3mm}\noindent\hspace*{1.2em}{\footnotesize\color{caseTool}$\circ$}\,\texttt{\textcolor{caseTool}{update\_plan}}: [in progress]~Collect Microsoft FY2025 reported financials and filing context; [pending]~Pull current market inputs and peer betas; derive WACC; [pending]~Set operating assumptions and build five-year unlevered FCF forecast; [pending]~Value terminal cash flows and bridge enterprise value to equity value; [pending]~Run WACC/perpetual-growth sensitivity and compare with market price; [pending]~Write the auditable DCF report with sources and assumptions~$\rightarrow$~Plan updated --- 0/6 step(s) done.\par\nopagebreak
\par\addvspace{3.6mm}\noindent\rbadge{R2}\par\nopagebreak
\par\addvspace{2.3mm}\noindent\hspace*{1.2em}{\footnotesize\color{caseTool}$\circ$}\,\texttt{\textcolor{caseTool}{load\_skill}}(dcf-valuation)\par\nopagebreak
\begin{fhresp}Skill: \texttt{dcf-valuation} \\
DCF valuation\par
{\color{black!45}\ldots\ (32 more lines)}\par\end{fhresp}
\par\addvspace{2.3mm}\noindent\hspace*{1.2em}{\footnotesize\color{caseTool}$\circ$}\,\texttt{\textcolor{caseTool}{load\_tool}}(data\_equity\_reference, data\_equity\_fundamentals, data\_market\_rates, compute\_valuation\_wacc, compute\_valuation\_dcf, compute\_valuation\_dcf\_sensitivity)\par\nopagebreak
\begin{fhresp}\texttt{data\_\allowbreak{}equity\_\allowbreak{}reference} (data.equity.reference) \\
Reference data for a public equity: name, sector/industry, country, exchange, market cap, shares outstanding, beta, current price, 52-week range, and P/E (snake\_case keys --- chain \texttt{prev:<id>.beta} into wacc and \texttt{prev:\allowbreak{}<id>.\allowbreak{}shares\_\allowbreak{}outstanding} into dcf). Use to identify and classify a ticker (see data.equity.ratios for full multiples incl. dividend yield). \\
Arguments schema:\par
{\color{black!45}\ldots\ (373 more lines)}\par\end{fhresp}
\par\addvspace{2.3mm}\noindent\hspace*{1.2em}{\footnotesize\color{caseTool}$\circ$}\,\texttt{\textcolor{caseTool}{search}}: [Microsoft 2025 10-K revenue operating income depreciation amortization capital expenditures working capital cash taxes FY2025, site:\allowbreak{}microsoft.\allowbreak{}com/\allowbreak{}en-\allowbreak{}us/\allowbreak{}Investor earnings FY 2025 Microsoft annual report]~(max\_results=6)\par\nopagebreak
\begin{fhresp}6 results: \\
\fhlink{https://www.microsoft.com/investor/reports/ar25/index.html}{Microsoft 2025 Annual Report} \\
\fhlink{https://www.microsoft.com/en-us/investor/earnings/fy-2025-q4/press-release-webcast}{FY25 Q4 - Press Releases - Investor Relations - Microsoft} \\
\fhlink{https://www.stocktitan.net/financials/MSFT/}{MSFT - Microsoft Corp Financials - Stock Titan} \\
\fhlink{https://www.microsoft.com/en-us/investor/earnings/fy-2025-q1/press-release-webcast}{FY25 Q1 - Press Releases - Investor Relations - Microsoft} \\
{\color{black!45}\ldots\ (2 more)}\end{fhresp}
\par\addvspace{3.6mm}\noindent\rbadge{R3}\par\nopagebreak
\par\addvspace{2.3mm}\noindent\hspace*{1.2em}{\footnotesize\color{caseTool}$\circ$}\,\texttt{\textcolor{caseTool}{visit}}: \fhlink{https://www.microsoft.com/investor/reports/ar25/index.html}{microsoft.com}, \fhlink{https://www.microsoft.com/en-us/investor/earnings/fy-2025-q4/press-release-webcast}{microsoft.com}~(goal=Extract FY2025 reported revenue,\allowbreak{} operating income,\allowbreak{} tax provision/\allowbreak{}effective tax rate,\allowbreak{} D\&A,\allowbreak{} capex (\allowbreak{}additions to property and equipment)\allowbreak{},\allowbreak{} working-\allowbreak{}capital balance-\allowbreak{}sheet items,\allowbreak{} operating cash flow,\allowbreak{} cash and investments,\allowbreak{} debt,\allowbreak{} diluted shares,\allowbreak{} fiscal year end,\allowbreak{} and management comments relevant to revenue growth/\allowbreak{}margins/\allowbreak{}capital spending.)\par\nopagebreak
\par\addvspace{2.3mm}\noindent\hspace*{1.2em}{\footnotesize\color{caseTool}$\circ$}\,\texttt{\textcolor{caseTool}{data\_equity\_fundamentals}}(ticker=MSFT)\par\nopagebreak
\begin{fhresp}\textbf{MSFT} fundamentals (TTM): \\
Revenue \$318.27B $\cdot$ EBITDA \$184.46B $\cdot$ FCF \$37.01B $\cdot$ OCF \$170.14B \\
Margins --- gross 68.3\%, op 46.3\%, net 39.3\% $\cdot$ ROE 34.0\% \\
Cash \$78.23B $\cdot$ Debt \$125.43B $\cdot$ Rev growth 18.3\% $\cdot$ EPS ttm/fwd 16.79/19.38 \\
Income trend (revenue / op income / net income): \\
  2025-06-30: \$281.72B / \$128.53B / \$101.83B \\
  2024-06-30: \$245.12B / \$109.43B / \$88.14B \\
  2023-06-30: \$211.91B / \$88.52B / \$72.36B \\
  2022-06-30: \$198.27B / \$83.38B / \$72.74B\par\end{fhresp}
\par\addvspace{2.3mm}\noindent\hspace*{1.2em}{\footnotesize\color{caseTool}$\circ$}\,\texttt{\textcolor{caseTool}{data\_equity\_reference}}(ticker=MSFT)\par\nopagebreak
\begin{fhresp}\textbf{Microsoft Corporation} (MSFT) --- NMS $\cdot$ USD \\
Sector: Technology / Software - Infrastructure $\cdot$ Country: United States \\
Market cap: \$2.84T $\cdot$ Shares out: 7.43B $\cdot$ Beta: 1.13 \\
Price: 381.70 $\cdot$ 52w: 349.20--555.45 $\cdot$ P/E (ttm/fwd): 22.73/19.70 \\
Microsoft Corporation develops and supports software, services, devices, and solutions worldwide. \ldots{}\par\end{fhresp}
\par\addvspace{2.3mm}\noindent\hspace*{1.2em}{\footnotesize\color{caseTool}$\circ$}\,\texttt{\textcolor{caseTool}{data\_equity\_reference}}(ticker=ORCL)\par\nopagebreak
\begin{fhresp}\textbf{Oracle Corporation} (ORCL) --- NYQ $\cdot$ USD \\
Sector: Technology / Software - Infrastructure $\cdot$ Country: United States \\
Market cap: \$331.23B $\cdot$ Shares out: 2.88B $\cdot$ Beta: 1.71 \\
Price: 114.99 $\cdot$ 52w: 114.75--345.72 $\cdot$ P/E (ttm/fwd): 19.72/10.56 \\
Oracle Corporation offers products and services that build, run and support enterprise information technology frameworks worldwide. \ldots{}\par\end{fhresp}
\par\addvspace{2.3mm}\noindent\hspace*{1.2em}{\footnotesize\color{caseTool}$\circ$}\,\texttt{\textcolor{caseTool}{data\_equity\_reference}}(ticker=GOOGL)\par\nopagebreak
\begin{fhresp}\textbf{Alphabet Inc.} (GOOGL) --- NMS $\cdot$ USD \\
Sector: Communication Services / Internet Content \& Information $\cdot$ Country: United States \\
Market cap: \$3.91T $\cdot$ Shares out: 5.87B $\cdot$ Beta: 1.25 \\
Price: 319.74 $\cdot$ 52w: 187.82--408.61 $\cdot$ P/E (ttm/fwd): 16.05/21.63 \\
Alphabet Inc. offers various products and platforms in the United States, Europe, the Middle East, Africa, the Asia-Pacific, Canada, and Latin America. \ldots{}\par\end{fhresp}
\par\addvspace{2.3mm}\noindent\hspace*{1.2em}{\footnotesize\color{caseTool}$\circ$}\,\texttt{\textcolor{caseTool}{data\_equity\_reference}}(ticker=AMZN)\par\nopagebreak
\begin{fhresp}\textbf{Amazon.com, Inc.} (AMZN) --- NMS $\cdot$ USD \\
Sector: Consumer Cyclical / Internet Retail $\cdot$ Country: United States \\
Market cap: \$2.50T $\cdot$ Shares out: 10.76B $\cdot$ Beta: 1.46 \\
Price: 232.11 $\cdot$ 52w: 196.00--278.56 $\cdot$ P/E (ttm/fwd): 27.76/23.41 \\
Amazon.com, Inc. engages in the retail sale of consumer products, advertising, and subscriptions service through online and physical stores in North America and internationally. \ldots{}\par\end{fhresp}
\par\addvspace{2.3mm}\noindent\hspace*{1.2em}{\footnotesize\color{caseTool}$\circ$}\,\texttt{\textcolor{caseTool}{data\_equity\_reference}}(ticker=AAPL)\par\nopagebreak
\begin{fhresp}\textbf{Apple Inc.} (AAPL) --- NMS $\cdot$ USD \\
Sector: Technology / Consumer Electronics $\cdot$ Country: United States \\
Market cap: \$4.89T $\cdot$ Shares out: 14.69B $\cdot$ Beta: 1.10 \\
Price: 333.02 $\cdot$ 52w: 201.50--334.99 $\cdot$ P/E (ttm/fwd): 40.32/34.54 \\
Apple Inc. designs, manufactures, and markets smartphones, personal computers, tablets, wearables, and accessories worldwide. \ldots{}\par\end{fhresp}
\par\addvspace{2.3mm}\noindent\hspace*{1.2em}{\footnotesize\color{caseTool}$\circ$}\,\texttt{\textcolor{caseTool}{data\_market\_rates}}()\par\nopagebreak
\begin{fhresp}\textbf{US Treasury yields:} 3m 3.81\% $\cdot$ 5y 4.43\% $\cdot$ 10y 4.68\% $\cdot$ 30y 5.16\% \\
Risk-free (10y) = 4.68\% $\rightarrow$ 0.0468 for WACC.\par\end{fhresp}
\par\addvspace{3.6mm}\noindent\rbadge{R4}\par\nopagebreak
\par\addvspace{2.3mm}\noindent\hspace*{1.2em}{\footnotesize\color{caseTool}$\circ$}\,\texttt{\textcolor{caseTool}{data\_equity\_reference}}(ticker=CRM)\par\nopagebreak
\begin{fhresp}\textbf{Salesforce, Inc.} (CRM) --- NYQ $\cdot$ USD \\
Sector: Technology / Software - Application $\cdot$ Country: United States \\
Market cap: \$134.04B $\cdot$ Shares out: 819.00M $\cdot$ Beta: 1.18 \\
Price: 163.66 $\cdot$ 52w: 146.32--274.00 $\cdot$ P/E (ttm/fwd): 18.99/10.55 \\
Salesforce, Inc. provides customer relationship management technology services that connect companies and customers together in the United States, Europe, and the Asia Pacific. \ldots{}\par\end{fhresp}
\par\addvspace{2.3mm}\noindent\hspace*{1.2em}{\footnotesize\color{caseTool}$\circ$}\,\texttt{\textcolor{caseTool}{data\_equity\_reference}}(ticker=ADBE)\par\nopagebreak
\begin{fhresp}\textbf{Adobe Inc.} (ADBE) --- NMS $\cdot$ USD \\
Sector: Technology / Software - Application $\cdot$ Country: United States \\
Market cap: \$89.48B $\cdot$ Shares out: 397.50M $\cdot$ Beta: 1.43 \\
Price: 225.11 $\cdot$ 52w: 190.12--376.16 $\cdot$ P/E (ttm/fwd): 12.87/8.19 \\
Adobe Inc. operates as a technology company worldwide. \ldots{}\par\end{fhresp}
\par\addvspace{2.3mm}\noindent\hspace*{1.2em}{\footnotesize\color{caseTool}$\circ$}\,\texttt{\textcolor{caseTool}{data\_equity\_reference}}(ticker=IBM)\par\nopagebreak
\begin{fhresp}\textbf{International Business Machines Corporation} (IBM) --- NYQ $\cdot$ USD \\
Sector: Technology / Information Technology Services $\cdot$ Country: United States \\
Market cap: \$201.80B $\cdot$ Shares out: 942.13M $\cdot$ Beta: 0.68 \\
Price: 214.19 $\cdot$ 52w: 199.19--332.46 $\cdot$ P/E (ttm/fwd): 19.01/16.21 \\
International Business Machines Corporation, together with its subsidiaries, provides integrated solutions and services in the Americas, Europe, the Middle East, Africa, and the Asia Pacific. \ldots{}\par\end{fhresp}
\par\addvspace{2.3mm}\noindent\hspace*{1.2em}{\footnotesize\color{caseTool}$\circ$}\,\texttt{\textcolor{caseTool}{search}}: [site:\allowbreak{}microsoft.\allowbreak{}com/\allowbreak{}investor/\allowbreak{}reports/\allowbreak{}ar25 FY2025 Microsoft accounts receivable 2024 unearned revenue accounts payable 2024, Microsoft FY2025 depreciation amortization 34,153 capital expenditures 64,551 changes in working capital annual report, Microsoft 2025 10-k cash flow changes in operating assets liabilities accounts receivable unearned revenue]~(max\_results=8)\par\nopagebreak
\begin{fhresp}12 results: \\
\fhlink{https://www.microsoft.com/investor/reports/ar25/index.html}{Microsoft Annual Report 2025} \\
\fhlink{https://www.scottsdaleaz.gov/docs/default-source/scottsdaleaz/finance/annual-comprehensive-financial-reports/fy2025-annual-comprehensive-financial-report.pdf?sfvrsn=dc8353cd_3}{ANNUAL} \\
\fhlink{https://www.microsoft.com/en-us/investor/default}{Home Page - Microsoft Investor Relations} \\
\fhlink{https://agilebrandguide.com/wiki/metrics/ebitda-earnings-before-interest-taxes-depreciation-and-amortization/}{EBITDA (Earnings Before Interest, Taxes, Depreciation, and...)} \\
{\color{black!45}\ldots\ (8 more)}\end{fhresp}
\par\addvspace{3.6mm}\noindent\rbadge{R5}\par\nopagebreak
\par\addvspace{2.3mm}\noindent\hspace*{1.2em}{\footnotesize\color{caseTool}$\circ$}\,\texttt{\textcolor{caseTool}{visit}}: \fhlink{https://www.microsoft.com/en-us/investor/earnings/fy-2025-q4/cash-flows}{microsoft.com}, \fhlink{https://www.microsoft.com/en-us/investor/earnings/fy-2025-q4/balance-sheets}{microsoft.com}, \fhlink{https://www.microsoft.com/en-us/investor/earnings/fy-2025-q4/income-statements}{microsoft.com}~(goal=Extract FY2025 and FY2024 cash-\allowbreak{}flow statement details,\allowbreak{} especially D\&A,\allowbreak{} capex/\allowbreak{}additions to PP\&E,\allowbreak{} cash flow changes in accounts receivable,\allowbreak{} inventories,\allowbreak{} other current assets,\allowbreak{} accounts payable,\allowbreak{} unearned revenue,\allowbreak{} other current liabilities; plus balance sheet cash,\allowbreak{} short-\allowbreak{}term investments,\allowbreak{} debt and diluted weighted average shares.)\par\nopagebreak
\par\addvspace{3.6mm}\noindent\rbadge{R6}\par\nopagebreak
\par\addvspace{2.3mm}\noindent\hspace*{1.2em}{\footnotesize\color{caseTool}$\circ$}\,\texttt{\textcolor{caseTool}{calc}}(expression=128528/\allowbreak{}281724)~$\rightarrow$~\textbf{0.45622}\par\nopagebreak
\noindent\hspace*{1.2em}{\footnotesize\color{caseTool}$\circ$}\,\texttt{\textcolor{caseTool}{calc}}(expression=21795/\allowbreak{}123627)~$\rightarrow$~\textbf{0.176296}\par\nopagebreak
\noindent\hspace*{1.2em}{\footnotesize\color{caseTool}$\circ$}\,\texttt{\textcolor{caseTool}{calc}}(expression=34153/\allowbreak{}281724)~$\rightarrow$~\textbf{0.121229}\par\nopagebreak
\noindent\hspace*{1.2em}{\footnotesize\color{caseTool}$\circ$}\,\texttt{\textcolor{caseTool}{calc}}(expression=64551/\allowbreak{}281724)~$\rightarrow$~\textbf{0.229129}\par\nopagebreak
\par\addvspace{3.6mm}\noindent\rbadge{R7}\par\nopagebreak
\par\addvspace{2.3mm}\noindent\hspace*{1.2em}{\footnotesize\color{caseTool}$\circ$}\,\texttt{\textcolor{caseTool}{compute\_valuation\_wacc}}(risk\_free\_rate=0.0468, equity\_risk\_premium=0.05, beta=1.406, cost\_of\_debt=0.0518, tax\_rate=0.176296, debt\_to\_equity=0.015194)\par\nopagebreak
\begin{fhresp}\textbf{Cost of capital} (CAPM + WACC) \\
Re = Rf + b*ERP = 4.7\% + 1.41*5.0\% = \textbf{11.7\%} \\
Implied ERP (Re-Rf = b*ERP): 7.0\% \\
After-tax Rd = 5.2\%*(1-17.6\%) = 4.3\% \\
Weights (D/E=0.015): E/V 98.5\%, D/V 1.5\% \\
\textbf{WACC = 11.6\%} $\cdot$ premium over Rf 6.9\%\par\end{fhresp}
\par\addvspace{3.6mm}\noindent\rbadge{R8}\par\nopagebreak
\par\addvspace{2.3mm}\noindent\hspace*{1.2em}{\footnotesize\color{caseTool}$\circ$}\,\texttt{\textcolor{caseTool}{calc}}(expression=0.0468+\allowbreak{}1.406*\allowbreak{}0.05)~$\rightarrow$~\textbf{0.1171}\par\nopagebreak
\noindent\hspace*{1.2em}{\footnotesize\color{caseTool}$\circ$}\,\texttt{\textcolor{caseTool}{calc}}(expression=(\allowbreak{}0.015194/\allowbreak{}(\allowbreak{}1+\allowbreak{}0.015194)\allowbreak{})\allowbreak{})~$\rightarrow$~\textbf{0.0149666}\par\nopagebreak
\noindent\hspace*{1.2em}{\footnotesize\color{caseTool}$\circ$}\,\texttt{\textcolor{caseTool}{calc}}(expression=1/\allowbreak{}(\allowbreak{}1+\allowbreak{}0.015194)\allowbreak{})~$\rightarrow$~\textbf{0.985033}\par\nopagebreak
\par\addvspace{3.6mm}\noindent\rbadge{R9}\par\nopagebreak
\par\addvspace{2.3mm}\noindent\hspace*{1.2em}{\footnotesize\color{caseTool}$\circ$}\,\texttt{\textcolor{caseTool}{calc}}(expression=281.724*\allowbreak{}1.14)~$\rightarrow$~\textbf{321.165}\par\nopagebreak
\noindent\hspace*{1.2em}{\footnotesize\color{caseTool}$\circ$}\,\texttt{\textcolor{caseTool}{calc}}(expression=281.724*\allowbreak{}1.14*\allowbreak{}1.13)~$\rightarrow$~\textbf{362.917}\par\nopagebreak
\noindent\hspace*{1.2em}{\footnotesize\color{caseTool}$\circ$}\,\texttt{\textcolor{caseTool}{calc}}(expression=281.724*\allowbreak{}1.14*\allowbreak{}0.457)~$\rightarrow$~\textbf{146.773}\par\nopagebreak
\noindent\hspace*{1.2em}{\footnotesize\color{caseTool}$\circ$}\,\texttt{\textcolor{caseTool}{calc}}(expression=281.724*\allowbreak{}1.14*\allowbreak{}0.135)~$\rightarrow$~\textbf{43.3573}\par\nopagebreak
\par\addvspace{3.6mm}\noindent\rbadge{R10}\par\nopagebreak
\par\addvspace{2.3mm}\noindent\hspace*{1.2em}{\footnotesize\color{caseTool}$\circ$}\,\texttt{\textcolor{caseTool}{compute\_valuation\_dcf}}(fcf\_schedule=[91.4768, 114.964, 139.717, \ldots; 5 values], discount\_rate=0.115986, terminal\_method=gordon\_growth, terminal\_growth\_rate=0.025, net\_debt=-66.819, shares\_outstanding=7.465)\par\nopagebreak
\begin{fhresp}\textbf{DCF valuation} \\
Discount rate 11.6\% $\cdot$ 5y explicit $\cdot$ Gordon growth (g=2.5\%, FCF\_N+1=\$187) \\
Sum PV explicit FCF: \$485 \\
Terminal value (yr N): \$2,058 $\cdot$ PV: \$1,189 \\
\textbf{Enterprise value: \$1,674} $\cdot$ net debt -\$67 \\
\textbf{Equity value: \$1,741} \\
\textbf{Intrinsic value/share: \$233} \\
Terminal value is 71.0\% of EV.\par\end{fhresp}
\par\addvspace{2.3mm}\noindent\hspace*{1.2em}{\footnotesize\color{caseTool}$\circ$}\,\texttt{\textcolor{caseTool}{compute\_valuation\_dcf\_sensitivity}}(fcf\_schedule=[91.4768, 114.964, 139.717, \ldots; 5 values], discount\_rates=[0.095986, 0.105986, 0.115986, \ldots; 5 values], terminal\_method=gordon\_growth, terminal\_growth\_rates=[0.015, 0.02, 0.025, \ldots; 5 values], net\_debt=-66.819, shares\_outstanding=7.465)\par\nopagebreak
\begin{fhresp}\textbf{DCF sensitivity} (Gordon growth) --- intrinsic value/share\par
\par\smallskip\begin{center}\resizebox{\ifdim\width>\linewidth\linewidth\else\width\fi}{!}{%
\begin{tabular}{lrrrrr}
\toprule
\textbf{WACC \textbackslash{} g} & \textbf{1.5\%} & \textbf{2.0\%} & \textbf{2.5\%} & \textbf{3.0\%} & \textbf{3.5\%} \\
\midrule
9.6\% & \$272 & \$285 & \$301 & \$319 & \$340 \\
10.6\% & \$241 & \$251 & \$263 & \$276 & \$291 \\
11.6\% & \$216 & \$224 & \$233 & \$243 & \$255 \\
12.6\% & \$196 & \$202 & \$209 & \$217 & \$226 \\
13.6\% & \$179 & \$184 & \$190 & \$196 & \$203 \\
\bottomrule
\end{tabular}}\end{center}\par\smallskip
\textbf{Range (intrinsic value/share): \$179 -- \$340}\par\end{fhresp}
\end{tcolorbox}

\end{document}